\documentclass[]{meta_preamble/fairmeta}
\pdfoutput=1

\usepackage{verbatim}
\usepackage{amsmath,amsfonts,amssymb} %
\usepackage{pifont}
\newcommand{\cmark}{\ding{51}}
\newcommand{\xmark}{\ding{55}}
\usepackage{bbm}
\usepackage{xspace}

\usepackage{url} %
\usepackage{algorithm} %
\usepackage{algpseudocode} %
\usepackage{array} %
\usepackage{tabularx} %
\usepackage{tikz}
\usetikzlibrary{calc}
\usepackage{adjustbox}
\usepackage{wrapfig}
\usepackage{nicematrix} %
\usepackage{makecell} %

\usepackage{lmodern}

\newtoggle{arxiv}
\toggletrue{arxiv}

\definecolor{metared}{RGB}{230, 51, 51}
\definecolor{primalcolor}{HTML}{A60000}
\definecolor{contrarycolor}{HTML}{00A6A6}
\definecolor{darkcontrarycolor}{HTML}{004C4C}
\definecolor{lightblue}{HTML}{2970CC}
\definecolor{lightpurple}{HTML}{673147}
\definecolor{ForestGreen}{HTML}{FF5733}
\definecolor{LimeGreen}{HTML}{90EE90}
\definecolor{myred}{HTML}{AA4A44}
\definecolor{hyppurple}{HTML}{800080}

\newcommand{\term}{{physics-based retargeting}}
\newcommand{\Term}{{Physics-based retargeting}}
\newcommand{\TERM}{{Physics-based Retargeting}}

\newif\ifincludenote
\includenotefalse
\ifincludenote
    \newcommand{\chaoyi}[1]{{\color{purple} Chaoyi: #1}}
    \newcommand{\guanya}[1]{{\color{cyan} Guanya: #1}}
    \newcommand{\akash}[1]{{\color{blue} Akash: #1}}
    \newcommand*{\HB}[1]{\textcolor{brown}{[homanga: #1]}}
    \newcommand*{\fhogan}[1]{\textcolor{red}{[fhogan: #1]}}

    \newcommand{\todo}[1]{{\color{red} TODO: #1}}
    \definecolor{haozhipurpole}{RGB}{122, 133, 193}
    \newcommand{\haozhi}[1]{{\color{haozhipurpole} Haozhi: #1}}

\else
    \newcommand{\chaoyi}[1]{}
    \newcommand{\guanya}[1]{}
    \newcommand{\akash}[1]{}
    \newcommand{\HB}[1]{}
    \newcommand{\fhogan}[1]{}
    \newcommand{\todo}[1]{}
    \newcommand{\haozhi}[1]{}
\fi

\newif\ifusepng
\iftoggle{arxiv}{\usepngfalse}{\usepngtrue}
\ifusepng
    
\else
    
\fi

\definecolor{metafg}{RGB}{100, 100, 100}
\definecolor{metabg}{RGB}{245, 245, 245}
\iftoggle{arxiv}{
    \newcommand{\graybold}[1]{{\color{metafg} {\textbf{#1}}}}
    
}{
    \newcommand{\graybold}[1]{\textbf{#1}}
    
}

\newcommand{\numcite}[1]{\ifcsname citenum\endcsname[\citenum{#1}]\else\cite{#1}\fi}
\newcommand{\taskname}[1]{{\texttt{#1}}}
\newcommand{\algname}[1]{\textbf{\texttt{#1}}}

\newcommand{\dexycb}{\taskname{DexYCB}\xspace}
\newcommand{\oakinkimage}{\taskname{OakInk-Image}\xspace}
\newcommand{\hrdexdb}{\taskname{HRDexDB}\xspace}
\newcommand{\hotthree}{\taskname{HOT3D}\xspace}
\newcommand{\oakink}{\taskname{Oakink}\xspace}
\newcommand{\gigahand}{\taskname{GigaHands}\xspace}
\newcommand{\lafan}{\taskname{LAFAN1}\xspace}
\newcommand{\amass}{\taskname{AMASS}\xspace}
\newcommand{\omomo}{\taskname{OMOMO}\xspace}
\newcommand{\arctic}{\taskname{ARCTIC}\xspace}

\newcommand{\sharpa}{\taskname{Sharpa}\xspace}
\newcommand{\allegro}{\taskname{Allegro}\xspace}
\newcommand{\inspire}{\taskname{Inspire}\xspace}
\newcommand{\schunk}{\taskname{Schunk}\xspace}
\newcommand{\xhand}{\taskname{XHand}\xspace}
\newcommand{\ability}{\taskname{Ability}\xspace}

\newcommand{\panda}{\taskname{Franka Emika Panda}\xspace}
\newcommand{\unitreeg}{\taskname{Unitree G1}\xspace}
\newcommand{\unitreeh}{\taskname{Unitree H1-2}\xspace}
\newcommand{\fourier}{\taskname{Fourier N1}\xspace}
\newcommand{\booster}{\taskname{Booster T1}\xspace}

\newcommand{\fpskinoak}{$\mathbf{248.9}$\xspace}
\newcommand{\fpsspideroak}{$6.1$\xspace}
\newcommand{\fpskingiga}{$\mathbf{244.2}$\xspace}
\newcommand{\fpsspidergiga}{$5.9$\xspace}
\newcommand{\fpsspidercoak}{$0.4$\xspace}
\newcommand{\fpsspidercgiga}{$0.5$\xspace}

\newcommand{\spider}{\algname{SPIDER}\xspace}

\newcommand{\dexmachina}{\algname{DexMachina}\xspace}
\newcommand{\maniptrans}{\algname{ManipTrans}\xspace}
\newcommand{\gmr}{\algname{GMR}\xspace}
\newcommand{\kinematic}{\algname{Kinematic}\xspace}

\newcommand{\methodfull}{\underline{S}calable \underline{P}hysics-\underline{I}nformed \underline{DE}xterous \underline{R}etargeting}

\renewcommand{\abstract}[1]{\def\abstractlist{{\color{black}#1}}}

\patchcmd{\mymaketitle}{\tcbset{top=0.5cm}}{\tcbset{top=0.3cm}}{}{\errmessage{Title top-padding patch failed}}
\patchcmd{\mymaketitle}{\tcbset{bottom=0.5cm}}{\tcbset{bottom=0.3cm}}{}{\errmessage{Title bottom-padding patch failed}}
\patchcmd{\mymaketitle}{\setlength{\parskip}{0.5cm}}{\setlength{\parskip}{0.25cm}}{}{\errmessage{Title paragraph-spacing patch failed}}
\patchcmd{\mymaketitle}{\vskip 0.5cm}{\vskip 0.15cm}{}{\errmessage{Title metadata-spacing patch failed}}

\newcommand{\titleicon}{\raisebox{-0.2ex}{\includegraphics[height=2ex]{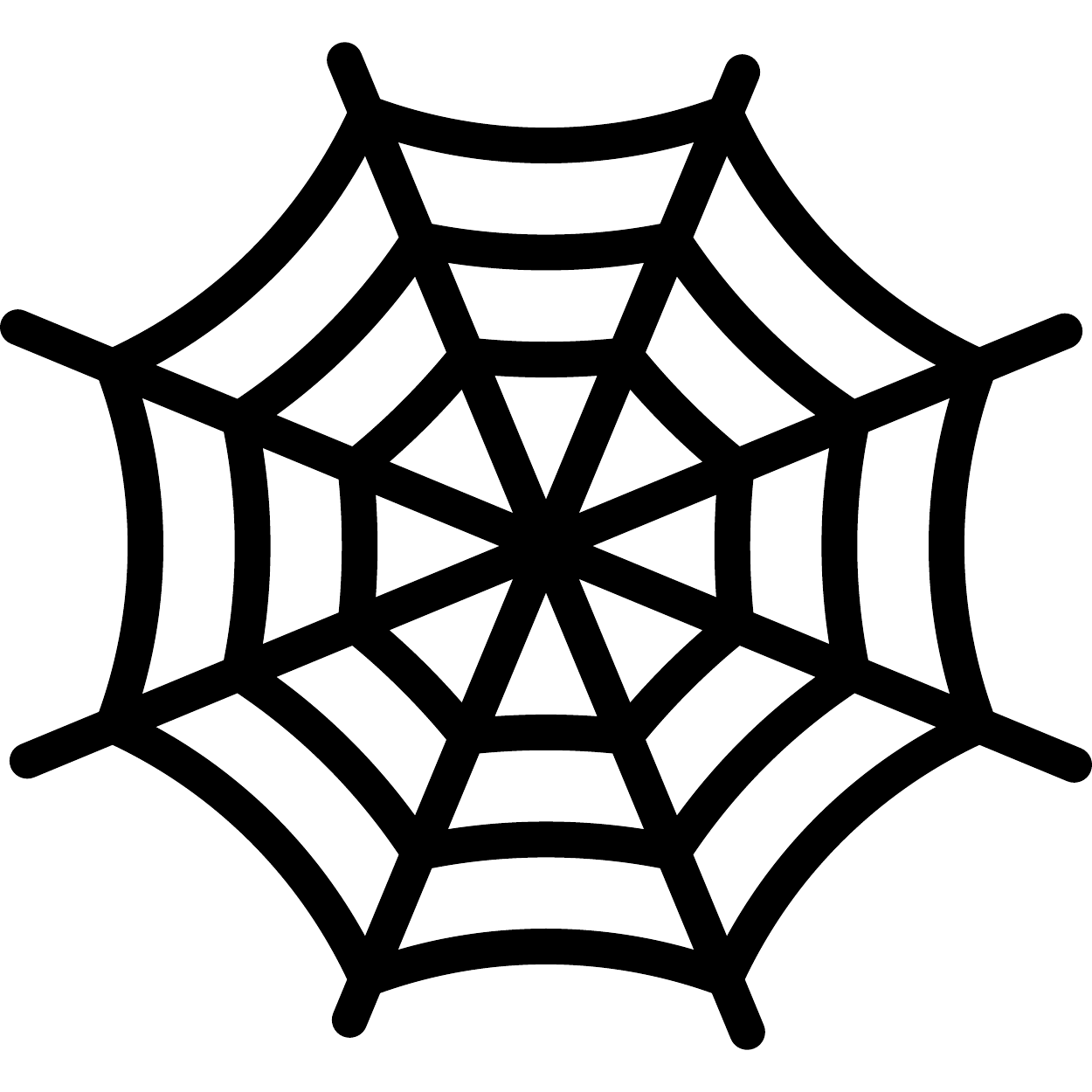}}}
\title{\titleicon\,SPIDER: Scalable Physics-Informed Dexterous Retargeting}

\author[1,2,*]{Chaoyi Pan}
\author[1]{Changhao Wang}
\author[1]{Haozhi Qi}
\author[1]{Julen Urain}
\author[1]{Zixi Liu}
\author[1]{Homanga Bharadhwaj}
\author[1]{Akash Sharma}
\author[1,\dagger]{Tingfan Wu}
\author[2,\dagger]{Guanya Shi}
\author[1,\dagger]{Jitendra Malik}
\author[1,\dagger]{Francois Hogan}

\affiliation[1]{FAIR at Meta}
\affiliation[2]{Carnegie Mellon University}
\contribution[*]{Work done at Meta}
\contribution[\dagger]{Joint last author}

\abstract{Learning agile robotic policies requires large-scale demonstrations, but translating abundant human motion data to robots is bottlenecked by the embodiment gap and missing dynamic information.
To bridge this gap, we propose \methodfull{} (\spider), a physics-based retargeting framework to transform and augment kinematic-only human demonstrations into dynamically feasible robot trajectories at scale.
Our key insight is that human demonstrations should provide global task structure and objective, while a sampling-based solver can be used to find the feasible solution given the physics constraints.
As a general framework, \spider is an efficient physics-based retargeting method that can be applied to both humanoid whole-body loco-manipulation and dexterous manipulation across $9$ humanoid/dexterous hand embodiments, $6$ datasets and $3$ simulators.
By bypassing the need for policy optimization, it achieves state-of-the-art performance while being 10x faster than reinforcement learning (RL) baselines.
Furthermore, \spider enables physics-based data augmentation—such as imposing external payloads, perturbations, and new contact patterns—to generate diverse data.
We demonstrate that the retargeted motion can be executed on the robot directly open-loop or serve as feasible reference motion for efficient RL policy learning.
Our pipeline enables large-scale generation of robot trajectories from human motion, and we release a dataset of 6,520 simulated demonstrations across four hands and 191 dataset-specific object identities to support future research.

}

\date{\today}
\correspondence{\email{chaoyip@andrew.cmu.edu}}
\metadata[Code]{\url{https://github.com/facebookresearch/spider}}
\metadata[Website]{\url{https://jc-bao.github.io/spider-project}}
\metadata[Dataset]{\url{https://huggingface.co/datasets/retarget/retarget_full}}

\begin{document}

\maketitle

\noindent\begin{minipage}{\textwidth}
    \centering
    \captionsetup{skip=5pt}
    \includegraphics[width=0.88\textwidth]{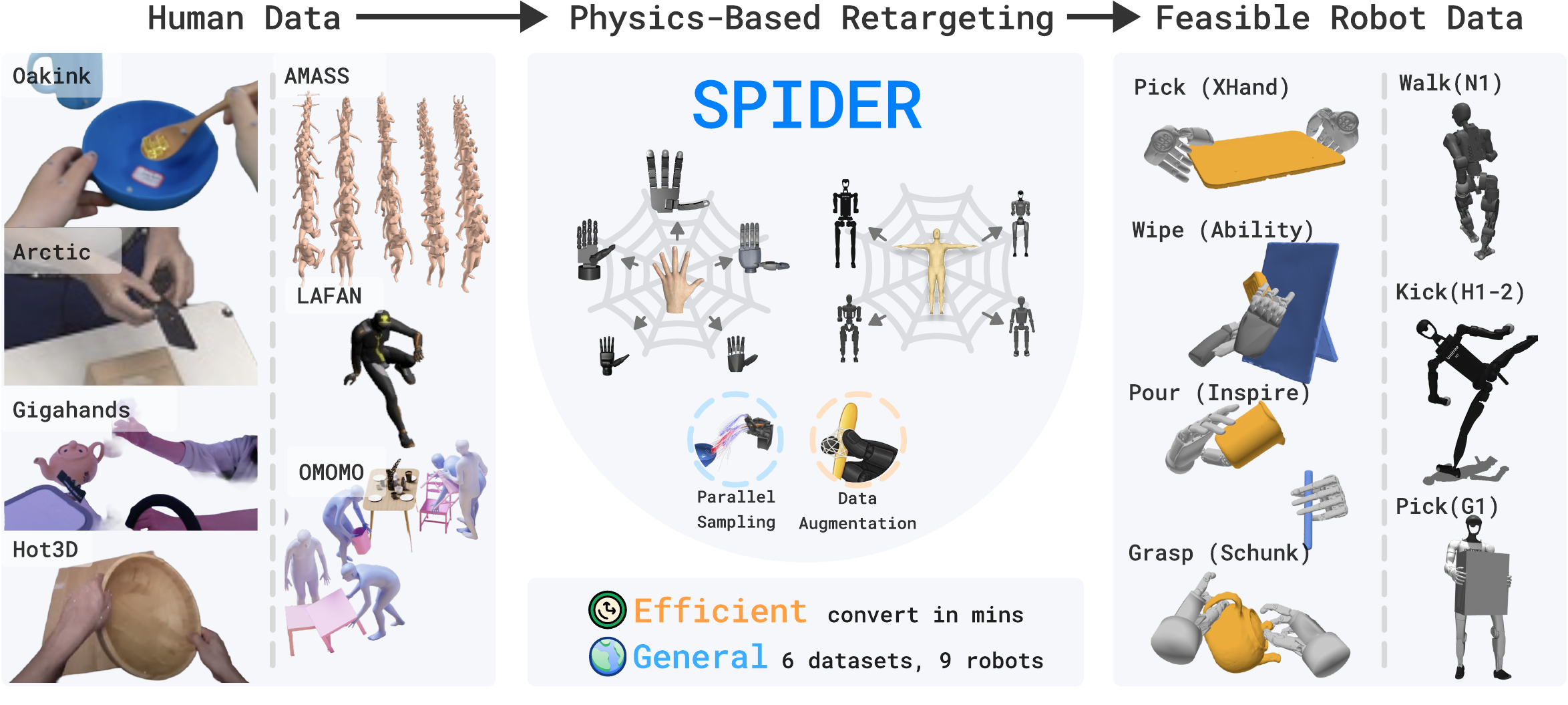}
    \captionof{figure}{\textbf{Overview of \spider.} The sampling-based physics pipeline converts human-object interaction into dynamically feasible robot trajectories across embodiments and task domains, with virtual contact guidance preserving intended contact modes.}
    \label{fig:teaser}
\end{minipage}
\par\medskip

\section{Introduction}

Whole-body dexterous manipulation has rapidly advanced by learning from massive internet-scale datasets~\citep{kimOpenVLAOpenSourceVisionLanguageAction2024} and human demonstrations~\citep{rdt2}.
Yet, acquiring generalizable whole-body manipulation across embodiments, from dexterous hands~\citep{yinDexterityGenFoundationController2025,zhongDexGraspVLAVisionLanguageActionFramework2025,qi2023general} to humanoid whole-body control~\citep{liAMOAdaptiveMotion2025,zeTWISTTeleoperatedWholeBody2025}, remains prohibitively expensive due to hardware constraints, task complexity, and lack of large-scale embodiment-specific data~\citep{zhangDOGloveDexterousManipulation2025,xuDexUMIUsingHuman2025}.
On the other hand, abundant human motion data is readily available, including large-scale motion-capture datasets~\citep{zhanOAKINK2DatasetBimanual2024,mahmoodAMASSArchiveMotion2019}, internet-scale video collections~\citep{renMotionTracksUnified2025}, and VR-based interfaces~\citep{hoqueEgoDexLearningDexterous2025}.
Recent advances in computer vision have made it possible to reconstruct 3D human body~\citep{goelHumans4DReconstructing2023} and hand motion~\citep{pavlakosReconstructingHands3D2023}, as well as object meshes~\citep{xiangStructured3DLatents2025} and trajectories~\citep{wenFoundationPoseUnified6D2024}, directly from videos.
These developments create a unique opportunity to leverage human motion as a scalable source of demonstrations for learning humanoid and dexterous robot control.
However, a fundamental challenge arises: the \emph{embodiment gap} - the mismatch in morphology, dynamics, and actuation between humans and robots - which leads to infeasible motion during transfer.
This motivates our central research question:

\begin{quote}
      How can we efficiently and reliably transform human motion into feasible robot trajectories that respect dynamics and contact?
\end{quote}

We formulate this as a \term{} problem~\citep{redaPhysicsbasedMotionRetargeting2023}: given kinematic human demonstrations, generate robot motions that (a) align the robot's poses with human poses, (b) follow physics constraints, and (c) preserve the task objectives of the demonstrated behavior (e.g. object manipulation).
The problem arises from the following challenges:
(a) \emph{Flexibility and generalizability:}
given diverse data sources and target embodiments, the framework should work for diverse tasks and human-object interactions.
(b) \emph{Scalability and efficiency:}
the abundance of internet-scale human data requires approaches that are both computationally efficient and scalable to large datasets.
Existing methods struggle to bridge these gaps:
inverse-kinematics (IK) approaches~\citep{qinOneHandMultiple2023,yinGeometricRetargetingPrincipled2025} are efficient but dynamically infeasible;
reinforcement learning (RL) approaches~\citep{liManipTransEfficientDexterous2025,lumCrossingHumanRobotEmbodiment2025} are general but require expensive trajectory-specific policy optimization;
teleoperation~\citep{yinDexterityGenFoundationController2025} is dynamically feasible but often labor-intensive and embodiment-specific.

To achieve scalable, general, and flexible physics-based cross-embodiment retargeting, we propose \methodfull{} (\spider), a sampling-based approach with contact guidance.
Our key insight is that human demonstrations provide high-level guidance in terms of robot motion and task specification, while a physics simulator grounds trajectories to ensure dynamical feasibility and contact correctness through large-scale sampling.
Our method lies in the middle of the trade-off between gradient-based optimization and reinforcement learning: it preserves the efficiency of optimization-based methods as a learning-free method, while achieving the generalizability of reinforcement learning as no gradient is required.
The resulting framework is embodiment-agnostic and task-general:
it can be applied to any robot-environment interaction as long as the scene can be simulated.

\graybold{Contributions.} Our contributions are summarized as follows:
\begin{itemize}
      \item We introduce \spider, a \emph{general} and \emph{scalable} physics-based retargeting framework that supports both dexterous manipulation and humanoid whole-body loco-manipulation across 9 diverse robot embodiments, 6 datasets, and 3 simulators. It achieves state-of-the-art performance while being $10\times$ faster than RL-based methods.
      \item We demonstrate our method's \emph{flexibility} and \emph{applicability} to diverse policy learning and data generation applications, including more efficient RL policy learning, data augmentation for novel physics constraints, user intent preservation with contact guidance, and trajectory robustification for direct real-robot deployment.
        \item Our pipeline enables the generation of large-scale, robot-feasible datasets, with 6532 demonstrations across 4 distinct robotic hands and 191 different objects, derived from human data. We release the full data generation pipeline to assist future research. 
\end{itemize}

\section{Related Work}

\subsection{Learning Manipulation from Human Data}

Recent work explores learning robot skills from large-scale human videos by first detecting actions and transferring them to robots.
One common strategy is kinematic retargeting, where human poses or keypoints are mapped to robot motions, as in DemoDiffusion~\citep{parkDemoDiffusionOneShotHuman2025}, OKAMI~\citep{liOKAMITeachingHumanoid2024}, R+X~\citep{papagiannisR+XRetrievalExecution2025}, and EgoZero~\citep{liuEgoZeroRobotLearning2025}.
Another strategy is to train a human-centric policy and then adapt it to robots through fine-tuning with in-domain data, as in MimicPlay~\citep{wangMimicPlayLongHorizonImitation2023}, Track2Act~\citep{bharadhwaj2024track2act}, and VideoDex~\citep{shawVideoDexLearningDexterity2022}.
Our approach is complementary to these pipelines. Specifically, \spider can serve as a drop-in replacement for the human-to-robot action transfer components, providing more accurate and contact-aware mappings for dexterous and whole-body control.

\subsection{Retargeting from Human Data}

Motion retargeting seeks to convert human data into robot trajectories that are physically consistent and executable.

\graybold{Kinematic Retargeting.}
These methods map human motion to robot configurations~\citep{qinOneHandMultiple2023}. While efficient and easy to compute, they often rely on specialized hardware~\citep{xuDexUMIUsingHuman2025}, handcrafted motion primitives~\citep{wuOneShotTransferLongHorizon2024}, and struggle with realism in contact-rich tasks~\citep{qinAnyTeleopGeneralVisionBased2024}. Being kinematic only, the generated motions are not compliant with physics constraints.

\graybold{Learning-Based Retargeting Networks.}
Neural mapping approaches train networks to convert human motions into robot motions~\citep{parkDemoDiffusionOneShotHuman2025,yinDexterityGenFoundationController2025}. Such models can outperform direct kinematic mappings and retain fast inference, but they require extensive pretraining and may fail when facing out-of-distribution motions or novel embodiments.

\graybold{Optimization-Based Retargeting.}
Optimization-based approaches explicitly incorporate physics and contact constraints to ensure dynamical feasibility~\citep{redaPhysicsbasedMotionRetargeting2023}. They can generate high-quality, physically plausible motions, but often depend on detailed contact data~\citep{lakshmipathyKinematicMotionRetargeting2024}, specific data pipelines~\citep{yangPhysicsDrivenDataGeneration2025}, and strong priors~\citep{nakaokaTaskModelLower2005}.
Due to the non-convex nature of the problem, sampling-based approaches~\citep{yangPhysicsDrivenDataGeneration2025,si2025exostartefficientlearningdexterous} have emerged as a promising solution.

\graybold{RL-Based Retargeting.}
RL has been used to retarget human demonstrations across embodiments~\citep{lumCrossingHumanRobotEmbodiment2025,liManipTransEfficientDexterous2025}. When combined with curriculum learning~\citep{mandiDexMachinaFunctionalRetargeting2025,liuQuasiSimParameterizedQuasiPhysical2024}, RL can produce dexterous, physically feasible robot motions. However, it typically requires training on each trajectory and significant computation, which limits scalability to internet-scale data and real-time deployment.

Existing methods trade off between efficiency (kinematics, neural networks) and physical fidelity (optimization, RL). \spider combines the generality of RL with the efficiency of optimization-based pipelines, making it a scalable and practical drop-in replacement for human-to-robot transfer.

\subsection{Sampling-based Optimization for Robot Control}

Sampling-based optimization methods such as the cross-entropy method~\citep{deboerTutorialCrossEntropyMethod2005}, evolutionary algorithms~\citep{salimansEvolutionStrategiesScalable2017}, and Bayesian optimization~\citep{frazierTutorialBayesianOptimization2018a} are powerful tools for solving non-convex and non-smooth problems.
Due to their parallelizability and flexibility, these methods have been applied to navigation~\citep{williamsAggressiveDrivingModel2016}, legged locomotion~\citep{xueFullOrderSamplingBasedMPC2024}, and dexterous manipulation~\citep{howellPredictiveSamplingRealtime2022,liDROPDexterousReorientation2024}. Despite their success in contact-rich control, they can suffer from instability and solution ambiguity in trajectory sampling~\citep{kimSmoothModelPredictive2022}.
\spider addresses these challenges by guiding sampling with contact information, which helps preserve the intended contact sequence.

n
\begin{figure*}[t]
    \centering
    \includegraphics[width=\textwidth]{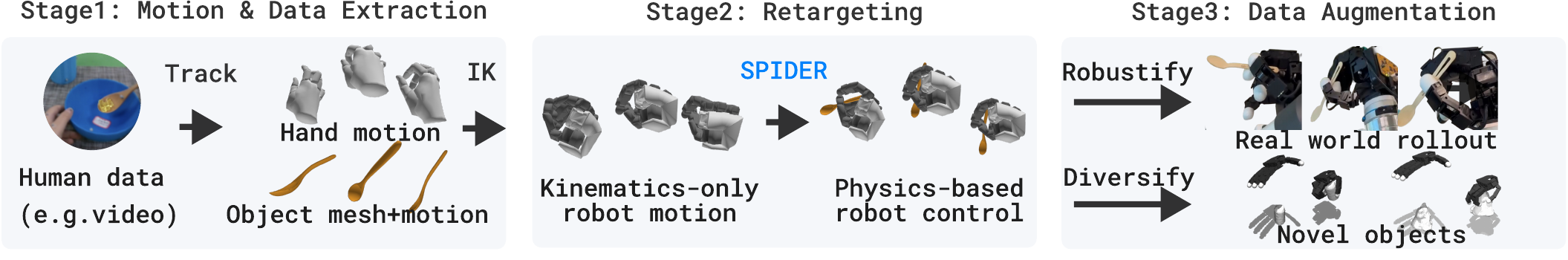}
    \caption{\textbf{Overview of the \spider pipeline.} The pipeline takes reconstructed object meshes, reference robot motion, and object motion and converts them into a dynamically feasible robot trajectory with corrected contacts. The generated trajectory is further robustified and augmented before deployment or policy learning.}
    \label{fig:method}
    \vspace{-0.4cm}
\end{figure*}

\section{Physics-based Retargeting with Sampling}
\label{sec:method}

This section introduces our pipeline for converting human data to feasible robot trajectories, as illustrated in~\Cref{fig:method}.
We first formulate the \term{} problem in~\Cref{subsec:motion_transfer_problem}, followed by its sampling-based solver design in~\Cref{subsec:sampling_based_solver}.
Then, we introduce different data augmentation methods in~\Cref{subsec:data_augmentation} for downstream applications including trajectory robustification for sim-to-real transfer, data augmentation for novel physics constraints and user intent preservation with contact guidance.

\subsection{\TERM{} Problem}
\label{subsec:motion_transfer_problem}

We formulate \term{} as a constrained optimization problem where a robot control sequence $u_{0:T-1}$ (e.g., joint position or torque commands) is optimized to minimize the distance to the reference trajectory $x^{\text{ref}}_{0:T}$ and the control effort.
As \emph{input}, a kinematic reference state sequence $x^{\text{ref}}_t, \forall t \in \{0, \ldots, T\}$ is first provided from human demonstrations.
The state $x^\text{ref}_t = \{q^\text{ref}_t, v^\text{ref}_t\}$ is composed of reference generalized coordinates $q^\text{ref}_t$ and generalized velocities $v^\text{ref}_t$.
Specifically, for robots with $n_u$ joints, the coordinates $q^\text{ref}_t$ comprise the robot joint positions $q^\text{ref,joint}_{t} \in \mathbb{R}^{n_u}$, the robot base transformation $X^\text{ref,base}_{t} \in \mathrm{SE}(3)$, and the object pose $X^\text{ref,object}_{t} \in \mathrm{SE}(3)$.
\Term{} is performed \emph{offline}: the \emph{output} is a dynamically feasible open-loop control sequence $u_{0:T-1}$ (e.g., the joint position commands used in this paper), valid under aligned initial conditions between the robot/object and the reference.
Real-robot experiments demonstrate feasibility from such aligned starts.
The optimization minimizes the distance to the reference trajectory and the control effort:

\iftoggle{arxiv}{
    \begin{subequations}
        \begin{align}
            \begin{split}
                \min_{u_{0:T-1}} J(u_{0:T-1})
                 & = \min_{u_{0:T-1}}  \|x_{T} \ominus x_{T}^\text{ref}\|_{Q_T}^2 +
                \sum_{t=0}^{T-1} \left( \left\| x_{t+1} \ominus x_{t+1}^\text{ref}\right\|_{Q_t}^2 + \|u_t\|_{R_t}^2 \right)
            \end{split} \label{eq:dynamical_retargeting_problem} \\
            \text{s.t.} \quad x_{t+1} & = f(x_t, u_t, t) \quad \forall t \in \{0, \ldots, T-1\}                         \\
            x_{0:T}                   & \in \mathcal{X}, \quad u_{0:T-1} \in \mathcal{U}
        \end{align}
    \end{subequations}
}{
    \begin{subequations}
        \small
        \begin{align}
            \begin{split}
                \min_{u_{0:T-1}} J(u_{0:T-1})
                 & = \min_{u_{0:T-1}}  \|x_{T} \ominus x_{T}^\text{ref}\|_{Q_T}^2 +                                             \\
                 & \sum_{t=0}^{T-1} \left( \left\| x_{t+1} \ominus x_{t+1}^\text{ref}\right\|_{Q_t}^2 + \|u_t\|_{R_t}^2 \right)
            \end{split} \label{eq:dynamical_retargeting_problem} \\
            \text{s.t.} \quad x_{t+1} & = f(x_t, u_t, t) \quad \forall t \in \{0, \ldots, T-1\}                                                 \\
            x_{0:T}                   & \in \mathcal{X}, \quad u_{0:T-1} \in \mathcal{U}
        \end{align}
    \end{subequations}
}

where $x_{0:T}$ is the optimized feasible state, $x_{t+1} = f(x_t, u_t, t)$ is the state transition function, $\mathcal{X}$ is the state space, $\mathcal{U}$ is the control input space, and $Q_t$ and $R_t$ are the state and control input weighting matrices.
Because the generalized coordinates contain poses on the $\mathrm{SE}(3)$ manifold, standard subtraction is undefined. We use the operator $\ominus$ to denote the difference in generalized coordinates, which evaluates the error between variables in Euclidean space normally, while mapping differences between $\mathrm{SE}(3)$ transformations to the local 6D tangent space using the logarithmic map.
In practice, we use diagonal weighting matrices for $Q_t$ and $R_t$, where $Q_t = \text{diag}(\{q_\text{robot}, q_\text{object}\})$ and $R_t = \text{diag}(\{r_\text{robot}, r_\text{object}\})$.

\subsection{Sampling for \TERM{}}
\label{subsec:sampling_based_solver}

\begin{figure*}[ht]
    \centering
    \includegraphics[width=1.0\linewidth]{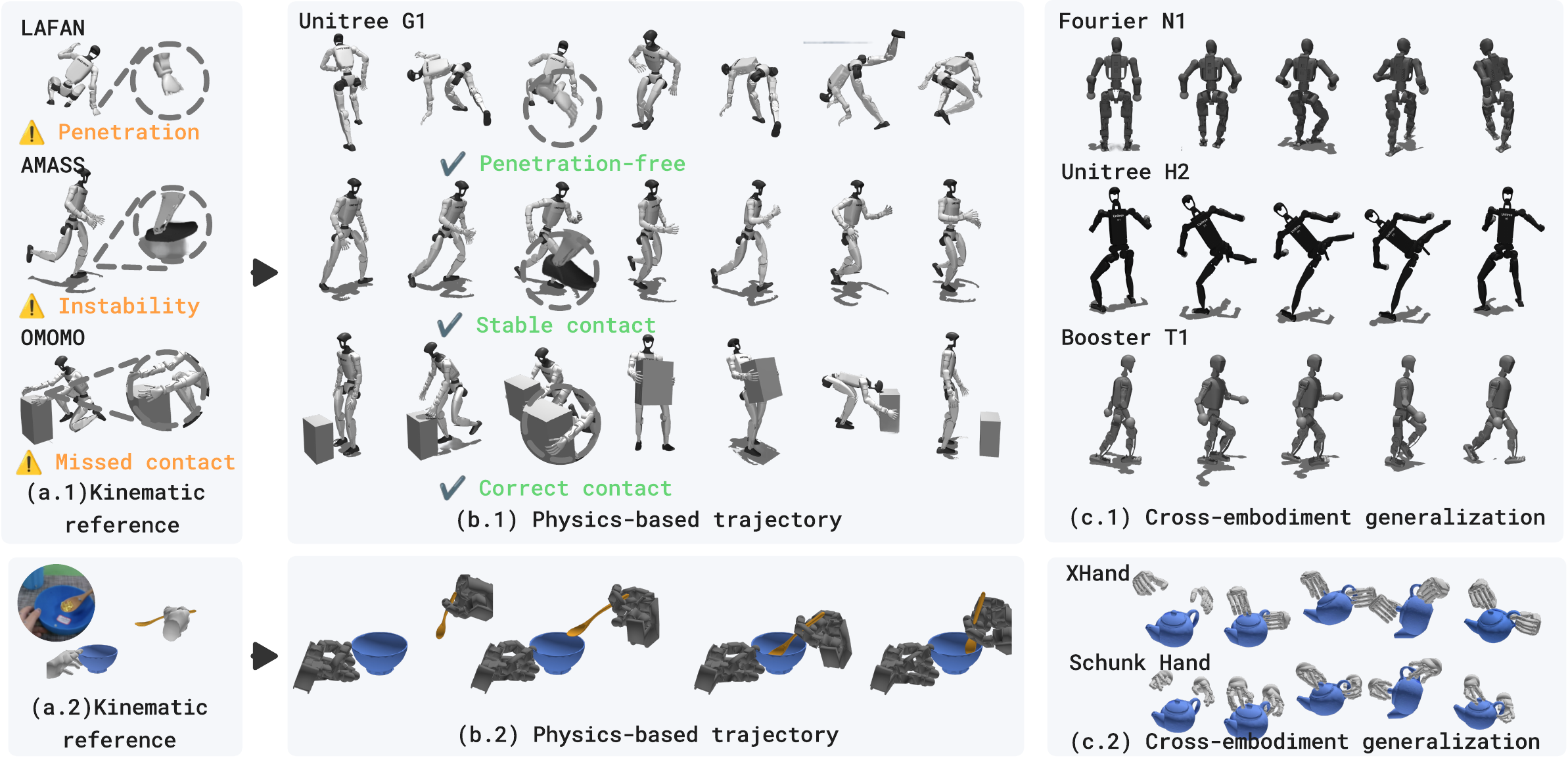}
    \caption{
        \textbf{Sampling for \TERM{} on different robot embodiments.}
        This figure shows the open-loop rollout of the retargeted control on different robot embodiments.
        The retargeting process corrects artifacts such as foot penetration and slipping, enabling the robot to execute highly dynamic motions.
        The method can be applied to both dexterous manipulation and humanoid whole-body control tasks.
    }
    \label{fig:cross_embodiment}
\end{figure*}

\begin{algorithm}[h]
    \caption{Sampling for \TERM{}}
    \label{alg:spider}
    \begin{algorithmic}[1]
        \State \textbf{Input:} $x^{\text{ref}}_{0:T}$, $U^0$, $H$, $N_W$, $\lambda$, $\beta_1, \beta_2$, $N$, $\epsilon_{\text{tol}}$
        \State $J_{\text{best}} \leftarrow \infty$, $i \leftarrow 0$
        \While{$i < N$ and not converged}
        \State $i \leftarrow i + 1$; Update $\Sigma^i_{h}$ using \Cref{eq:cov} \iftoggle{arxiv}{\Comment{Annealing schedule for sampling covariance}}{}
        \For{$j \in \{1, \ldots, N_W\}$}
        \State Sample {\iftoggle{arxiv}{}{\footnotesize}$[W]_j \sim \mathcal{N}(0, \Sigma^i_{0:H-1})$; $[U^{i}]_j \leftarrow U^{i} + [W]_j$} \label{alg:spider:sample}
        \State Compute $J([U^{i}]_j)$ by simulation rollout (\Cref{eq:dynamical_retargeting_problem})
        \EndFor
        \State Update $U^{i+1}$ using weighted average from \Cref{eq:dial_mpc}
        \If{$|J_{\text{best}} - \min_j J(U_j)| < \epsilon_{\text{tol}}$} \label{alg:spider:early_stop}
        \State \textbf{break} \Comment{Early stop if improvement is small}
        \EndIf
        \State $J_{\text{best}} \leftarrow \min_j J(U_j)$
        \EndWhile
        \State \textbf{Return} $U^*$
    \end{algorithmic}
\end{algorithm}

Due to the contact-rich nature of \term{}, the optimization problem in \Cref{eq:dynamical_retargeting_problem} is highly non-convex and often discontinuous.
Sampling-based optimization~\citep{mannorCrossEntropyMethod} provides a natural way to handle such landscapes, as it does not rely on smoothness or convexity assumptions.
Intuitively, this resembles RL in that both rely on sampled trajectories from parallelized simulation to guide decision-making, but instead of updating a policy network, we directly use the samples to optimize the control sequence $U = u_{0:T-1}$.
To this end, we adopt a sampling-based optimizer with an annealed sampling kernel~\citep{xueFullOrderSamplingBasedMPC2024,panModelBasedDiffusionTrajectory2024} as shown in \Cref{alg:spider}.
\iftoggle{arxiv}{}{\small}%
\begin{align}
    U^{i+1}      & = U^i + \frac{\sum_{j=1}^{N_W} \exp\left(-\frac{J(U^i + [W]_j)}{\lambda}\right) [W]_j}{\sum_{j=1}^{N_W} \exp\left(-\frac{J(U^i + [W]_j)}{\lambda}\right)}, \label{eq:dial_mpc} \\
    \Sigma^i_{h} & = \exp\left(-\frac{N-i}{\beta_1 N} - \frac{H-h}{\beta_2 H} \right) I, \label{eq:cov}
\end{align}

where $H$ is the planning horizon, $\lambda$ is the temperature parameter, $U^i$ is the solution at iteration $i$, $i = 1, \ldots, N$ is the iteration index with $N$ being the total number of iterations, $[W]_{j} \sim \mathcal{N}(0, \Sigma_{0:H-1}),~j=1,\ldots, N_W$ are $N_W$ sampled Gaussian noise sequences with scheduled covariance $\Sigma_h, h\in \{0, \ldots, H-1\}$, and $\beta_1, \beta_2 \in (0, 1)$ are annealing parameters controlling sampling covariance in line~\ref{alg:spider:sample} of \Cref{alg:spider}.

In contact-rich dexterous manipulation, the \emph{feasible solution set is typically narrow}, so effective optimization requires a combination of \emph{coarse search} to discover feasible contact modes and \emph{fine refinement} to achieve stable contact.
The annealed sampling covariance in \Cref{eq:dial_mpc} implements this exploration-exploitation trade-off: $\beta_1$ controls the rate across outer iterations (global-to-local search over optimization updates), and $\beta_2$ controls the rate along the prediction horizon (allocating more or less perturbation across timesteps).
As the schedule progresses, the effective sampling radius transitions from broad exploration to targeted exploitation around promising trajectories.
To further improve speed, we adopt an early stopping strategy~\citep{kamatBITKOMOCombiningSampling2022}, where optimization halts when improvements from further sampling become small.

\subsection{Sampling for Physics-based Data Augmentation}
\label{subsec:data_augmentation}

Apart from generating feasible robot actions, another application of \term{} is the ability to systematically augment the retargeted data for trajectory robustness, data diversity and user intent preservation.

\graybold{Trajectory Robustification.}
To bridge the gap from reconstructed demonstrations to real hardware, trajectories must be robust against unknown or misspecified dynamics (e.g., friction, contact compliance) and reconstruction noise (e.g., object mesh estimation error, pose estimation error) that could otherwise make a nominal plan infeasible.
Our approach optimizes a control sequence with a pessimistic (min--max) objective over a bounded parameter set $\mathcal{D}$, similar to the domain randomization (DR) method~\citep{tobinDomainRandomizationTransferring2017a} but replacing expectation with worst-case cost to ensure universal feasibility:
\begin{equation}
    J_{\text{rob}}(U) = \max_{d \in \mathcal{D}} J(U, d),
\end{equation}
where $J(U, d)$ is the cost in \Cref{eq:dynamical_retargeting_problem} under dynamics parameter $d$.
In practice, $\mathcal{D}$ spans variations such as contact margin size, friction coefficients, and object mass.
During optimization, each candidate sequence $U_j$ is rolled out over a mini-batch $d_{1:K} \sim \mathcal{D}$ and evaluated by $\max_{k \le K} J(U_j, d_k)$, following robust sampling-based control~\citep{williamsRobustSamplingBased2018}.

\begin{figure}[t]
    \centering
    \includegraphics[width=1.0\linewidth]{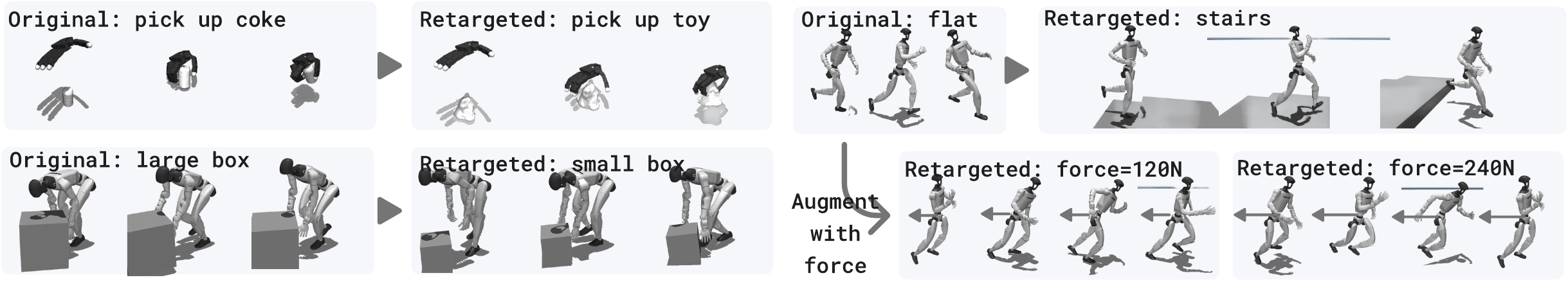}
    \caption{
        \textbf{Geometric and physics augmentation.}
        We augment the retargeted data from a single demonstration into a diverse set of physically feasible actions.
        Here we demonstrate (a) generating motion with a new object mesh for dexterous manipulation, (b) moving a lighter and smaller object for humanoid robot tasks, (c) adding stairs to the scene for humanoid running motion, and (d) applying external forces to the robot when it is pulling a heavy object.
    }
    \label{fig:data_augmentation}
\end{figure}

\graybold{Geometric and Physics Augmentation.}
To generate diverse data, one human demonstration can be augmented into a diverse set of physically feasible actions that could be used for policy learning~\citep{mandlekar2023mimicgen}.
\spider supports generating diverse geometric variations of the objects while still grounding them with physics.
For dexterous manipulation tasks, we perturb the initial object pose and scale to generate diverse interaction behaviors while the reference motion is kept the same.
We can even directly replace the object mesh to generate new interaction behaviors.
For instance, in \Cref{fig:data_augmentation}, we adapt a grasping motion for a Coke bottle to a cat toy simply by replacing the object mesh.
Similarly, the same idea can be applied to human-object interaction for humanoid robot tasks.
By modifying the object geometry to make it smaller and lighter, the robot can automatically adapt to the new object by lowering its body and switching to a single-hand pushing motion with less force.
One can also change the terrain to transform a flat-ground running motion into a stair-climbing motion, which is challenging for kinematic-only retargeting because climbing stairs requires new contact patterns.
A key advantage of physics-based retargeting is its force-awareness.
By applying external forces to the robot when it is pulling a heavy object, the sampling-based solver can generate different behaviors that respond to the force.
For instance, if a large force ($120$\,N and $240$\,N, respectively, in~\Cref{fig:data_augmentation}) is applied to pull the robot back, the generated trajectory will lean forward and move more slowly.

\begin{figure}[t]
    \centering
    \includegraphics[width=1.0\linewidth]{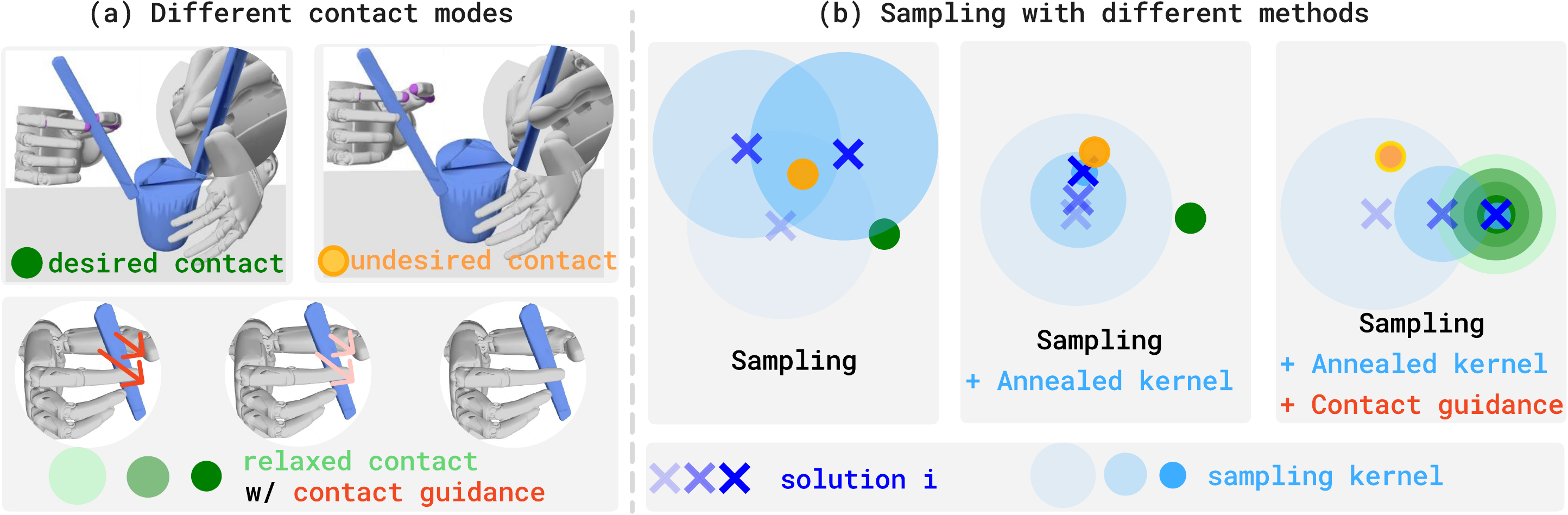}
    \caption{
        \textbf{Human-preferred contact preservation.} Due to a poor initial reference, the solution may converge to a different contact mode than the human demonstration. Contact guidance corrects this behavior by making the target contact mode easier to sample. With a virtual force applied between the object and the hand, the object is more likely to move towards the target contact region, guiding the sampling towards the desired contact mode.
    }
    \label{fig:contact_mismatch}
    \label{fig:guidance}
\end{figure}

\graybold{Human-preferred Contact Preservation.}
Due to the multimodal nature of dexterous manipulation, solution ambiguity could lead to unexpected contact behaviors that do not match the reference motion, as illustrated in \Cref{fig:contact_mismatch}.
For gradient-based optimization methods, one can specify the contact mode in dynamics constraints to enforce the desired contact mode.
Inspired by this idea, we apply a virtual constraint between intended contact points on the object and robot fingers, ``sticking'' the object to the target configuration in early stages and gradually relaxing this constraint in a curriculum-like fashion.
For the $k$-th object-hand contact pair (e.g., handle-thumb), we define the current relative position as ${}^\text{robot}p_{k,t}^\text{object} = p^\text{robot}_{k,t} - p^\text{object}_{k,t}$ and the desired reference position as ${}^\text{robot}p_{k,t}^{\text{object,ref}} = p^\text{robot,ref}_{k,t} - p^\text{object,ref}_{k,t}$.
The constraint is activated when the contact indicator $c_{k,t} = 1$, which occurs when the reference relative position is within the contact threshold: $c_{k,t} = \mathbf{1}(\|{}^\text{robot}p_{k,t}^{\text{object,ref}}\|_2 \leq \epsilon_\text{contact})$.
The constraint strength is controlled by a penalty parameter $\eta_i \rightarrow \infty$ when $i \rightarrow N$:
\begin{equation}
    c_{k,t} \|{}^\text{robot}p_{k,t}^\text{object} - {}^\text{robot}p_{k,t}^{\text{object,ref}}\|_2^2 \leq \eta_i
\end{equation}
The implementation details are described in \Cref{sec:contact_guidance_details}.
Compared to previous methods~\citep{mandiDexMachinaFunctionalRetargeting2025}, our formulation focuses on \emph{relative} hand--object relationships instead of their absolute positions.
The method requires annotations of the intended contact points and labels, which we detect automatically using the simulator's collision detection on kinematic data.

\section{Experiments}
\label{sec:experiments}

This section evaluates \spider across four applications: retargeting human demonstrations to feasible robot trajectories, downstream policy training, physical-robot deployment, and large-scale dataset generation.
We report retargeting performance separately for dexterous manipulation and humanoid whole-body control in~\Cref{subsec:tracking_performance}, then evaluate policy learning for each embodiment in~\Cref{subsec:policy_training}.
To verify the generated trajectory feasibility, we apply \spider to in-the-wild data and deploy the resulting motions on physical robots in~\Cref{subsec:deployment_results}.
Finally, we apply \spider to retarget common hand-object interaction datasets at scale and report its performance in~\Cref{subsec:large_scale_results}.
Common failure cases and limitations are discussed in~\Cref{subsec:failure_cases}.
Throughout, we adopt unified notation for tracking errors: $e_t$ and $e_r$ denote translation and rotation errors, $e_j$ denotes joint angle error, $e_{ft}$ denotes fingertip position error, and $e_\text{arti}$ denotes articulation joint angle error for articulated objects.
Full implementation details including datasets, robot embodiments, simulator settings, and hyperparameters are provided in~\Cref{sec:implementation_details}.

\subsection{Retargeting Performance}
\label{subsec:tracking_performance}

\begin{table}[t]
    \centering
    \begin{minipage}[t]{0.55\textwidth}
        \begin{minipage}[t][7.5em][t]{\linewidth}
            \captionof{table}{\textbf{Dexterous manipulation tracking.}
                Comparisons align each RL or kinematic baseline's setting.
                ADD-AUC evaluates object-pose tracking from $0$ to $10$\,cm; ``mixed'' aggregates five hands.
                FPS uses one NVIDIA L40S; full results are in \Cref{sec:comparison_with_baselines}.}
            \label{tab:dexterous_comparison_main}
        \end{minipage}
        \resizebox{\linewidth}{!}{%
            \begin{NiceTabular}{ll|lll|cc}
                \CodeBefore
                \rectanglecolor{metabg}{3-6}{3-7}
                \rectanglecolor{metabg}{5-6}{5-7}
                \rectanglecolor{metabg}{6-7}{6-7}
                \rectanglecolor{metabg}{8-6}{8-6}
                \rectanglecolor{metabg}{9-7}{9-7}
                \rectanglecolor{metabg}{11-6}{11-6}
                \Body
                \toprule
                                             & \textbf{Method}    & \textbf{Simulator}     & \textbf{Dataset} & \textbf{Hand} & AUC (\%) $\uparrow$      & FPS $\uparrow$ \\
                \midrule
                \Block{4-1}{\rotate vs.\ RL} & \maniptrans        & \taskname{Isaac Gym}   & \oakink          & \inspire      & $19.4 \pm 27.5$          & $0.1$          \\
                                             & \spider            & \taskname{Isaac Gym}   & \oakink          & \inspire      & $\mathbf{22.3 \pm 28.1}$ & $\mathbf{2.5}$ \\
                \cmidrule{2-7}
                                             & \dexmachina        & \taskname{Genesis}     & \arctic          & \inspire      & $41.0 \pm 25.3$          & $0.05$         \\
                                             & \spider            & \taskname{Genesis}     & \arctic          & \inspire      & $\mathbf{43.6 \pm 29.8}$ & $\mathbf{1.5}$ \\
                \midrule
                \Block{6-1}{\rotate vs.\ kin.}
                                             & \kinematic         & \taskname{MuJoCo}      & \oakink          & mixed         & $11.0 \pm 11.1$          & \fpskinoak      \\
                                             & \spider            & \taskname{MuJoCo Warp} & \oakink          & mixed         & $48.8 \pm 24.9$          & \fpsspideroak   \\
                                             & \algname{SPIDER+C} & \taskname{MuJoCo Warp} & \oakink          & mixed         & $\mathbf{56.9 \pm 24.7}$ & \fpsspidercoak  \\
                \cmidrule{2-7}
                                             & \kinematic         & \taskname{MuJoCo}      & \gigahand        & mixed         & $24.1 \pm 22.3$          & \fpskingiga     \\
                                             & \spider            & \taskname{MuJoCo Warp} & \gigahand        & mixed         & $56.9 \pm 20.6$          & \fpsspidergiga  \\
                                             & \algname{SPIDER+C} & \taskname{MuJoCo Warp} & \gigahand        & mixed         & $\mathbf{57.7 \pm 23.9}$ & \fpsspidercgiga \\
                \bottomrule
            \end{NiceTabular}}
    \end{minipage}
    \hfill
    \begin{minipage}[t]{0.43\textwidth}
        \begin{minipage}[t][7.5em][t]{\linewidth}
            \captionof{table}{\textbf{Humanoid whole-body tracking.}
                Open-loop robot and object tracking.
                Robot errors are joint (degrees), pelvis translation (cm), and orientation (degrees); object errors apply only to \omomo (cm/degrees).
                FPS is trajectory-generation throughput.}
            \label{tab:humanoid_tracking_results}
        \end{minipage}
        \resizebox{\linewidth}{!}{%
            \begin{NiceTabular}{ll|ccc cc c}
                \CodeBefore
                \rectanglecolor{metabg}{4-3}{4-5}
                \rectanglecolor{metabg}{6-3}{6-5}
                \rectanglecolor{metabg}{8-3}{8-5}
                \rectanglecolor{metabg}{8-6}{8-7}
                \rectanglecolor{metabg}{3-8}{3-8}
                \rectanglecolor{metabg}{5-8}{5-8}
                \rectanglecolor{metabg}{7-8}{7-8}
                \Body
                \toprule
                                 &                 & \multicolumn{3}{c}{\textbf{Robot}} & \multicolumn{2}{c}{\textbf{Object}} &                           \\
                \cmidrule(lr){3-5} \cmidrule(lr){6-7}
                \textbf{Data} & \textbf{Method} & $e_j$ $\downarrow$           & $e_t$ $\downarrow$             & $e_r$ $\downarrow$ & $e_t$ $\downarrow$ & $e_r$ $\downarrow$ & \textbf{FPS} $\uparrow$ \\
                \midrule
                \Block{2-1}{\lafan}
                                 & \gmr            & $1.08$                             & $2.01$                              & $2.40$                   & --                      & --                       & $\mathbf{35.2}$         \\
                                 & \spider         & $\mathbf{0.58}$                    & $\mathbf{0.11}$                     & $\mathbf{0.07}$          & --                      & --                       & $23.1$                  \\
                \midrule
                \Block{2-1}{\amass}
                                 & \gmr            & $6.2$                              & $4.1$                               & $18.7$                   & --                      & --                       & $\mathbf{37.2}$         \\
                                 & \spider         & $\mathbf{0.75}$                    & $\mathbf{0.23}$                     & $\mathbf{0.08}$          & --                      & --                       & $22.0$                  \\
                \midrule
                \Block{2-1}{\omomo}
                                 & \gmr            & $0.57$                             & $1.43$                              & $2.28$                   & $1.18$                  & $1.68$                   & $\mathbf{28.0}$         \\
                                 & \spider         & $\mathbf{0.83}$                    & $\mathbf{0.20}$                     & $\mathbf{0.17}$          & $\mathbf{0.18}$         & $\mathbf{0.06}$          & $19.6$                  \\
                \bottomrule
            \end{NiceTabular}}
    \end{minipage}
\end{table}

\subsubsection{Dexterous Manipulation}
\label{subsec:dexterous_retargeting_results}

\graybold{Baselines.}
We compare against two RL-based physics retargeting methods, \maniptrans~\citep{liManipTransEfficientDexterous2025} and \dexmachina~\citep{mandiDexMachinaFunctionalRetargeting2025}, and a \kinematic retargeting baseline (fingertip-based IK, kinematically feasible but dynamically infeasible).
Since RL-based methods train per-trajectory policies whereas \spider directly optimizes control sequences, this comparison serves as a performance reference.
To isolate the effect of retargeting from confounds, each comparison aligns the simulator, dataset, hand, and reward setting with the corresponding baseline: we evaluate against \maniptrans on \oakink in \taskname{Isaac Gym} and against \dexmachina on \arctic in \taskname{Genesis}, both using the 12-DoF \inspire hand, and additionally align the nominal base controller with \maniptrans's universal base imitator; the kinematic comparison is run in \taskname{MuJoCo}.
Full protocol details are in \Cref{sec:comparison_with_baselines}.

\graybold{Metrics.}
We report two metrics.
\emph{ADD-AUC} measures object tracking quality without committing to a single success threshold: we sample keypoints on the object mesh, transform them by both the achieved and the reference object poses at every frame, and take the area under the per-frame success-rate curve as the mean keypoint distance threshold sweeps from $0$ to $10$\,cm (normalized so the metric lies in $[0, 100]\%$).
Because it is computed on transformed mesh keypoints, it captures both translational and rotational error in a single number; the full definition is in \Cref{sec:implementation_details}.
We use ADD-AUC rather than a thresholded success rate because success is defined differently across baseline papers and neither has released a public evaluation protocol.
\emph{FPS} measures retargeting throughput, defined as the total number of retargeted frames divided by the total retargeting wall-clock time (higher is better).

\graybold{Comparison Results.}
\Cref{tab:dexterous_comparison_main} aggregates all dexterous-hand comparisons.
Against the RL-based baselines, \spider achieves comparable or better ADD-AUC while generating trajectories over an order of magnitude faster, since it optimizes control sequences with a parallel simulator instead of training a policy per trajectory.
Against \kinematic retargeting, \spider more than doubles the ADD-AUC, and contact guidance (\algname{SPIDER+C}, with contacts detected by MuJoCo's contact detector) improves it further.
The high variance is because results are aggregated across tasks of varying difficulty; detailed per-task, per-hand tracking performance is reported in \Cref{sec:comparison_with_baselines}.

\subsubsection{Humanoid Whole-Body Control}
\label{subsec:humanoid_retargeting_results}

\graybold{Baselines and Reference Methods.}
For humanoid whole-body loco-manipulation tasks, there is no existing physics-based retargeting method since the kinematic motion is directly used as the reference motion for RL training.
Thus, we compare \spider with the state-of-the-art kinematics-based retargeting method general motion retargeting (\gmr)~\citep{araujo2025retargetingmattersgeneralmotion}, where we report the open-loop rollout motion tracking error.
Since humanoid robots are unstable systems, a stabilizer policy is required to track the retargeted motion, where reference motion quality is crucial.
For \lafan, we use representative dance and stand-up sequences; for \amass, we randomly sample eight motions; and for \omomo, we evaluate an object-transport trajectory.
The table reports averages over every frame and sequence in each dataset.
Robot translation and rotation errors measure pelvis tracking, while the object columns measure pose tracking for the \omomo interaction.

\graybold{Comparison Against Kinematics-Based Baselines.}
As summarized in~\Cref{tab:humanoid_tracking_results}, \spider achieves lower open-loop tracking error thanks to physics grounding with minimal speed degradation, including significantly better object tracking on the \omomo dataset.

\subsection{Policy Training}
\label{subsec:policy_training}

\begin{figure}[t]
    \centering
    \begin{minipage}[t]{\textwidth}
        \centering
        \includegraphics[width=\linewidth]{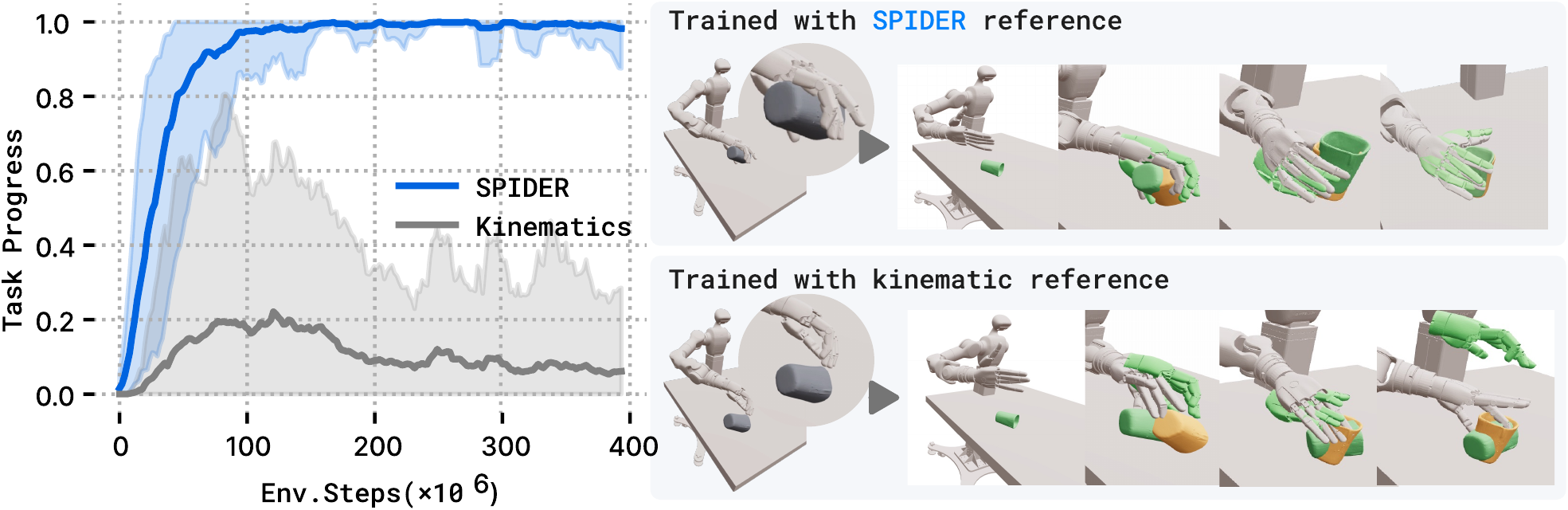}
        \captionof{figure}{\textbf{Dexterous grasping policy training.}
            With identical hyperparameters over seven seeds, the \spider reference yields fast, consistent task completion; the kinematic reference contains physical artifacts and fails. Yellow shows learned motion and green shows the reference.}
        \label{fig:grasp_rl_training}
    \end{minipage}
    \vspace{1em}

    \begin{minipage}[t]{\textwidth}
        \centering
        \includegraphics[width=\linewidth]{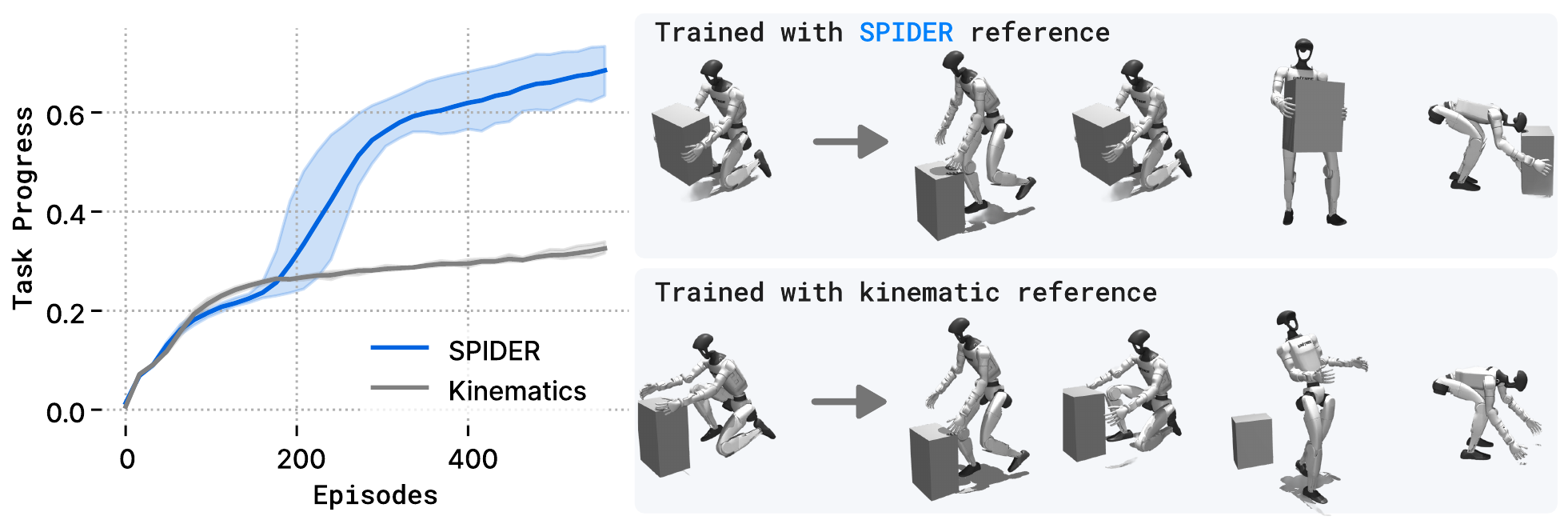}
        \captionof{figure}{\textbf{Humanoid loco-manipulation policy training.}
            The original reference misses box contact, whereas \spider corrects contact and supplies feasible feedforward control. The resulting policy converges faster and completes more of the target motion.}
        \label{fig:rl_training}
    \end{minipage}
\end{figure}

\subsubsection{Dexterous Hand Policy Training}
\spider provides an already-feasible reference motion, making the policy's tracking problem substantially easier.
We evaluate a tabletop cup-grasping task with a fixed-base 7-DoF right arm and a 22-DoF, five-fingered hand.
The policy outputs 6D wrist and 22D finger target deltas at 20\,Hz and is trained with PPO in 4096 parallel environments using only imitation rewards for object pose, wrist pose, finger joints, and hand landmarks.
The reference comes from an in-the-wild human pouring demonstration reconstructed by the pipeline in~\Cref{par:in_the_wild}.
Occlusion causes the kinematic reference to let the cup drift out of the hand; \spider removes this artifact before training.
Over seven seeds with identical hyperparameters, the \spider-reference policy completes the task ($0.99 \pm 0.01$ task progress), whereas the kinematic-reference policy stalls ($0.07 \pm 0.11$), as shown in~\Cref{fig:grasp_rl_training}.
Full action, observation, reward, reset, and PPO settings are given in~\Cref{subsec:grasp_policy_training_details}.

\subsubsection{Humanoid Policy Training}
We evaluate a whole-body loco-manipulation policy that must pick up a box, transport it, and place it on the floor using the framework of~\citet{weng2025hdmilearninginteractivehumanoid}.
We train PPO with body and object tracking rewards plus an action-rate penalty, while removing contact rewards and auxiliary regularizers so the comparison isolates reference-motion quality.
The original kinematic motion is offset from the object and misses the grasp contact; the \spider version corrects the contact and provides a dynamically feasible nominal control.
Under identical training settings, the kinematic-reference policy learns body tracking but fails to grasp the object, while the \spider-reference policy converges faster and achieves substantially better object tracking (\Cref{fig:rl_training}).
Humanoid reward and termination details are provided in~\Cref{subsec:rl_policy_training_details}.

\subsection{Deployment Results}
\label{subsec:deployment_results}

\begin{figure*}[tb]
    \centering
    \includegraphics[width=\textwidth]{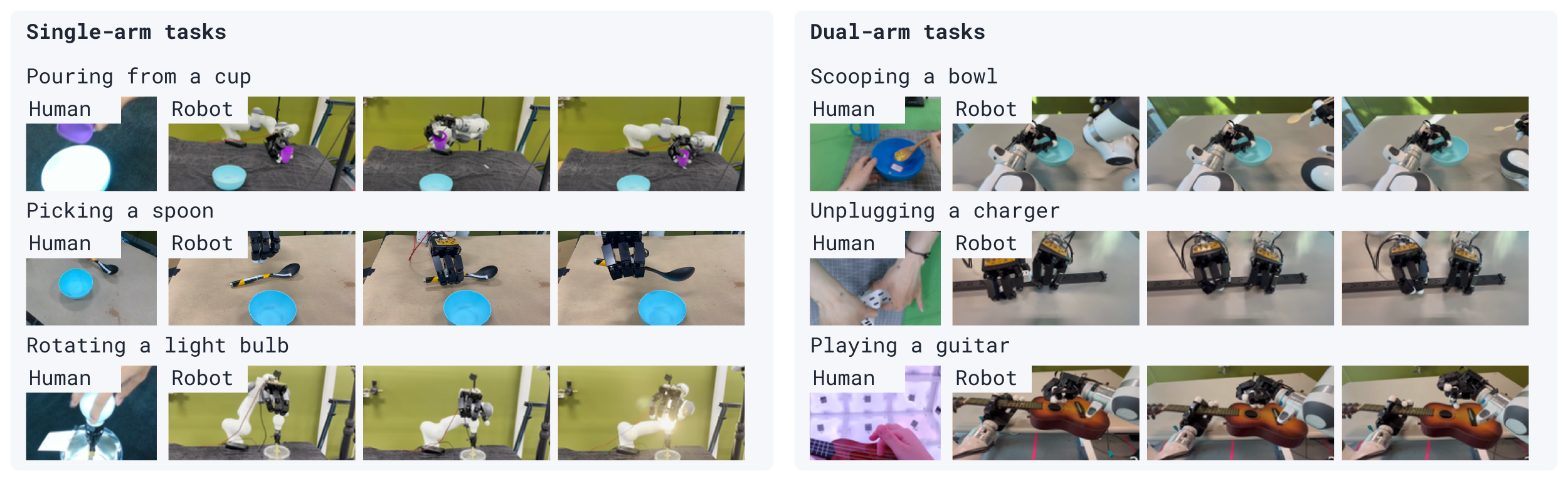}
    \caption{\textbf{Open-loop rollout with robustification.}
        Left: showing single-arm tasks including pouring from a cup, picking a spoon, and rotating a light bulb.
        Right: showing dual-arm tasks including scooping a bowl, unplugging a charger, and playing a guitar.
        The retargeted trajectories are executed directly on the physical robot.}
    \label{fig:deployment_results}
\end{figure*}

\begin{figure*}[tb]
    \centering
    \includegraphics[width=\textwidth]{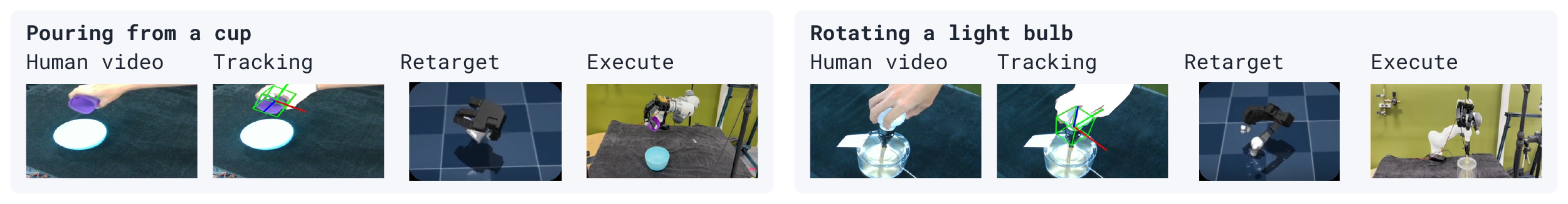}
    \caption{\textbf{Retargeting from a Single RGB Camera.} Starting from a single RGB observation of a user manipulating an object, we first reconstruct the 3D scene with hand and object meshes, which often contain reconstruction artifacts and penetrations. Our method then retargets this noisy data into a physically plausible robot trajectory with corrected contacts. Finally, the retargeted trajectory is successfully executed on the physical robot, demonstrating the pipeline's robustness to real-world data quality issues.}
    \label{fig:real2sim2real}
\end{figure*}

\graybold{Replay Generated Motion on Physical Robots.}
To demonstrate the dynamical feasibility of our retargeted trajectories, we deploy \spider on a physical system consisting of a 7-DoF \panda with an \allegro hand, using the robustification method in~\Cref{subsec:data_augmentation}.
The arm is controlled using operational space control (OSC), while the hand directly executes the optimized joint actions from our method.
We evaluate trajectory feasibility on four dexterous manipulation tasks—rotating a light bulb, manipulating a small spoon, playing a guitar, and unplugging a charger—each demanding precise finger coordination and showcasing the practical applicability of our retargeting approach.
Representative execution sequences are shown in~\Cref{fig:deployment_results}.
We note an inherent limitation of this open-loop execution: while the robustification and randomization in~\Cref{subsec:data_augmentation} make the trajectory tolerant to moderate perturbations and model mismatch, they cannot compensate for large deviations in the object's initial position.
If the object is placed significantly off from the reference initial pose, the open-loop rollout has no feedback mechanism to recover, and the execution fails; closing the loop with a learned tracking policy (e.g.,~\Cref{subsec:rl_policy_training_details}) is the natural remedy.

\graybold{Using In-the-Wild Data.}
\phantomsection\label{par:in_the_wild}
To demonstrate the flexibility of \spider for upstreaming data, we use the pipeline to convert video from a single RGB camera into executable robot trajectories.
The pipeline consists of: (1) 3D mesh reconstruction with Trellis~\citep{xiangStructured3DLatents2025}, (2) hand pose estimation with HAMER~\citep{pavlakosReconstructingHands3D2023} to obtain MANO parameters, and (3) object pose tracking with FoundationPose~\citep{wenFoundationPoseUnified6D2024} for 6D trajectories.
Due to occlusion and limited resolution, the human and object motions are noisier than those in the dataset, which requires more robust physics grounding.
As shown in~\Cref{fig:real2sim2real}, we validate on single-hand manipulation tasks like pouring with a cup and installing a bolt.
Real-world noise, artifacts, and penetrations are corrected through physics grounding in \spider, and the retargeted trajectories are directly executable on physical robots.
One limitation is that, since it is an open-loop rollout, it can only handle a narrow range of initial states and disturbances.
We leave the closed-loop control for future work.

\subsection{Large Scale Data Generation}
\label{subsec:large_scale_results}

\begin{figure}[t]
    \centering
    \includegraphics[width=1.0\linewidth]{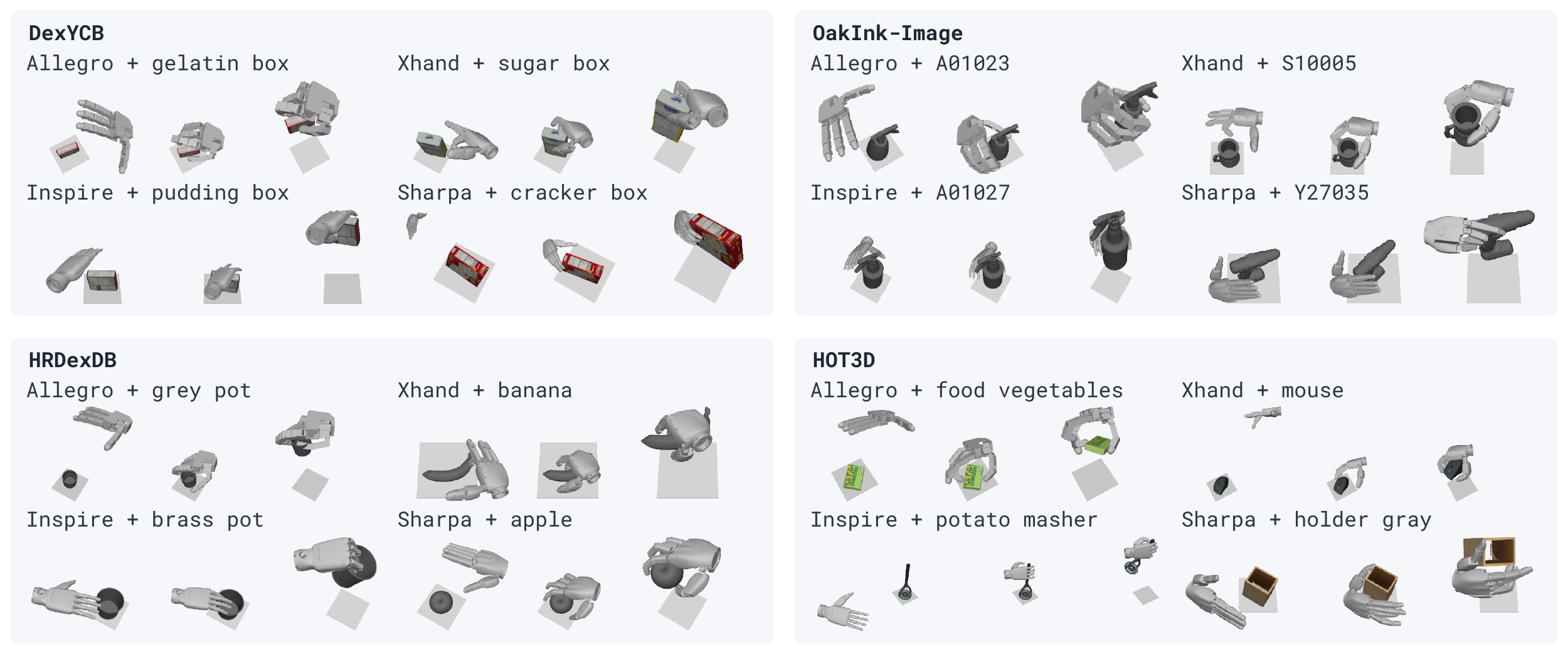}
    \caption{\textbf{Dataset trajectory visualization.} We apply \spider to human demonstrations and retain 6,520 simulated robot trajectories spanning 191 dataset-specific object identities. }
    \label{fig:large_scale}
\end{figure}

\begin{table*}[t]
    \centering
    \caption{\textbf{Large-scale dataset generation with \spider.}
        We retarget human demonstrations to four robotic hands.
        ADD-AUC (\%, higher is better) averages all completed seed-0 rollouts, including failed grasps. Bold shaded cells identify the best hand for each dataset; the aggregated row weights trajectories equally. }
    \label{tab:large_scale_generation}
    \begin{NiceTabular}{lrr|rrrr}
        \CodeBefore
        \rectanglecolor{metabg}{3-4}{3-4}
        \rectanglecolor{metabg}{4-4}{4-4}
        \rectanglecolor{metabg}{5-5}{5-5}
        \rectanglecolor{metabg}{6-4}{6-4}
        \rectanglecolor{metabg}{7-4}{7-4}
        \Body
        \toprule
        & \multicolumn{2}{c|}{\textbf{Human Data}} & \multicolumn{4}{c}{\textbf{ADD-AUC (\%) $\uparrow$}} \\
        \textbf{Dataset} & \textbf{\# Demos} & \textbf{\# Objects} & \textbf{\allegro} & \textbf{\xhand} & \textbf{\inspire} & \textbf{\sharpa} \\
        \midrule
        \dexycb~\numcite{chaoDexYCB2021} & 499 & 20 & \textbf{55.70} & 53.21 & 54.47 & 45.53 \\
        \oakinkimage~\numcite{yangOakInk2022} & 169 & 66 & \textbf{59.17} & 58.07 & 59.14 & 56.94 \\
        \hrdexdb~\numcite{limHRDexDB2026} & 441 & 94 & 64.21 & \textbf{64.44} & 63.76 & 62.14 \\
        \hotthree~\numcite{banerjeeHOT3DHandObject} & 2,718 & 33 & \textbf{71.16} & 70.09 & 69.41 & 40.74 \\
        \midrule
        \textbf{Aggregated} & \textbf{3,827} & \textbf{213} & \textbf{67.86} & 66.76 & 66.41 & 44.46 \\
        \bottomrule
    \end{NiceTabular}
\end{table*}

\graybold{Source Data Processing.}
We aggregate 3,827 right-hand demonstrations from \dexycb~\citep{chaoDexYCB2021}, \oakinkimage~\citep{yangOakInk2022}, \hrdexdb~\citep{limHRDexDB2026}, and \hotthree~\citep{banerjeeHOT3DHandObject}, covering 213 dataset-specific object identities (\Cref{tab:large_scale_generation}).
The key processing step is to identify hand--object interaction intervals and select the contact-establishment segments for retargeting, retaining a short approach to initialize the hand before contact.

\graybold{Generation Method.}
We apply \spider without contact guidance to four hands: \allegro, \xhand, \inspire, and \sharpa, since noisy demonstrations can make contact signals unreliable.
We use shared optimization hyperparameters across datasets and hands, with dataset-specific reference processing described in~\Cref{subsec:large_scale_details}.

\graybold{Generation Scale and Quality.}
With the shared optimization recipe, the completed rollouts achieve $61.48\%$ ADD-AUC, evaluated using the same surface-point metric as~\Cref{subsec:dexterous_retargeting_results}.
\allegro achieves the highest aggregated ADD-AUC, while \xhand performs best on \hrdexdb; performance varies across hands and datasets.
Example trajectories can be found in~\Cref{fig:large_scale}.

\subsection{Limitations and Common Failure Cases}
\label{subsec:failure_cases}

\spider has several failure modes.
(1)~\textbf{Perception errors}: inaccurate upstream reconstruction (\algname{HAMER}~\citep{pavlakosReconstructingHands3D2023}, \algname{FoundationPose}~\citep{wenFoundationPoseUnified6D2024}, \algname{Trellis}~\citep{xiangStructured3DLatents2025}) can introduce erroneous reference trajectories; large misalignments may cause the optimizer to converge to a semantically incorrect contact mode despite physical consistency.
(2)~\textbf{Severe occlusion}: heavy occlusion produces large temporal gaps in the reference, degrading retargeting quality; temporal consistency filtering is a promising mitigation.
(3)~\textbf{Open-loop sensitivity}: the generated trajectory is valid only near the aligned initial state---any initial pose mismatch leads to growing execution errors; closed-loop policy distillation is left as future work.
(4)~\textbf{Contact mode ambiguity}: without contact guidance, sampling may converge to an alternative valid grasp; \algname{SPIDER+C} resolves most such cases at the cost of contact annotations.

\section{Conclusion and Future Work}

This paper introduces \spider, a general, flexible and efficient physics-based retargeting framework that enables large-scale data transfer from human data.
Our framework achieves competitive performance compared to state-of-the-art methods while being an order of magnitude faster.
Despite its generality, \spider's performance depends on the quality of reconstructed 3D human-object interaction data; noisy meshes and motion can yield degraded trajectories.
Future work includes improving the robustness of the framework to noisy data and distilling the framework output into a policy for closed-loop control.

\clearpage
\newpage
\bibliographystyle{plainnat}
\bibliography{refs}

\clearpage
\newpage
\beginappendix

\section{Implementation Details}
\label{sec:implementation_details}

\subsection{Experimental Setup}
\label{subsec:experimental_setup}

\graybold{Dataset Selection.}
For dexterous manipulation tasks, we evaluate on 3 bimanual manipulation datasets: \gigahand \citep{fuGigaHandsMassiveAnnotated2024}, \oakink \citep{zhanOAKINK2DatasetBimanual2024} and \arctic \citep{li2023object}.
For humanoid tasks, we choose the commonly used \lafan~\citep{harvey2020robust} and \amass~\citep{mahmoodAMASSArchiveMotion2019} datasets for locomotion and the \omomo~\citep{li2023object} dataset for human-object interaction.
To the best of our knowledge, \textbf{\spider is the first universal retargeting method that can handle both dexterous manipulation and humanoid whole-body control tasks at this scale.}

\begin{figure}[htbp]
    \centering
    \includegraphics[width=\textwidth]{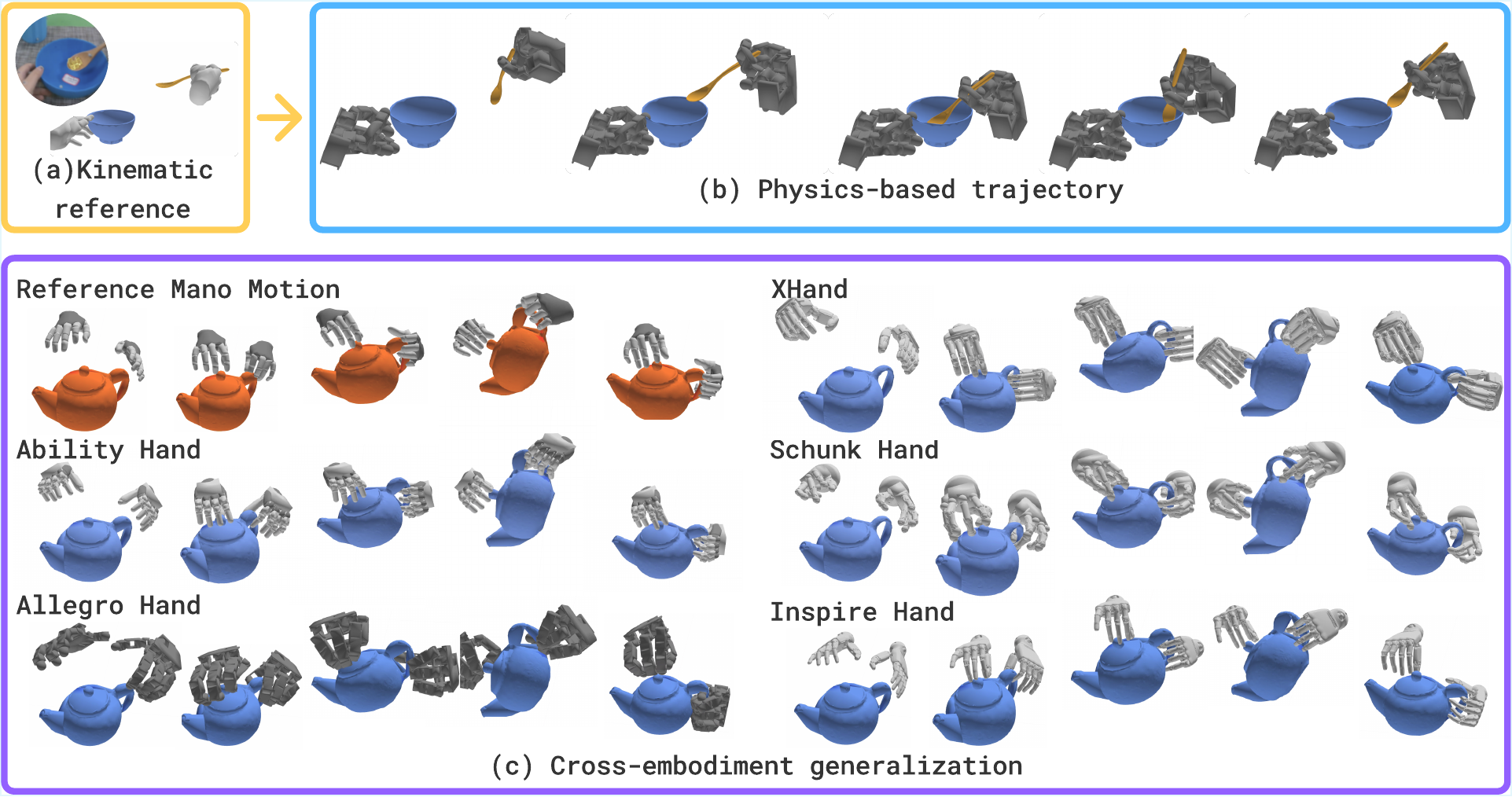}
    \caption{\textbf{Retargeted dexterous manipulation trajectories on different hands.} Starting with an infeasible motion with penetration and infeasible contact, \spider grounds it with physics. Different hands exhibit different behaviors with the same sampling parameters.}
    \label{fig:cross_embodiment_appendix}
\end{figure}

\graybold{Robot Embodiments.}
For dexterous manipulation, we evaluate across 5 different robotic hands to demonstrate the generalizability of our method: \allegro, \xhand, \inspire, \ability, and \schunk.
On humanoid whole-body control tasks, we use \unitreeg, \unitreeh, \fourier, and \booster as the target embodiments.
These platforms exhibit significant variations in degrees of freedom, dimensions, and finger configurations (see~\Cref{tab:hand_specifications}), showcasing our method's cross-embodiment capabilities.
\Cref{fig:cross_embodiment_appendix} shows some example trajectories generated by \spider on different hands.

\begin{figure*}[htbp]
    \centering
    \includegraphics[width=\textwidth]{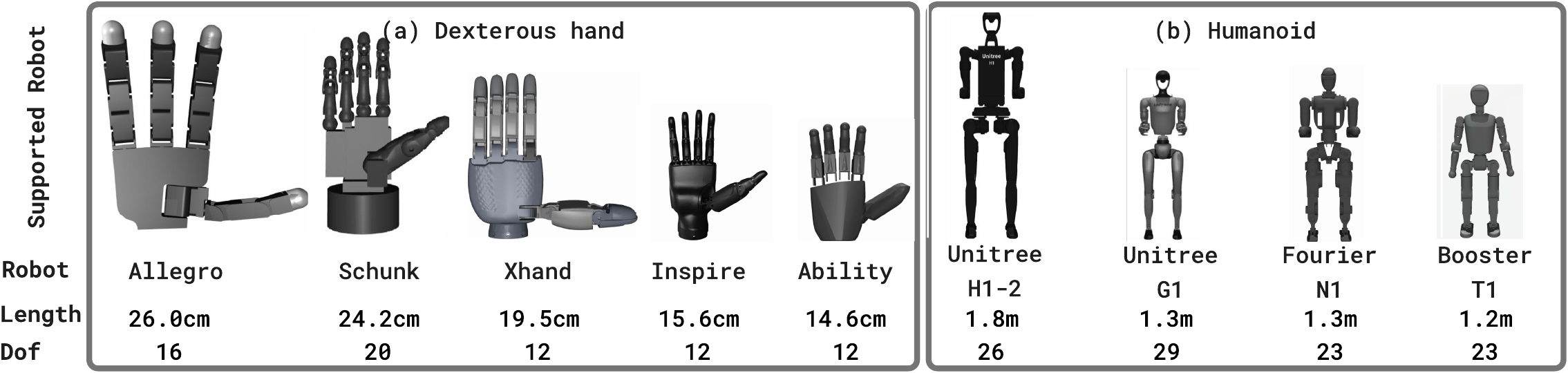}
    \caption{\textbf{Specifications of robots used in evaluation.}
        \spider supports both dexterous hands and humanoid robots.
        The significant variations in DoF, dimensions, and finger count demonstrate the cross-embodiment generalizability of our approach.
        We employ a simulated 12-DoF configuration~\citep{liManipTransEfficientDexterous2025} of the \inspire and \ability hands, removing the joint constraints compared to their real-world versions.}
    \label{fig:hand_specifications}
\end{figure*}

\begin{table}[htbp]
    \centering
    \begin{tabular}{l>{\centering}p{0.8cm}>{\centering}p{1.2cm}>{\centering}p{1.3cm}>{\centering\arraybackslash}p{1.0cm}}
        \toprule
        \textbf{Hand} & \textbf{DOF} & \textbf{Width (cm)} & \textbf{Length (cm)} & \textbf{Fingers} \\
        \midrule
        Allegro       & $16$         & $12.0$              & $26.0$               & $4$              \\
        XHAND         & $12$         & $9.2$               & $19.5$               & $5$              \\
        Inspire       & $12$         & $8.8$               & $15.6$               & $5$              \\
        Ability       & $12$         & $7.5$               & $14.6$               & $5$              \\
        Schunk SVH    & $20$         & $9.2$               & $24.2$               & $5$              \\
        \bottomrule
    \end{tabular}
    \caption{\textbf{Specifications of robotic hands used in evaluation.} The significant variations in DoF, dimensions, and finger count demonstrate the cross-embodiment generalizability of our approach. We employ a simulated 12-DoF configuration~\citep{liManipTransEfficientDexterous2025} of the Inspire and Ability hands, removing the joint constraints compared to their real-world versions.}
    \label{tab:hand_specifications}
\end{table}

\graybold{Evaluation Metrics.}
In our own ablation study where we validate the effect of our annealing and contact guidance design, we report tracking performance on the robot and object motion.
For object tracking error, we compute per-step averaged rotation error $E_\text{rot} = \frac{1}{T} \sum_{t=1}^T \arccos \left( 2 \langle q_{{\text{obj}},t}, q_{{\text{obj}},t}^{\text{ref}} \rangle^2 - 1 \right)$ and position error $E_\text{pos} = \frac{1}{T} \sum_{t=1}^T \|p_{{\text{obj}},t} - p_{{\text{obj}},t}^{\text{ref}}\|_2$, where $T$ is the number of timesteps, $\langle\cdot,\cdot\rangle$ is the dot product, and $q_{{\text{obj}},t}$ and $p_{{\text{obj}},t}$ denote the object quaternion and position at timestep $t$, respectively.
We define success only on the object motion since the goal is to move the object rather than to precisely track the robot hand.
Task success is defined using: (1) the rotation error $E_\text{rot} < 0.5$ rad, and (2) the average per-step translation error $E_\text{pos} < 0.1$ m.
Since a single success threshold is sensitive to its choice, we additionally report ADD-AUC10, following the ADD-AUC protocol~\citep{mandiDexMachinaFunctionalRetargeting2025}: we sample $512$ keypoints on the object mesh surface, transform them by both the achieved and the reference object poses at every frame, and compute the mean per-keypoint L2 distance (the classic ADD metric for rigid bodies).
The metric (denoted ADD-AUC, or ADD-AUC10 to make the $10$\,cm range explicit) is the area under the per-frame success-rate curve as the distance threshold sweeps from $0$ to $10$\,cm, with the threshold axis normalized so the metric lies in $[0, 100]\%$ (higher is better).
This threshold-free metric summarizes tracking quality without committing to a single cutoff and is sensitive to both translational and rotational errors through the mesh keypoints.
For humanoid locomotion, the joint tracking error, pelvis position error and pelvis orientation error are computed.

When comparing with baselines, we use the same evaluation metrics in the original papers to make a fair comparison.
Specifically, we follow the same evaluation setting as \maniptrans~\citep{liManipTransEfficientDexterous2025} and \dexmachina~\citep{mandiDexMachinaFunctionalRetargeting2025} with their published evaluation protocol.

\subsection{Data Preprocessing}
We extract a 21D keypoint representation per hand using 3D fingertip positions and 6D wrist pose from MANO. An IK solver maps these keypoints to robot-specific joint positions by minimizing
$\mathcal{L}_{\text{IK}} = \sum_{i=1}^{n} \|\mathbf{p}_{i}^{\text{robot}} - \mathbf{p}_{i}^{\text{human}}\|^2 + 0.1 \|\mathbf{R}_{\text{wrist}}^{\text{robot}} - \mathbf{R}_{\text{wrist}}^{\text{human}}\|_F^2$.
Trajectories are resampled at 50 Hz and low-pass filtered at 10 Hz.

\subsection{Hyperparameters and Setup}
The settings below describe the retargeting comparisons; the large-scale generation campaign uses the separate recipe in~\Cref{subsec:large_scale_details}.
Simulations for these comparisons run at 100 Hz physics and 50 Hz control in MuJoCo Warp (most datasets) and Genesis (\arctic dataset, as in DexMachina~\citep{mandiDexMachinaFunctionalRetargeting2025}). Ablations and speed tests use RTX 4090 GPUs; the large-scale generation campaign uses L40S GPUs.
All hyperparameters are listed in \Cref{tab:spider-mjwp-params}.

\graybold{Automatic difficulty boosting. }
In practice, \spider requires little per-task hyperparameter tuning because per-step failures are handled by an automatic boosting mechanism.
When the optimizer fails to reach the error tolerance at a step, the system retries from the last checkpoint with a boosted budget: each boost level doubles the number of optimization iterations and scales the sampling noise by $\sqrt{2}$ (both relative to the original configuration), and the noise-decay rate is re-derived so that the final-iteration noise still matches the configured target.
If local retries are exhausted, the system reverts to an earlier checkpoint and re-optimizes that segment with the boosted budget before continuing.
This acts as an automatic per-step hyperparameter adaptation: easy segments are solved cheaply with the default budget, while hard segments (e.g., contact-rich transitions) automatically receive more samples and wider exploration, so the iteration count and noise scale do not need to be tuned for the hardest step in advance.

\begin{table*}[htbp]
    \centering
    \caption{\spider hyperparameters for \oakink and \gigahand retargeting with MuJoCo Warp Simulator. Values that differ from the default are shown per dataset.}
    \label{tab:spider-mjwp-params}
    \begin{tabular}{lcccc}
        \toprule
        \textbf{Parameter}                               & \textbf{\oakink} & \textbf{\oakink (w/ guidance)} & \textbf{\gigahand} & \textbf{\gigahand (w/ guidance)} \\
        \midrule
        \multicolumn{5}{l}{\textit{Simulation}}                                                                                                                      \\
        Sim.\ timestep $\Delta t_{\text{sim}}$           & $0.01$ s         & $0.005$ s                      & $0.01$ s           & $0.005$ s                        \\
        Control timestep $\Delta t_{\text{ctrl}}$        & $0.4$ s          & $0.05$ s                       & $0.4$ s            & $0.2$ s                          \\
        Knot spacing $\Delta t_{\text{knot}}$            & $0.4$ s          & $0.1$ s                        & $0.4$ s            & $0.4$ s                          \\
        \midrule
        \multicolumn{5}{l}{\textit{MPC / CEM Optimization}}                                                                                                          \\
        Planning horizon $H$                             & $2.0$ s          & $0.4$ s                        & $1.6$ s            & $1.6$ s                          \\
        Horizon steps                                    & 200              & 80                             & 160                & 320                              \\
        Parallel samples $N_w$                           & 1024             & 1024                           & 1024               & 1024                             \\
        Max CEM iterations $N$                           & 32               & 32                             & 32                 & 32                               \\
        CEM temperature $\tau$                           & 1.0              & 1.0                            & 1.0                & 1.0                              \\
        Early stop threshold $\epsilon$                  & $0.01$           & $0.0$                          & $0.01$             & $0.0$                            \\
        Early stop patience                              & 2                & 1                              & 1                  & 1                                \\
        \midrule
        \multicolumn{5}{l}{\textit{Noise Schedule}}                                                                                                                  \\
        Initial noise $\sigma_{\text{first}}$            & $0.5$            & $1.0$                          & $0.5$              & $1.0$                            \\
        Final noise (horizon end) $\sigma_{\text{last}}$ & 1.0              & 1.0                            & 1.0                & 1.0                              \\
        Noise decay $\beta$                              & $0.931$          & $0.963$                        & $0.931$            & $0.963$                          \\
        Joint noise scale                                & $0.03$           & $0.02$                         & $0.03$             & $0.15$                           \\
        Wrist pos noise scale                            & $0.01$           & $0.005$                        & $0.03$             & $0.01$                           \\
        Wrist rot noise scale                            & 0.01             & 0.01                           & 0.01               & 0.01                             \\
        \midrule
        \multicolumn{5}{l}{\textit{Reward Weights}}                                                                                                                  \\
        Base pos $w_{\text{pos}}^{\text{base}}$          & $0.1$            & $0.3$                          & $0.1$              & $0.03$                           \\
        Base rot $w_{\text{rot}}^{\text{base}}$          & $0.03$           & $0.3$                          & $0.03$             & $0.03$                           \\
        Fingertip pos $w_{\text{pos}}$                   & 1.0              & 1.0                            & 1.0                & 1.0                              \\
        Fingertip rot $w_{\text{rot}}$                   & $0.3$            & $1.0$                          & $0.3$              & $1.0$                            \\
        \midrule
        \multicolumn{5}{l}{\textit{Contact Guidance (w/ guidance only)}}                                                                                             \\
        Contact guidance                                 & \xmark           & \cmark                         & \xmark             & \cmark                           \\
        Guidance decay $\gamma_g$                        & --               & $0.8$                          & --                 & $0.5$                            \\
        Pos actuator gain                                & --               & $10.0$                         & --                 & $2.0$                            \\
        Rot actuator gain                                & --               & $0.3$                          & --                 & $0.3$                            \\
        \bottomrule
    \end{tabular}
\end{table*}

\subsection{Large-Scale Dataset Generation}
\label{subsec:large_scale_details}

\graybold{Source Data and Clip Selection.}
The central processing objective is to isolate the hand--object interaction interval, particularly contact establishment, and retarget the motion that forms and maintains the grasp.
We preserve a short pre-contact approach for initialization; contact-focused selection therefore includes the transition into contact rather than only frames in which contact is already present.
The campaign in~\Cref{subsec:large_scale_results} uses 499 usable right-hand \dexycb~\citep{chaoDexYCB2021} demonstrations (out of 501 source folders), 169 \oakinkimage~\citep{yangOakInk2022} sequences with the \texttt{liftup} intent, 441 \hrdexdb~\citep{limHRDexDB2026} sequences, and 2,718 single-hand \hotthree clips.
\oakinkimage is distinct from the bimanual \taskname{OAKINK2} data used in the retargeting comparisons.
\hotthree~\citep{banerjeeHOT3DHandObject} clips are extracted from 424 recordings (198 Aria and 226 Quest3).
Hand and object timestamps are matched within 20\,ms, and Quest3 trajectories are rotated from Y-up to Z-up.
A grasp event requires a fingertip within 1.5\,cm of the object surface for at least five frames; lift onset requires object displacement above 2\,mm per frame during contact.
Clips begin 30 frames before first touch and end 10 frames after lift onset.
Static grasps and clips with the carrying hand untracked in more than $20\%$ of frames are discarded.
The 141 additional right-hand \hotthree bimanual clips and all left-hand clips are excluded from the reported results.
Object counts are dataset-local identifiers, not globally deduplicated physical objects or semantic classes.

\graybold{Representation and Lift Augmentation.}
Each source episode contains object poses, 21 hand landmarks in MediaPipe order, MANO vertices, and an object mesh, in metres at 30\,Hz.
GLB scene transforms are preserved, including \hotthree's millimetre-to-metre scale, and the initial object is translated so its lowest mesh vertex lies on the ground.
The wrist orientation is constructed from the wrist-to-middle-MCP direction and the index-to-ring-MCP direction; the five fingertip targets use landmark indices 4, 8, 12, 16, and 20.
\dexycb and \oakinkimage packages include an appended synthetic lift; \hotthree packages are extended to a target height, whereas \hrdexdb contains recorded lifts.
The common processor further appends a lift if needed to reach 0.30\,m above the initial object position, translating the hand and object together at 0.3\,m/s while holding their relative configuration fixed.
References are resampled to 12.5\,Hz with duration-preserving interpolation and a nominal 12--64-frame range before IK.
\hrdexdb instead uses a fixed 24-frame recipe, yielding 20 saved reference frames after IK trimming.
Trimming and resampling can reduce the final reference lift below 0.30\,m.
No hand-size normalization or fingertip tightening is applied.

\graybold{Embodiments and Initialization.}
The four floating hands have six base coordinates and 16 (\allegro), 12 (\xhand), 12 (\inspire), or 22 (\sharpa) finger control dimensions; \allegro has four fingers and the others have five.
We generate convex object collision geometry, detect reference contacts, and construct a MuJoCo scene for each hand.
Open-hand IK transitions each finger from an open pose to its reference target at its first detected contact frame.
An IK rollout is rejected if it contains non-finite states, has fewer than eight frames, or places the object outside the vertical range $[-0.3,2.0]$\,m.
The hands are simulated without attached robot arms.

\graybold{MPC Configuration.}
All reported runs use seed~0 and the same optimization settings in~\Cref{tab:large_scale_parameters}, without contact guidance or external force/torque perturbations.
The campaign uses NVIDIA L40S GPUs with concurrent optimization jobs; the archived runs report a median task wall time of approximately 220\,s, including preprocessing.
This runtime measures generation cost, not the duration of the resulting trajectory.

\begin{table}[htbp]
    \centering
    \caption{\textbf{Shared large-scale generation configuration.} \hrdexdb differs only in its reference-length processing.}
    \label{tab:large_scale_parameters}
    \begin{tabular}{lr|lr}
        \toprule
        \textbf{Parameter} & \textbf{Value} & \textbf{Parameter} & \textbf{Value} \\
        \midrule
        Simulation timestep & 0.01\,s & Reference timestep & 0.08\,s \\
        Control interval & 0.25\,s & Knot spacing & 0.5\,s \\
        Planning horizon & 2.0\,s & Samples per iteration & 1,024 \\
        Maximum iterations & 16 & Temperature & 1.0 \\
        Initial control-noise scale & 2.0 & Horizon-end noise scale & 4.0 \\
        Joint noise scale & 0.1 & Position / rotation noise scale & 0.01 / 0.01 \\
        Final optimization noise scale & 0.01 & Contact reward weight & 0 \\
        \bottomrule
    \end{tabular}
\end{table}

\graybold{Tracking Evaluation.}
We recompute surface-point ADD-AUC directly from saved object poses and source meshes, following the metric in~\Cref{sec:implementation_details}.
For each sequence, we sample $K=512$ points $v_j$ uniformly by surface area on the object mesh and use the same points for all robot hands.
The per-frame error is
\begin{equation}
    d_t = \frac{1}{K}\sum_{j=1}^{K}
    \left\|R_t v_j + p_t - \left(R_t^{\mathrm{ref}}v_j+p_t^{\mathrm{ref}}\right)\right\|_2.
\end{equation}
We evaluate the fraction of frames with $d_t\leq\tau$ at 100 equally spaced thresholds $\tau\in[0,0.10]$\,m and apply trapezoidal integration after normalizing the threshold axis to $[0,1]$.
Reported values are multiplied by 100.
Surface samples use a deterministic per-sequence seed.
For temporal alignment, we take every eighth state from the 100\,Hz simulation and truncate simulation and reference to their common length at 12.5\,Hz, without mean-centering or extending the reference.
Each trajectory contributes equally to dataset and overall averages.
The main table includes all 15,041 completed single-hand rollouts, including poor reference lifts and failed grasps.
The 281 rollouts containing non-finite simulated states are retained; invalid evaluated frames contribute zero threshold successes.
A position-only AUC implementation, which ignores rotation, gives $63.25\%$ overall; the reported surface-point ADD-AUC is $61.48\%$.

\graybold{Completion and Retention Criteria.}
The 3,827 source demonstrations and four hands define 15,308 planned tasks.
There are 15,302 task records: 15,041 completed rollouts, 161 IK-sanity failures, and 100 preparation failures; six further planned tasks have no record.
Failures without a completed trajectory have no ADD-AUC and are excluded from its denominator.
For dataset retention, we require a reference rise of at least 0.15\,m, a final simulated rise of at least 0.10\,m, and a final object-position error of at most 0.10\,m.
For this final-error test, reference positions are linearly interpolated onto the simulation timestamps, with endpoint holding beyond the reference duration.
These labels are recomputed from the saved trajectories.
Excluding the 437 completed rollouts with insufficient reference lift leaves 14,604 eligible rollouts with ADD-AUC $61.67\%$; applying the full retention filter yields the 6,520 trajectories in~\Cref{tab:large_scale_retained}.
Their higher ADD-AUC describes the selected dataset rather than unfiltered generation performance.
No perturbation robustness or real-robot execution is evaluated for this campaign.

\begin{table}[htbp]
    \centering
    \caption{\textbf{Retained simulated demonstrations, seed~0.} Object identities are counted within each source dataset. Hours equal saved frames divided by 100\,Hz and 3,600.}
    \label{tab:large_scale_retained}
    \begin{tabular}{lrrrrr}
        \toprule
        \textbf{Dataset} & \textbf{Trajectories} & \textbf{Objects} & \textbf{Frames} & \textbf{Hours} & \textbf{ADD-AUC (\%)} \\
        \midrule
        \dexycb~\numcite{chaoDexYCB2021}       & 870   & 20  & 151,650   & 0.42 & 73.27 \\
        \oakinkimage~\numcite{yangOakInk2022} & 241   & 58  & 36,850    & 0.10 & 76.52 \\
        \hrdexdb~\numcite{limHRDexDB2026}      & 871   & 80  & 152,425   & 0.42 & 79.13 \\
        \hotthree~\numcite{banerjeeHOT3DHandObject}        & 4,538 & 33  & 950,650   & 2.64 & 79.19 \\
        \midrule
        \textbf{Total / overall} & \textbf{6,520} & \textbf{191} & \textbf{1,291,575} & \textbf{3.59} & \textbf{78.30} \\
        \bottomrule
    \end{tabular}
\end{table}

\subsection{Humanoid Policy Training}
\label{subsec:rl_policy_training_details}

This section describes the training details for the humanoid loco-manipulation policy experiment in the main paper.
We use the PPO algorithm~\cite{schulman2017proximalpolicyoptimizationalgorithms} and port the implementation from~\cite{weng2025hdmilearninginteractivehumanoid}.

\graybold{Rewards. }
We use the following rewards:
\begin{itemize}
    \item Body tracking reward: tracking humanoid pelvis, hand, and leg motions.
    \item Object tracking reward: tracking the object motion.
    \item Action rate penalty: penalizing the action rate to avoid jittering.
\end{itemize}
We remove the contact reward, self-collision penalty as well as the auxiliary contact reward proposed in the paper to evaluate the reference motion retargeting quality.

\graybold{Termination. }
We terminate the training when the average body tracking error is greater than 10 cm and the average object tracking error is greater than 10 cm.

\subsection{Dexterous Hand Policy Training}
\label{subsec:grasp_policy_training_details}

This section describes the training details for the dexterous grasping policy experiment in the main paper.
\spider is complementary to RL algorithms: it provides an already-feasible reference motion, so the policy only needs to track it rather than discover the behavior from scratch.
The task uses a right arm ($7$-DoF) paired with a $22$-DoF dexterous hand ($5$ fingers, $29$ DoF total) grasping a cup on a table.
The reference is collected from an in-the-wild human demonstration and physics-grounded with \spider (\Cref{par:in_the_wild}).
The \spider nominal feedforward control is used both to initialize the policy targets and as the reference for the tracking reward; we do \emph{not} use it as a residual control term, though such a variant is also possible.

\graybold{Scene and simulation. }
The scene consists of the right arm mounted at a fixed base, the object (mesh loaded from the demonstration manifest) as a dynamic rigid body, and a table with its top surface at $0.75$\,m.
The environment runs at $20$\,Hz control over a $120$\,Hz PhysX simulation (decimation $6$; TGS solver with $16$ position iterations, $2$\,mm contact offset), with episodes of $12$\,s ($\approx 240$ policy steps) and $4096$ parallel environments.
The arm uses per-joint PD gains identified from the physical system.

\graybold{Action space ($28$-D). }
The policy outputs a $28$-D vector in $[-1,1]$ that is scaled by per-step limits and accumulated onto persistent wrist and finger targets (delta control).
The $6$-D wrist delta (position $\pm 1$\,cm per step, rotation $\pm 4.5^\circ$ per step as axis-angle) drives a nullspace differential-IK controller with damped least squares; the $22$-D finger delta ($\pm 0.2$\,rad per step) is sent to a joint-position controller clamped to joint limits.

\graybold{Observation space ($325$-D). }
The observation concatenates: relative joint positions and velocities ($29+29$), wrist pose ($3+6$, using $6$-D rotation), accumulated wrist and finger targets ($3+6+22$), object pose ($3+6$), the reference wrist pose, finger joints, and object pose at the current reference step ($3+6+22+3+6$), the goal object bounding box ($24$) and goal hand landmarks in the object frame ($51$), the current object bounding box ($24$) and hand landmarks ($51$), and the last action ($28$).
We use an asymmetric actor-critic: the critic additionally receives privileged information (per-joint PD gains, effort limits and armature, object mass and friction, object velocities, and the object--wrist relative pose).
The policy network receives the current frame concatenated with a $4$-step history buffer; actor and critic use $(512, 256, 128)$ MLPs.

\graybold{Rewards. }
We use dense tracking rewards of the form $\exp(-s \cdot e)$, where $e$ is the tracking error and $s$ is a per-term scale, listed in \Cref{tab:grasp-rewards}, together with a small action-magnitude penalty.
The hand-landmark term tracks $17$ landmarks on the hand expressed in the object frame, which couples the hand and object motion and is given the largest weight.
We remove the lifting reward and the success bonus, keeping only the imitation tracking terms so that performance reflects the quality of the reference motion, mirroring the humanoid setting.

\begin{table}[h]
    \centering
    \begin{tabular}{lccc}
        \toprule
        \textbf{Term} & \textbf{Error $e$} & \textbf{Scale $s$} & \textbf{Weight} \\
        \midrule
        Object position  & $\|\Delta p_{\text{obj}}\|$        & $30$ & $0.065$ \\
        Object rotation  & geodesic angle                     & $1.9$ & $0.065$ \\
        Wrist position   & $\|\Delta p_{\text{wrist}}\|$      & $20$ & $0.039$ \\
        Wrist rotation   & geodesic angle                     & $1.9$ & $0.039$ \\
        Finger joints    & mean $|\Delta q|$                  & $10$ & $0.039$ \\
        Hand landmarks   & per-point distance (object frame)  & $60$ & $0.279$ \\
        Action penalty   & $\mathrm{mean}(a_t^2)$             & --   & $-0.10$ \\
        \bottomrule
    \end{tabular}
    \caption{\textbf{Reward terms for the dexterous grasping policy.} Each tracking term is $\exp(-s \cdot e) \in (0, 1]$; weights are normalized. The lifting reward and success bonuses of the original environment are removed.}
    \label{tab:grasp-rewards}
\end{table}

\graybold{Termination and success. }
An episode terminates if the object leaves the workspace (height below $0.5$\,m or above $1.5$\,m in the environment frame), if the wrist drifts more than $1.2$\,m from the object, or if any arm joint exceeds its position limit; episodes truncate at $12$\,s.
An episode is counted as successful when the object is lifted at least $4$\,cm above the table while the hand remains in contact (minimum hand-to-object distance below $2$\,cm).

\graybold{Initial-state generation via sequential IK rollout. }
To turn the single \spider-retargeted demonstration into a diverse initial-state distribution, we augment it into thousands of randomized, reachability-validated placements.
For each candidate placement, the object is randomly translated on the table and rotated in yaw (uniform over the table workspace, discarding placements outside the table bounds), and the demonstrated wrist trajectory---stored in an object-centered frame---is re-expressed in the robot frame at that placement.
A GPU-batched damped-least-squares IK solver then rolls the trajectory forward waypoint-by-waypoint: the first waypoint is warm-started from the nearest configuration in a precomputed table of healthy arm configurations (with a larger iteration budget of $400$ vs.\ $50$ for subsequent waypoints), and every later waypoint is warm-started from the previous solution.
A placement is accepted only if the arm can track the \emph{entire} trajectory within tolerance (wrist position error $<3$\,cm, orientation error $<0.15$\,rad) while keeping the proximal joints within $0.5$\,rad of the healthy-configuration manifold (rejecting contorted elbow postures).
Before the object is lifted, its pose follows the transformed reference; after lift-off, it is treated as rigidly attached to the wrist and its pose is recomputed from the achieved forward kinematics.
Accepted placements---each carrying a complete arm, finger, and object trajectory---are further filtered by a physics rollout that verifies grasp stability before being used for training.

\graybold{Reset. }
On reset, each environment samples a placement and a starting waypoint along its trajectory (biased toward early waypoints, with a fixed fraction starting from the beginning) and initializes the robot and object from the corresponding validated state; for early waypoints, the object position is additionally perturbed by $\pm 4$\,cm in the horizontal plane with $\pm 30^\circ$ yaw, and joint targets receive small noise.
Episode start times are desynchronized across environments so that resets do not coincide.

\graybold{PPO hyperparameters. }
We train with PPO~\citep{schulman2017proximalpolicyoptimizationalgorithms}: adaptive learning rate initialized at $3\times 10^{-4}$ (KL-targeted at $0.016$), discount $\gamma=0.95$, GAE $\lambda=0.9$, clip $0.2$, horizon length $16$, minibatch size $16384$, $5$ mini-epochs, gradient norm clipping at $1.0$, and observation/value normalization.
Each policy is trained on a single L40 GPU for roughly $3$--$4$ hours.

\section{Comparison with Baselines}
\label{sec:comparison_with_baselines}

\subsection{Comparison Against Kinematics-Based Retargeting Baselines}
\graybold{Setup and Per-Hand Results.}
To ensure representative evaluation while maintaining computational feasibility, we conduct our ablation study on a collection of challenging manipulation tasks from established datasets: (1) for \oakink \citep{zhanOAKINK2DatasetBimanual2024}, we evaluate on 7 distinct bimanual two-object manipulation tasks that were previously benchmarked in \maniptrans \citep{liManipTransEfficientDexterous2025}; and (2) for \gigahand \citep{fuGigaHandsMassiveAnnotated2024}, we utilize their released example dataset comprising 3 manipulation trajectories over five hands.
Per-hand tracking results are presented in~\Cref{tab:optimization_oakink,tab:optimization_gigahand}, where \spider wins in terms of object tracking precision, especially with contact guidance enabled.
The aggregated comparison is summarized in the main paper (\Cref{tab:dexterous_comparison_main}).
\begin{table}[h]
    \centering
    \resizebox{\columnwidth}{!}{%
        \begin{NiceTabular}{ll|ccc}
            \CodeBefore
            \rectanglecolor{metabg}{2-5}{4-5}
            \rectanglecolor{metabg}{5-4}{5-4}
            \rectanglecolor{metabg}{6-5}{7-5}
            \rectanglecolor{metabg}{8-4}{9-4}
            \rectanglecolor{metabg}{10-5}{10-5}
            \rectanglecolor{metabg}{11-5}{16-5}
            \Body
            \toprule
            \textbf{Robot}        &                          & \kinematic          & \spider                    & \algname{SPIDER+C}         \\
            \midrule
            \Block{3-1}{\allegro} & $e_t$ (cm) $\downarrow$  & $9.4_{\pm 6.8}$     & $7.7_{\pm 7.5}$            & $\mathbf{4.6_{\pm 2.8}}$   \\
                                  & $e_r$ (deg) $\downarrow$ & $44.2_{\pm 30.9}$   & $37.6_{\pm 33.1}$          & $\mathbf{19.9_{\pm 18.9}}$ \\
                                  & AUC $\uparrow$           & $21.5$              & $53.4$                     & $\mathbf{56.6}$            \\
            \Block{3-1}{\ability} & $e_t$ (cm) $\downarrow$  & $76.3_{\pm 143.5}$  & $\mathbf{9.3_{\pm 9.1}}$   & $10.9_{\pm 8.2}$           \\
                                  & $e_r$ (deg) $\downarrow$ & $75.2_{\pm 26.5}$   & $44.4_{\pm 39.1}$          & $\mathbf{36.4_{\pm 32.7}}$ \\
                                  & AUC $\uparrow$           & $8.2$               & $43.7$                     & $\mathbf{57.9}$            \\
            \Block{3-1}{\inspire} & $e_t$ (cm) $\downarrow$  & $196.6_{\pm 545.3}$ & $\mathbf{9.1_{\pm 8.1}}$   & $11.7_{\pm 11.7}$          \\
                                  & $e_r$ (deg) $\downarrow$ & $91.0_{\pm 31.2}$   & $\mathbf{41.0_{\pm 29.4}}$ & $43.3_{\pm 29.2}$          \\
                                  & AUC $\uparrow$           & $7.7$               & $46.4$                     & $\mathbf{63.4}$            \\
            \Block{3-1}{\schunk}  & $e_t$ (cm) $\downarrow$  & $17.8_{\pm 9.7}$    & $9.1_{\pm 8.7}$            & $\mathbf{7.6_{\pm 4.7}}$   \\
                                  & $e_r$ (deg) $\downarrow$ & $86.5_{\pm 24.7}$   & $33.1_{\pm 26.2}$          & $\mathbf{28.8_{\pm 19.6}}$ \\
                                  & AUC $\uparrow$           & $14.6$              & $46.7$                     & $\mathbf{52.4}$            \\
            \Block{3-1}{\xhand}   & $e_t$ (cm) $\downarrow$  & $39.5_{\pm 21.4}$   & $7.1_{\pm 7.6}$            & $\mathbf{5.2_{\pm 3.3}}$   \\
                                  & $e_r$ (deg) $\downarrow$ & $108.4_{\pm 15.6}$  & $32.4_{\pm 29.7}$          & $\mathbf{24.1_{\pm 19.6}}$ \\
                                  & AUC $\uparrow$           & $5.7$               & $53.4$                     & $\mathbf{54.5}$            \\
            \bottomrule
        \end{NiceTabular}}
    \caption{\textbf{Comparison with \kinematic retargeting on \oakink.} Mean$_{\pm\text{std}}$ over 5 seeds. AUC denotes ADD-AUC in percent (\Cref{sec:implementation_details}).}
    \label{tab:optimization_oakink}
\end{table}

\begin{table}[h]
    \centering
    \resizebox{\columnwidth}{!}{%
        \begin{NiceTabular}{ll|ccc}
            \CodeBefore
            \rectanglecolor{metabg}{2-4}{3-4}
            \rectanglecolor{metabg}{4-5}{4-5}
            \rectanglecolor{metabg}{5-4}{5-4}
            \rectanglecolor{metabg}{6-5}{6-5}
            \rectanglecolor{metabg}{7-4}{8-4}
            \rectanglecolor{metabg}{9-5}{10-5}
            \rectanglecolor{metabg}{11-4}{11-4}
            \rectanglecolor{metabg}{12-5}{12-5}
            \rectanglecolor{metabg}{13-4}{14-4}
            \rectanglecolor{metabg}{15-5}{16-5}
            \Body
            \toprule
            \textbf{Robot}        &                          & \kinematic        & \spider                   & \algname{SPIDER+C}        \\
            \midrule
            \Block{3-1}{\allegro} & $e_t$ (cm) $\downarrow$  & $23.9_{\pm 11.2}$ & $\mathbf{4.5_{\pm 1.0}}$  & $8.7_{\pm 3.8}$           \\
                                  & $e_r$ (deg) $\downarrow$ & $45.0_{\pm 22.5}$ & $\mathbf{9.1_{\pm 5.7}}$  & $9.9_{\pm 2.9}$           \\
                                  & AUC $\uparrow$           & $41.4$            & $58.7$                    & $\mathbf{83.1}$           \\
            \Block{3-1}{\ability} & $e_t$ (cm) $\downarrow$  & $37.3_{\pm 30.8}$ & $\mathbf{7.9_{\pm 5.5}}$  & $12.5_{\pm 4.8}$          \\
                                  & $e_r$ (deg) $\downarrow$ & $67.7_{\pm 44.2}$ & $19.7_{\pm 10.1}$         & $\mathbf{17.8_{\pm 6.3}}$ \\
                                  & AUC $\uparrow$           & $17.4$            & $\mathbf{55.5}$           & $30.8$                    \\
            \Block{3-1}{\inspire} & $e_t$ (cm) $\downarrow$  & $44.2_{\pm 27.3}$ & $\mathbf{4.5_{\pm 0.3}}$  & $11.4_{\pm 7.6}$          \\
                                  & $e_r$ (deg) $\downarrow$ & $71.3_{\pm 29.6}$ & $15.1_{\pm 12.3}$         & $\mathbf{14.8_{\pm 8.1}}$ \\
                                  & AUC $\uparrow$           & $12.5$            & $53.1$                    & $\mathbf{73.4}$           \\
            \Block{3-1}{\schunk}  & $e_t$ (cm) $\downarrow$  & $38.3_{\pm 17.5}$ & $\mathbf{11.1_{\pm 9.4}}$ & $14.0_{\pm 2.5}$          \\
                                  & $e_r$ (deg) $\downarrow$ & $59.2_{\pm 16.4}$ & $19.9_{\pm 20.6}$         & $\mathbf{17.1_{\pm 7.1}}$ \\
                                  & AUC $\uparrow$           & $18.1$            & $\mathbf{60.8}$           & $30.9$                    \\
            \Block{3-1}{\xhand}   & $e_t$ (cm) $\downarrow$  & $46.4_{\pm 42.8}$ & $\mathbf{4.5_{\pm 1.1}}$  & $8.1_{\pm 4.5}$           \\
                                  & $e_r$ (deg) $\downarrow$ & $73.4_{\pm 45.6}$ & $14.3_{\pm 6.2}$          & $\mathbf{13.2_{\pm 8.1}}$ \\
                                  & AUC $\uparrow$           & $31.1$            & $56.6$                    & $\mathbf{64.0}$           \\
            \bottomrule
        \end{NiceTabular}}
    \caption{\textbf{Comparison with \kinematic retargeting on \gigahand.} Mean$_{\pm\text{std}}$ over 5 seeds. AUC denotes ADD-AUC in percent (\Cref{sec:implementation_details}).}
    \label{tab:optimization_gigahand}
\end{table}

We report per-trajectory tracking metrics for \oakink and \gigahand datasets in the following tables.
\begin{table}[p]
    \centering\small
    \begin{tabular}{lllccc}
        \toprule
        Dataset                    & Task                                    & Metric             & Kinematic                 & SPIDER                     & SPIDER+C                   \\
        \midrule
        \multirow{32}{*}{oakink}   & \multirow{4}{*}{lift\_board}            & $e_\text{ft}$ (cm) & \textbf{$1.6_{\pm 0.0}$}  & $2.8_{\pm 0.4}$            & $4.4_{\pm 0.9}$            \\
                                   &                                         & $e_j$ (rad)        & $0.06_{\pm 0.00}$         & \textbf{$0.03_{\pm 0.00}$} & $0.04_{\pm 0.00}$          \\
                                   &                                         & $e_t$ (cm)         & $3.8_{\pm 0.0}$           & $1.9_{\pm 0.2}$            & \textbf{$1.8_{\pm 0.2}$}   \\
                                   &                                         & $e_r$ (deg)        & $12.5_{\pm 0.0}$          & $7.6_{\pm 1.1}$            & \textbf{$4.4_{\pm 1.1}$}   \\
        \cline{2-6}
                                   & \multirow{4}{*}{pick\_spoon\_bowl}      & $e_\text{ft}$ (cm) & \textbf{$0.9_{\pm 0.0}$}  & $2.4_{\pm 0.6}$            & $3.8_{\pm 1.9}$            \\
                                   &                                         & $e_j$ (rad)        & $0.05_{\pm 0.00}$         & \textbf{$0.03_{\pm 0.00}$} & $0.08_{\pm 0.09}$          \\
                                   &                                         & $e_t$ (cm)         & $6.1_{\pm 0.0}$           & \textbf{$4.5_{\pm 0.8}$}   & $7.2_{\pm 1.4}$            \\
                                   &                                         & $e_r$ (deg)        & $56.3_{\pm 0.0}$          & \textbf{$20.9_{\pm 2.0}$}  & $50.7_{\pm 26.4}$          \\
        \cline{2-6}
                                   & \multirow{4}{*}{pour\_tube}             & $e_\text{ft}$ (cm) & $0.9_{\pm 0.0}$           & \textbf{$0.7_{\pm 0.2}$}   & $5.7_{\pm 0.2}$            \\
                                   &                                         & $e_j$ (rad)        & $0.06_{\pm 0.00}$         & $0.20_{\pm 0.15}$          & \textbf{$0.04_{\pm 0.00}$} \\
                                   &                                         & $e_t$ (cm)         & $23.3_{\pm 0.0}$          & $23.6_{\pm 2.5}$           & \textbf{$5.7_{\pm 0.4}$}   \\
                                   &                                         & $e_r$ (deg)        & $109.0_{\pm 0.0}$         & $110.2_{\pm 14.2}$         & \textbf{$20.6_{\pm 2.5}$}  \\
        \cline{2-6}
                                   & \multirow{4}{*}{stir\_beaker}           & $e_\text{ft}$ (cm) & \textbf{$1.3_{\pm 0.0}$}  & $1.9_{\pm 0.5}$            & $2.9_{\pm 0.2}$            \\
                                   &                                         & $e_j$ (rad)        & $0.06_{\pm 0.00}$         & $0.02_{\pm 0.00}$          & \textbf{$0.02_{\pm 0.00}$} \\
                                   &                                         & $e_t$ (cm)         & $15.4_{\pm 0.0}$          & $13.5_{\pm 1.4}$           & \textbf{$7.3_{\pm 3.1}$}   \\
                                   &                                         & $e_r$ (deg)        & $46.9_{\pm 0.0}$          & $53.1_{\pm 11.7}$          & \textbf{$24.7_{\pm 5.2}$}  \\
        \cline{2-6}
                                   & \multirow{4}{*}{uncap\_alcohol\_burner} & $e_\text{ft}$ (cm) & \textbf{$1.1_{\pm 0.0}$}  & $1.4_{\pm 0.2}$            & $2.6_{\pm 0.5}$            \\
                                   &                                         & $e_j$ (rad)        & $0.06_{\pm 0.00}$         & \textbf{$0.01_{\pm 0.00}$} & $0.02_{\pm 0.00}$          \\
                                   &                                         & $e_t$ (cm)         & $3.3_{\pm 0.0}$           & \textbf{$2.6_{\pm 0.2}$}   & $3.1_{\pm 0.4}$            \\
                                   &                                         & $e_r$ (deg)        & $22.6_{\pm 0.0}$          & $24.4_{\pm 0.9}$           & \textbf{$9.9_{\pm 2.5}$}   \\
        \cline{2-6}
                                   & \multirow{4}{*}{unplug}                 & $e_\text{ft}$ (cm) & \textbf{$1.0_{\pm 0.0}$}  & $1.2_{\pm 0.1}$            & $2.7_{\pm 0.3}$            \\
                                   &                                         & $e_j$ (rad)        & $0.06_{\pm 0.00}$         & \textbf{$0.02_{\pm 0.00}$} & $0.06_{\pm 0.06}$          \\
                                   &                                         & $e_t$ (cm)         & $6.4_{\pm 0.0}$           & \textbf{$3.8_{\pm 1.6}$}   & $6.0_{\pm 1.5}$            \\
                                   &                                         & $e_r$ (deg)        & \textbf{$14.9_{\pm 0.0}$} & $22.4_{\pm 4.4}$           & $25.4_{\pm 12.2}$          \\
        \cline{2-6}
                                   & \multirow{4}{*}{wipe\_board}            & $e_\text{ft}$ (cm) & \textbf{$0.9_{\pm 0.0}$}  & $2.1_{\pm 0.3}$            & $3.0_{\pm 0.2}$            \\
                                   &                                         & $e_j$ (rad)        & $0.05_{\pm 0.00}$         & \textbf{$0.02_{\pm 0.00}$} & $0.05_{\pm 0.00}$          \\
                                   &                                         & $e_t$ (cm)         & $7.5_{\pm 0.0}$           & $3.7_{\pm 0.6}$            & \textbf{$0.9_{\pm 0.1}$}   \\
                                   &                                         & $e_r$ (deg)        & $47.0_{\pm 0.0}$          & $24.3_{\pm 4.6}$           & \textbf{$3.3_{\pm 0.4}$}   \\
        \cline{2-6}
                                   & \multirow{4}{*}{\textbf{Avg.}}          & $e_\text{ft}$ (cm) & \textbf{$1.1_{\pm 0.2}$}  & $1.8_{\pm 0.7}$            & $3.6_{\pm 1.0}$            \\
                                   &                                         & $e_j$ (rad)        & $0.06_{\pm 0.00}$         & $0.05_{\pm 0.06}$          & \textbf{$0.04_{\pm 0.02}$} \\
                                   &                                         & $e_t$ (cm)         & $9.4_{\pm 6.8}$           & $7.7_{\pm 7.4}$            & \textbf{$4.6_{\pm 2.4}$}   \\
                                   &                                         & $e_r$ (deg)        & $44.2_{\pm 30.9}$         & $37.6_{\pm 32.2}$          & \textbf{$19.9_{\pm 15.2}$} \\
        \midrule
        \multirow{16}{*}{gigahand} & \multirow{4}{*}{p36-tea}                & $e_\text{ft}$ (cm) & \textbf{$1.7_{\pm 0.4}$}  & $4.9_{\pm 0.5}$            & $7.0_{\pm 1.5}$            \\
                                   &                                         & $e_j$ (rad)        & $0.29_{\pm 0.17}$         & $0.12_{\pm 0.13}$          & \textbf{$0.11_{\pm 0.01}$} \\
                                   &                                         & $e_t$ (cm)         & $11.1_{\pm 0.0}$          & \textbf{$3.6_{\pm 1.1}$}   & $4.4_{\pm 2.9}$            \\
                                   &                                         & $e_r$ (deg)        & $72.9_{\pm 0.0}$          & $16.1_{\pm 4.5}$           & \textbf{$7.9_{\pm 2.6}$}   \\
        \cline{2-6}
                                   & \multirow{4}{*}{p44-dog}                & $e_\text{ft}$ (cm) & $4.0_{\pm 0.1}$           & $5.8_{\pm 0.6}$            & \textbf{$2.0_{\pm 0.0}$}   \\
                                   &                                         & $e_j$ (rad)        & $0.20_{\pm 0.09}$         & $0.10_{\pm 0.08}$          & \textbf{$0.05_{\pm 0.00}$} \\
                                   &                                         & $e_t$ (cm)         & $37.7_{\pm 0.0}$          & \textbf{$4.7_{\pm 0.7}$}   & $11.3_{\pm 0.8}$           \\
                                   &                                         & $e_r$ (deg)        & $40.0_{\pm 0.0}$          & \textbf{$4.6_{\pm 0.4}$}   & $13.1_{\pm 0.8}$           \\
        \cline{2-6}
                                   & \multirow{4}{*}{p52-instrument}         & $e_\text{ft}$ (cm) & \textbf{$3.1_{\pm 0.0}$}  & $4.2_{\pm 0.1}$            & $4.5_{\pm 0.3}$            \\
                                   &                                         & $e_j$ (rad)        & $0.08_{\pm 0.00}$         & \textbf{$0.07_{\pm 0.00}$} & $0.13_{\pm 0.12}$          \\
                                   &                                         & $e_t$ (cm)         & $22.6_{\pm 0.0}$          & \textbf{$5.2_{\pm 0.3}$}   & $10.4_{\pm 2.3}$           \\
                                   &                                         & $e_r$ (deg)        & $17.6_{\pm 0.0}$          & \textbf{$6.7_{\pm 0.1}$}   & $8.7_{\pm 1.5}$            \\
        \cline{2-6}
                                   & \multirow{4}{*}{\textbf{Avg.}}          & $e_\text{ft}$ (cm) & \textbf{$3.0_{\pm 0.9}$}  & $5.0_{\pm 0.6}$            & $4.5_{\pm 2.0}$            \\
                                   &                                         & $e_j$ (rad)        & $0.19_{\pm 0.09}$         & \textbf{$0.10_{\pm 0.02}$} & $0.10_{\pm 0.04}$          \\
                                   &                                         & $e_t$ (cm)         & $23.8_{\pm 10.9}$         & \textbf{$4.5_{\pm 0.7}$}   & $8.7_{\pm 3.1}$            \\
                                   &                                         & $e_r$ (deg)        & $43.5_{\pm 22.7}$         & \textbf{$9.1_{\pm 5.0}$}   & $9.9_{\pm 2.3}$            \\
        \bottomrule
    \end{tabular}
    \caption{Per-trajectory tracking metrics for \texttt{allegro}. Mean $\pm$ std across 5 seeds.}
    \label{tab:tracking_detail_allegro}
\end{table}

\begin{table}[p]
    \centering\small
    \begin{tabular}{lllccc}
        \toprule
        Dataset                    & Task                                    & Metric             & Kinematic                  & SPIDER                     & SPIDER+C                   \\
        \midrule
        \multirow{32}{*}{oakink}   & \multirow{4}{*}{lift\_board}            & $e_\text{ft}$ (cm) & $6.5_{\pm 0.0}$            & $3.7_{\pm 1.1}$            & \textbf{$2.0_{\pm 0.4}$}   \\
                                   &                                         & $e_j$ (rad)        & $0.19_{\pm 0.07}$          & $0.04_{\pm 0.00}$          & \textbf{$0.02_{\pm 0.00}$} \\
                                   &                                         & $e_t$ (cm)         & $18.0_{\pm 0.0}$           & $7.8_{\pm 2.0}$            & \textbf{$5.5_{\pm 2.6}$}   \\
                                   &                                         & $e_r$ (deg)        & $19.3_{\pm 0.0}$           & $6.6_{\pm 2.3}$            & \textbf{$4.9_{\pm 1.8}$}   \\
        \cline{2-6}
                                   & \multirow{4}{*}{pick\_spoon\_bowl}      & $e_\text{ft}$ (cm) & \textbf{$1.3_{\pm 0.0}$}   & $3.3_{\pm 0.6}$            & $7.0_{\pm 0.3}$            \\
                                   &                                         & $e_j$ (rad)        & $0.13_{\pm 0.07}$          & \textbf{$0.02_{\pm 0.00}$} & $0.12_{\pm 0.00}$          \\
                                   &                                         & $e_t$ (cm)         & $12.4_{\pm 0.0}$           & \textbf{$6.5_{\pm 1.8}$}   & $9.0_{\pm 0.3}$            \\
                                   &                                         & $e_r$ (deg)        & $53.7_{\pm 0.0}$           & $64.3_{\pm 12.4}$          & \textbf{$41.1_{\pm 12.3}$} \\
        \cline{2-6}
                                   & \multirow{4}{*}{pour\_tube}             & $e_\text{ft}$ (cm) & \textbf{$1.8_{\pm 0.4}$}   & $2.2_{\pm 0.4}$            & $8.6_{\pm 1.3}$            \\
                                   &                                         & $e_j$ (rad)        & $0.16_{\pm 0.10}$          & $0.06_{\pm 0.09}$          & \textbf{$0.01_{\pm 0.00}$} \\
                                   &                                         & $e_t$ (cm)         & $30.9_{\pm 0.0}$           & $30.0_{\pm 2.3}$           & \textbf{$26.6_{\pm 1.0}$}  \\
                                   &                                         & $e_r$ (deg)        & $93.2_{\pm 0.0}$           & $126.7_{\pm 7.2}$          & \textbf{$92.6_{\pm 8.7}$}  \\
        \cline{2-6}
                                   & \multirow{4}{*}{stir\_beaker}           & $e_\text{ft}$ (cm) & \textbf{$3.5_{\pm 0.1}$}   & $3.5_{\pm 0.3}$            & $8.4_{\pm 0.3}$            \\
                                   &                                         & $e_j$ (rad)        & $0.09_{\pm 0.05}$          & \textbf{$0.05_{\pm 0.04}$} & $0.08_{\pm 0.00}$          \\
                                   &                                         & $e_t$ (cm)         & $14.2_{\pm 0.0}$           & \textbf{$8.3_{\pm 3.2}$}   & $19.5_{\pm 0.7}$           \\
                                   &                                         & $e_r$ (deg)        & $72.6_{\pm 0.0}$           & \textbf{$38.8_{\pm 6.1}$}  & $74.2_{\pm 5.7}$           \\
        \cline{2-6}
                                   & \multirow{4}{*}{uncap\_alcohol\_burner} & $e_\text{ft}$ (cm) & \textbf{$3.5_{\pm 0.4}$}   & $3.8_{\pm 1.6}$            & $4.0_{\pm 0.8}$            \\
                                   &                                         & $e_j$ (rad)        & $0.26_{\pm 0.15}$          & $0.07_{\pm 0.09}$          & \textbf{$0.05_{\pm 0.00}$} \\
                                   &                                         & $e_t$ (cm)         & $10.1_{\pm 0.0}$           & $5.1_{\pm 2.4}$            & \textbf{$4.0_{\pm 2.0}$}   \\
                                   &                                         & $e_r$ (deg)        & $75.6_{\pm 0.0}$           & $43.5_{\pm 11.7}$          & \textbf{$12.4_{\pm 2.9}$}  \\
        \cline{2-6}
                                   & \multirow{4}{*}{unplug}                 & $e_\text{ft}$ (cm) & $3.2_{\pm 0.1}$            & \textbf{$2.2_{\pm 0.5}$}   & $2.6_{\pm 0.4}$            \\
                                   &                                         & $e_j$ (rad)        & $0.09_{\pm 0.05}$          & $0.05_{\pm 0.05}$          & \textbf{$0.01_{\pm 0.00}$} \\
                                   &                                         & $e_t$ (cm)         & $422.5_{\pm 0.0}$          & \textbf{$1.6_{\pm 0.3}$}   & $5.4_{\pm 2.0}$            \\
                                   &                                         & $e_r$ (deg)        & $98.9_{\pm 0.0}$           & \textbf{$12.6_{\pm 0.8}$}  & $19.9_{\pm 7.3}$           \\
        \cline{2-6}
                                   & \multirow{4}{*}{wipe\_board}            & $e_\text{ft}$ (cm) & \textbf{$3.0_{\pm 0.1}$}   & $3.3_{\pm 0.3}$            & $4.3_{\pm 0.4}$            \\
                                   &                                         & $e_j$ (rad)        & $0.15_{\pm 0.08}$          & $0.07_{\pm 0.07}$          & \textbf{$0.01_{\pm 0.00}$} \\
                                   &                                         & $e_t$ (cm)         & $16.6_{\pm 0.0}$           & \textbf{$4.9_{\pm 0.2}$}   & $6.2_{\pm 1.5}$            \\
                                   &                                         & $e_r$ (deg)        & $104.0_{\pm 0.0}$          & $22.4_{\pm 0.6}$           & \textbf{$9.8_{\pm 1.0}$}   \\
        \cline{2-6}
                                   & \multirow{4}{*}{\textbf{Avg.}}          & $e_\text{ft}$ (cm) & $3.3_{\pm 1.6}$            & \textbf{$3.1_{\pm 0.6}$}   & $5.3_{\pm 2.5}$            \\
                                   &                                         & $e_j$ (rad)        & $0.15_{\pm 0.05}$          & $0.05_{\pm 0.02}$          & \textbf{$0.04_{\pm 0.04}$} \\
                                   &                                         & $e_t$ (cm)         & $74.9_{\pm 142.0}$         & \textbf{$9.2_{\pm 8.8}$}   & $10.9_{\pm 8.0}$           \\
                                   &                                         & $e_r$ (deg)        & $73.9_{\pm 27.5}$          & $45.0_{\pm 38.0}$          & \textbf{$36.4_{\pm 32.0}$} \\
        \midrule
        \multirow{16}{*}{gigahand} & \multirow{4}{*}{p36-tea}                & $e_\text{ft}$ (cm) & \textbf{$2.0_{\pm 0.2}$}   & $4.3_{\pm 0.4}$            & $7.2_{\pm 1.6}$            \\
                                   &                                         & $e_j$ (rad)        & $0.11_{\pm 0.04}$          & \textbf{$0.08_{\pm 0.03}$} & $0.11_{\pm 0.02}$          \\
                                   &                                         & $e_t$ (cm)         & $10.8_{\pm 0.0}$           & \textbf{$4.2_{\pm 1.1}$}   & $11.3_{\pm 5.2}$           \\
                                   &                                         & $e_r$ (deg)        & $49.0_{\pm 0.0}$           & \textbf{$21.3_{\pm 0.6}$}  & $21.7_{\pm 2.7}$           \\
        \cline{2-6}
                                   & \multirow{4}{*}{p44-dog}                & $e_\text{ft}$ (cm) & $5.8_{\pm 0.0}$            & $5.5_{\pm 0.1}$            & \textbf{$3.9_{\pm 0.0}$}   \\
                                   &                                         & $e_j$ (rad)        & $0.15_{\pm 0.04}$          & $0.11_{\pm 0.04}$          & \textbf{$0.09_{\pm 0.00}$} \\
                                   &                                         & $e_t$ (cm)         & $24.7_{\pm 0.0}$           & \textbf{$4.7_{\pm 0.2}$}   & $10.9_{\pm 0.5}$           \\
                                   &                                         & $e_r$ (deg)        & $30.3_{\pm 0.0}$           & \textbf{$6.7_{\pm 0.4}$}   & $9.4_{\pm 0.4}$            \\
        \cline{2-6}
                                   & \multirow{4}{*}{p52-instrument}         & $e_\text{ft}$ (cm) & $6.7_{\pm 0.0}$            & \textbf{$3.5_{\pm 0.1}$}   & $3.7_{\pm 0.3}$            \\
                                   &                                         & $e_j$ (rad)        & \textbf{$0.08_{\pm 0.00}$} & $0.10_{\pm 0.00}$          & $0.14_{\pm 0.02}$          \\
                                   &                                         & $e_t$ (cm)         & $84.2_{\pm 0.0}$           & \textbf{$14.8_{\pm 4.0}$}  & $15.4_{\pm 5.5}$           \\
                                   &                                         & $e_r$ (deg)        & $135.1_{\pm 0.0}$          & $31.0_{\pm 1.9}$           & \textbf{$22.2_{\pm 2.3}$}  \\
        \cline{2-6}
                                   & \multirow{4}{*}{\textbf{Avg.}}          & $e_\text{ft}$ (cm) & $4.8_{\pm 2.0}$            & \textbf{$4.4_{\pm 0.8}$}   & $4.9_{\pm 1.6}$            \\
                                   &                                         & $e_j$ (rad)        & $0.11_{\pm 0.03}$          & \textbf{$0.10_{\pm 0.01}$} & $0.11_{\pm 0.02}$          \\
                                   &                                         & $e_t$ (cm)         & $39.9_{\pm 31.8}$          & \textbf{$7.9_{\pm 4.9}$}   & $12.5_{\pm 2.0}$           \\
                                   &                                         & $e_r$ (deg)        & $71.5_{\pm 45.6}$          & $19.7_{\pm 10.0}$          & \textbf{$17.8_{\pm 5.9}$}  \\
        \bottomrule
    \end{tabular}
    \caption{Per-trajectory tracking metrics for \texttt{ability}. Mean $\pm$ std across 5 seeds.}
    \label{tab:tracking_detail_ability}
\end{table}

\begin{table}[p]
    \centering\small
    \begin{tabular}{lllccc}
        \toprule
        Dataset                    & Task                                    & Metric             & Kinematic            & SPIDER                     & SPIDER+C                   \\
        \midrule
        \multirow{32}{*}{oakink}   & \multirow{4}{*}{lift\_board}            & $e_\text{ft}$ (cm) & $7.1_{\pm 0.0}$      & \textbf{$1.8_{\pm 0.2}$}   & $2.4_{\pm 0.7}$            \\
                                   &                                         & $e_j$ (rad)        & $0.05_{\pm 0.00}$    & \textbf{$0.03_{\pm 0.00}$} & $0.04_{\pm 0.00}$          \\
                                   &                                         & $e_t$ (cm)         & $66.1_{\pm 0.0}$     & $8.1_{\pm 0.2}$            & \textbf{$2.6_{\pm 1.9}$}   \\
                                   &                                         & $e_r$ (deg)        & $21.7_{\pm 0.0}$     & \textbf{$6.3_{\pm 0.8}$}   & $8.3_{\pm 1.4}$            \\
        \cline{2-6}
                                   & \multirow{4}{*}{pick\_spoon\_bowl}      & $e_\text{ft}$ (cm) & $6.5_{\pm 0.0}$      & \textbf{$1.1_{\pm 0.3}$}   & $2.0_{\pm 0.3}$            \\
                                   &                                         & $e_j$ (rad)        & $0.14_{\pm 0.05}$    & $0.05_{\pm 0.06}$          & \textbf{$0.02_{\pm 0.01}$} \\
                                   &                                         & $e_t$ (cm)         & $184.1_{\pm 0.0}$    & \textbf{$9.4_{\pm 2.5}$}   & $12.8_{\pm 1.7}$           \\
                                   &                                         & $e_r$ (deg)        & $87.0_{\pm 0.0}$     & \textbf{$44.5_{\pm 8.5}$}  & $52.6_{\pm 12.6}$          \\
        \cline{2-6}
                                   & \multirow{4}{*}{pour\_tube}             & $e_\text{ft}$ (cm) & $7.0_{\pm 0.0}$      & \textbf{$0.4_{\pm 0.1}$}   & $3.7_{\pm 1.4}$            \\
                                   &                                         & $e_j$ (rad)        & $0.05_{\pm 0.00}$    & \textbf{$0.01_{\pm 0.00}$} & $0.02_{\pm 0.01}$          \\
                                   &                                         & $e_t$ (cm)         & $149.4_{\pm 0.0}$    & \textbf{$27.0_{\pm 1.5}$}  & $31.7_{\pm 17.8}$          \\
                                   &                                         & $e_r$ (deg)        & $124.3_{\pm 0.0}$    & \textbf{$85.2_{\pm 8.2}$}  & $88.3_{\pm 17.7}$          \\
        \cline{2-6}
                                   & \multirow{4}{*}{stir\_beaker}           & $e_\text{ft}$ (cm) & $6.9_{\pm 0.0}$      & \textbf{$2.2_{\pm 0.2}$}   & $3.9_{\pm 0.5}$            \\
                                   &                                         & $e_j$ (rad)        & $0.05_{\pm 0.00}$    & $0.03_{\pm 0.00}$          & \textbf{$0.02_{\pm 0.00}$} \\
                                   &                                         & $e_t$ (cm)         & $169.2_{\pm 0.0}$    & \textbf{$4.5_{\pm 1.4}$}   & $17.5_{\pm 1.8}$           \\
                                   &                                         & $e_r$ (deg)        & $106.7_{\pm 0.0}$    & \textbf{$32.4_{\pm 10.7}$} & $65.4_{\pm 9.2}$           \\
        \cline{2-6}
                                   & \multirow{4}{*}{uncap\_alcohol\_burner} & $e_\text{ft}$ (cm) & $7.0_{\pm 0.0}$      & \textbf{$1.5_{\pm 0.2}$}   & $3.4_{\pm 0.5}$            \\
                                   &                                         & $e_j$ (rad)        & $0.05_{\pm 0.00}$    & \textbf{$0.04_{\pm 0.00}$} & $0.07_{\pm 0.00}$          \\
                                   &                                         & $e_t$ (cm)         & $23.9_{\pm 0.0}$     & \textbf{$3.6_{\pm 1.0}$}   & $4.6_{\pm 0.6}$            \\
                                   &                                         & $e_r$ (deg)        & $82.2_{\pm 0.0}$     & \textbf{$30.1_{\pm 5.8}$}  & $36.8_{\pm 28.1}$          \\
        \cline{2-6}
                                   & \multirow{4}{*}{unplug}                 & $e_\text{ft}$ (cm) & $9.2_{\pm 0.0}$      & \textbf{$2.5_{\pm 0.7}$}   & $2.7_{\pm 0.9}$            \\
                                   &                                         & $e_j$ (rad)        & $0.22_{\pm 0.07}$    & $0.07_{\pm 0.09}$          & \textbf{$0.03_{\pm 0.00}$} \\
                                   &                                         & $e_t$ (cm)         & $731.4_{\pm 1313.0}$ & \textbf{$2.8_{\pm 1.2}$}   & $7.1_{\pm 1.0}$            \\
                                   &                                         & $e_r$ (deg)        & $109.1_{\pm 1.1}$    & \textbf{$13.6_{\pm 6.5}$}  & $24.7_{\pm 10.5}$          \\
        \cline{2-6}
                                   & \multirow{4}{*}{wipe\_board}            & $e_\text{ft}$ (cm) & $7.0_{\pm 0.2}$      & \textbf{$1.7_{\pm 0.3}$}   & $3.7_{\pm 0.3}$            \\
                                   &                                         & $e_j$ (rad)        & $0.19_{\pm 0.07}$    & $0.14_{\pm 0.09}$          & \textbf{$0.03_{\pm 0.00}$} \\
                                   &                                         & $e_t$ (cm)         & $51.9_{\pm 18.7}$    & \textbf{$5.9_{\pm 3.7}$}   & $5.9_{\pm 0.8}$            \\
                                   &                                         & $e_r$ (deg)        & $106.1_{\pm 4.7}$    & $70.3_{\pm 21.9}$          & \textbf{$27.0_{\pm 4.6}$}  \\
        \cline{2-6}
                                   & \multirow{4}{*}{\textbf{Avg.}}          & $e_\text{ft}$ (cm) & $7.2_{\pm 0.8}$      & \textbf{$1.6_{\pm 0.6}$}   & $3.1_{\pm 0.7}$            \\
                                   &                                         & $e_j$ (rad)        & $0.11_{\pm 0.07}$    & $0.05_{\pm 0.04}$          & \textbf{$0.03_{\pm 0.02}$} \\
                                   &                                         & $e_t$ (cm)         & $196.6_{\pm 225.8}$  & \textbf{$8.8_{\pm 7.7}$}   & $11.7_{\pm 9.4}$           \\
                                   &                                         & $e_r$ (deg)        & $91.0_{\pm 31.2}$    & \textbf{$40.3_{\pm 26.6}$} & $43.3_{\pm 25.3}$          \\
        \midrule
        \multirow{16}{*}{gigahand} & \multirow{4}{*}{p36-tea}                & $e_\text{ft}$ (cm) & $8.2_{\pm 0.0}$      & \textbf{$4.7_{\pm 0.2}$}   & $6.5_{\pm 0.6}$            \\
                                   &                                         & $e_j$ (rad)        & $0.07_{\pm 0.00}$    & \textbf{$0.07_{\pm 0.00}$} & $0.13_{\pm 0.01}$          \\
                                   &                                         & $e_t$ (cm)         & $11.5_{\pm 0.0}$     & $4.9_{\pm 0.3}$            & \textbf{$4.2_{\pm 1.1}$}   \\
                                   &                                         & $e_r$ (deg)        & $108.4_{\pm 0.0}$    & $32.4_{\pm 1.6}$           & \textbf{$13.7_{\pm 1.3}$}  \\
        \cline{2-6}
                                   & \multirow{4}{*}{p44-dog}                & $e_\text{ft}$ (cm) & $6.3_{\pm 2.4}$      & $3.8_{\pm 0.1}$            & \textbf{$3.6_{\pm 0.0}$}   \\
                                   &                                         & $e_j$ (rad)        & $0.12_{\pm 0.09}$    & $0.07_{\pm 0.00}$          & \textbf{$0.06_{\pm 0.00}$} \\
                                   &                                         & $e_t$ (cm)         & $65.0_{\pm 24.4}$    & \textbf{$4.3_{\pm 0.2}$}   & $10.5_{\pm 0.3}$           \\
                                   &                                         & $e_r$ (deg)        & $60.0_{\pm 21.4}$    & \textbf{$5.9_{\pm 0.2}$}   & $8.4_{\pm 0.2}$            \\
        \cline{2-6}
                                   & \multirow{4}{*}{p52-instrument}         & $e_\text{ft}$ (cm) & $6.5_{\pm 0.0}$      & $4.1_{\pm 0.1}$            & \textbf{$1.7_{\pm 0.3}$}   \\
                                   &                                         & $e_j$ (rad)        & $0.07_{\pm 0.00}$    & \textbf{$0.06_{\pm 0.00}$} & $0.10_{\pm 0.01}$          \\
                                   &                                         & $e_t$ (cm)         & $55.9_{\pm 0.0}$     & \textbf{$4.3_{\pm 0.2}$}   & $19.6_{\pm 7.3}$           \\
                                   &                                         & $e_r$ (deg)        & $45.4_{\pm 0.0}$     & \textbf{$7.1_{\pm 0.5}$}   & $22.4_{\pm 9.8}$           \\
        \cline{2-6}
                                   & \multirow{4}{*}{\textbf{Avg.}}          & $e_\text{ft}$ (cm) & $7.0_{\pm 0.9}$      & $4.2_{\pm 0.4}$            & \textbf{$3.9_{\pm 2.0}$}   \\
                                   &                                         & $e_j$ (rad)        & $0.09_{\pm 0.02}$    & \textbf{$0.06_{\pm 0.00}$} & $0.10_{\pm 0.03}$          \\
                                   &                                         & $e_t$ (cm)         & $44.2_{\pm 23.4}$    & \textbf{$4.5_{\pm 0.3}$}   & $11.4_{\pm 6.3}$           \\
                                   &                                         & $e_r$ (deg)        & $71.3_{\pm 26.9}$    & $15.1_{\pm 12.2}$          & \textbf{$14.8_{\pm 5.8}$}  \\
        \bottomrule
    \end{tabular}
    \caption{Per-trajectory tracking metrics for \texttt{inspire}. Mean $\pm$ std across 5 seeds.}
    \label{tab:tracking_detail_inspire}
\end{table}

\begin{table}[p]
    \centering\small
    \begin{tabular}{lllccc}
        \toprule
        Dataset                    & Task                                    & Metric             & Kinematic         & SPIDER                     & SPIDER+C                   \\
        \midrule
        \multirow{32}{*}{oakink}   & \multirow{4}{*}{lift\_board}            & $e_\text{ft}$ (cm) & $5.2_{\pm 0.0}$   & $3.1_{\pm 0.4}$            & \textbf{$3.0_{\pm 0.3}$}   \\
                                   &                                         & $e_j$ (rad)        & $0.28_{\pm 0.00}$ & $0.03_{\pm 0.00}$          & \textbf{$0.02_{\pm 0.00}$} \\
                                   &                                         & $e_t$ (cm)         & $17.8_{\pm 0.0}$  & $9.2_{\pm 3.6}$            & \textbf{$6.6_{\pm 8.6}$}   \\
                                   &                                         & $e_r$ (deg)        & $54.6_{\pm 0.0}$  & $10.7_{\pm 3.7}$           & \textbf{$9.9_{\pm 11.4}$}  \\
        \cline{2-6}
                                   & \multirow{4}{*}{pick\_spoon\_bowl}      & $e_\text{ft}$ (cm) & $4.1_{\pm 0.0}$   & \textbf{$3.3_{\pm 0.3}$}   & $4.2_{\pm 0.4}$            \\
                                   &                                         & $e_j$ (rad)        & $0.06_{\pm 0.00}$ & \textbf{$0.02_{\pm 0.00}$} & $0.03_{\pm 0.00}$          \\
                                   &                                         & $e_t$ (cm)         & $9.2_{\pm 0.0}$   & \textbf{$4.9_{\pm 0.4}$}   & $6.6_{\pm 1.2}$            \\
                                   &                                         & $e_r$ (deg)        & $98.2_{\pm 0.0}$  & \textbf{$43.3_{\pm 11.1}$} & $54.0_{\pm 11.8}$          \\
        \cline{2-6}
                                   & \multirow{4}{*}{pour\_tube}             & $e_\text{ft}$ (cm) & $5.1_{\pm 0.0}$   & \textbf{$1.7_{\pm 0.5}$}   & $5.9_{\pm 0.3}$            \\
                                   &                                         & $e_j$ (rad)        & $0.28_{\pm 0.00}$ & \textbf{$0.02_{\pm 0.00}$} & $0.03_{\pm 0.00}$          \\
                                   &                                         & $e_t$ (cm)         & $35.8_{\pm 0.0}$  & $28.8_{\pm 0.6}$           & \textbf{$14.6_{\pm 0.4}$}  \\
                                   &                                         & $e_r$ (deg)        & $103.4_{\pm 0.0}$ & $82.5_{\pm 10.5}$          & \textbf{$55.6_{\pm 5.9}$}  \\
        \cline{2-6}
                                   & \multirow{4}{*}{stir\_beaker}           & $e_\text{ft}$ (cm) & $4.1_{\pm 0.0}$   & \textbf{$2.8_{\pm 0.2}$}   & $3.9_{\pm 0.2}$            \\
                                   &                                         & $e_j$ (rad)        & $0.15_{\pm 0.00}$ & \textbf{$0.02_{\pm 0.00}$} & $0.02_{\pm 0.00}$          \\
                                   &                                         & $e_t$ (cm)         & $21.2_{\pm 0.0}$  & $10.6_{\pm 1.7}$           & \textbf{$9.9_{\pm 0.5}$}   \\
                                   &                                         & $e_r$ (deg)        & $112.1_{\pm 0.0}$ & $48.3_{\pm 8.2}$           & \textbf{$35.4_{\pm 6.1}$}  \\
        \cline{2-6}
                                   & \multirow{4}{*}{uncap\_alcohol\_burner} & $e_\text{ft}$ (cm) & $3.3_{\pm 0.0}$   & \textbf{$1.4_{\pm 0.1}$}   & $2.1_{\pm 0.2}$            \\
                                   &                                         & $e_j$ (rad)        & $0.05_{\pm 0.00}$ & $0.02_{\pm 0.00}$          & \textbf{$0.01_{\pm 0.00}$} \\
                                   &                                         & $e_t$ (cm)         & $22.3_{\pm 0.0}$  & \textbf{$3.0_{\pm 0.8}$}   & $4.5_{\pm 0.9}$            \\
                                   &                                         & $e_r$ (deg)        & $106.4_{\pm 0.0}$ & \textbf{$5.8_{\pm 0.4}$}   & $16.9_{\pm 9.1}$           \\
        \cline{2-6}
                                   & \multirow{4}{*}{unplug}                 & $e_\text{ft}$ (cm) & $8.0_{\pm 1.9}$   & \textbf{$2.2_{\pm 0.0}$}   & $3.9_{\pm 0.2}$            \\
                                   &                                         & $e_j$ (rad)        & $0.28_{\pm 0.11}$ & \textbf{$0.02_{\pm 0.00}$} & $0.02_{\pm 0.00}$          \\
                                   &                                         & $e_t$ (cm)         & $12.3_{\pm 7.4}$  & \textbf{$1.9_{\pm 0.1}$}   & $3.8_{\pm 0.2}$            \\
                                   &                                         & $e_r$ (deg)        & $58.8_{\pm 22.5}$ & $12.5_{\pm 0.7}$           & \textbf{$11.5_{\pm 5.2}$}  \\
        \cline{2-6}
                                   & \multirow{4}{*}{wipe\_board}            & $e_\text{ft}$ (cm) & $4.1_{\pm 0.1}$   & \textbf{$3.4_{\pm 0.1}$}   & $4.3_{\pm 0.1}$            \\
                                   &                                         & $e_j$ (rad)        & $0.13_{\pm 0.04}$ & $0.03_{\pm 0.00}$          & \textbf{$0.02_{\pm 0.00}$} \\
                                   &                                         & $e_t$ (cm)         & $6.0_{\pm 0.9}$   & \textbf{$5.1_{\pm 0.3}$}   & $8.2_{\pm 1.0}$            \\
                                   &                                         & $e_r$ (deg)        & $71.8_{\pm 16.8}$ & $28.3_{\pm 1.8}$           & \textbf{$23.6_{\pm 9.3}$}  \\
        \cline{2-6}
                                   & \multirow{4}{*}{\textbf{Avg.}}          & $e_\text{ft}$ (cm) & $4.8_{\pm 1.4}$   & \textbf{$2.6_{\pm 0.7}$}   & $3.9_{\pm 1.1}$            \\
                                   &                                         & $e_j$ (rad)        & $0.18_{\pm 0.10}$ & \textbf{$0.02_{\pm 0.00}$} & $0.02_{\pm 0.01}$          \\
                                   &                                         & $e_t$ (cm)         & $17.8_{\pm 9.3}$  & $9.1_{\pm 8.5}$            & \textbf{$7.8_{\pm 3.4}$}   \\
                                   &                                         & $e_r$ (deg)        & $86.5_{\pm 22.3}$ & $33.1_{\pm 25.3}$          & \textbf{$29.6_{\pm 17.8}$} \\
        \midrule
        \multirow{16}{*}{gigahand} & \multirow{4}{*}{p36-tea}                & $e_\text{ft}$ (cm) & $4.4_{\pm 0.0}$   & $4.8_{\pm 0.7}$            & \textbf{$4.0_{\pm 0.2}$}   \\
                                   &                                         & $e_j$ (rad)        & $0.07_{\pm 0.00}$ & \textbf{$0.05_{\pm 0.00}$} & $0.08_{\pm 0.00}$          \\
                                   &                                         & $e_t$ (cm)         & $15.2_{\pm 0.0}$  & \textbf{$7.1_{\pm 9.1}$}   & $14.1_{\pm 0.9}$           \\
                                   &                                         & $e_r$ (deg)        & $80.7_{\pm 0.0}$  & $36.1_{\pm 28.7}$          & \textbf{$26.9_{\pm 1.2}$}  \\
        \cline{2-6}
                                   & \multirow{4}{*}{p44-dog}                & $e_\text{ft}$ (cm) & $4.8_{\pm 0.1}$   & $5.5_{\pm 0.3}$            & \textbf{$4.3_{\pm 0.0}$}   \\
                                   &                                         & $e_j$ (rad)        & $0.12_{\pm 0.10}$ & $0.05_{\pm 0.00}$          & \textbf{$0.05_{\pm 0.00}$} \\
                                   &                                         & $e_t$ (cm)         & $45.3_{\pm 8.5}$  & \textbf{$4.5_{\pm 0.2}$}   & $11.0_{\pm 0.3}$           \\
                                   &                                         & $e_r$ (deg)        & $42.0_{\pm 5.8}$  & \textbf{$6.6_{\pm 0.3}$}   & $10.2_{\pm 0.2}$           \\
        \cline{2-6}
                                   & \multirow{4}{*}{p52-instrument}         & $e_\text{ft}$ (cm) & $4.9_{\pm 0.0}$   & $5.9_{\pm 0.6}$            & \textbf{$4.1_{\pm 0.1}$}   \\
                                   &                                         & $e_j$ (rad)        & $0.07_{\pm 0.00}$ & \textbf{$0.04_{\pm 0.00}$} & $0.07_{\pm 0.00}$          \\
                                   &                                         & $e_t$ (cm)         & $54.5_{\pm 0.0}$  & $21.8_{\pm 3.1}$           & \textbf{$16.8_{\pm 1.1}$}  \\
                                   &                                         & $e_r$ (deg)        & $54.9_{\pm 0.0}$  & $16.9_{\pm 1.9}$           & \textbf{$14.3_{\pm 0.8}$}  \\
        \cline{2-6}
                                   & \multirow{4}{*}{\textbf{Avg.}}          & $e_\text{ft}$ (cm) & $4.7_{\pm 0.2}$   & $5.4_{\pm 0.5}$            & \textbf{$4.1_{\pm 0.1}$}   \\
                                   &                                         & $e_j$ (rad)        & $0.09_{\pm 0.03}$ & \textbf{$0.05_{\pm 0.00}$} & $0.07_{\pm 0.01}$          \\
                                   &                                         & $e_t$ (cm)         & $38.3_{\pm 16.8}$ & \textbf{$11.1_{\pm 7.6}$}  & $14.0_{\pm 2.4}$           \\
                                   &                                         & $e_r$ (deg)        & $59.2_{\pm 16.1}$ & $19.9_{\pm 12.2}$          & \textbf{$17.1_{\pm 7.1}$}  \\
        \bottomrule
    \end{tabular}
    \caption{Per-trajectory tracking metrics for \texttt{schunk}. Mean $\pm$ std across 5 seeds.}
    \label{tab:tracking_detail_schunk}
\end{table}

\begin{table}[p]
    \centering\small
    \begin{tabular}{lllccc}
        \toprule
        Dataset                    & Task                                    & Metric             & Kinematic                  & SPIDER                     & SPIDER+C                   \\
        \midrule
        \multirow{32}{*}{oakink}   & \multirow{4}{*}{lift\_board}            & $e_\text{ft}$ (cm) & $6.4_{\pm 0.1}$            & \textbf{$2.4_{\pm 0.1}$}   & $2.4_{\pm 0.2}$            \\
                                   &                                         & $e_j$ (rad)        & $0.08_{\pm 0.06}$          & \textbf{$0.03_{\pm 0.00}$} & $0.04_{\pm 0.00}$          \\
                                   &                                         & $e_t$ (cm)         & $74.6_{\pm 18.3}$          & $1.9_{\pm 0.5}$            & \textbf{$1.5_{\pm 0.1}$}   \\
                                   &                                         & $e_r$ (deg)        & $74.3_{\pm 8.0}$           & $9.4_{\pm 0.4}$            & \textbf{$4.3_{\pm 0.7}$}   \\
        \cline{2-6}
                                   & \multirow{4}{*}{pick\_spoon\_bowl}      & $e_\text{ft}$ (cm) & $6.1_{\pm 0.4}$            & \textbf{$2.8_{\pm 0.4}$}   & $4.7_{\pm 1.0}$            \\
                                   &                                         & $e_j$ (rad)        & $0.08_{\pm 0.07}$          & $0.02_{\pm 0.00}$          & \textbf{$0.01_{\pm 0.00}$} \\
                                   &                                         & $e_t$ (cm)         & $29.0_{\pm 0.8}$           & \textbf{$5.4_{\pm 0.2}$}   & $7.5_{\pm 1.3}$            \\
                                   &                                         & $e_r$ (deg)        & $109.1_{\pm 5.0}$          & \textbf{$21.6_{\pm 2.8}$}  & $33.2_{\pm 15.4}$          \\
        \cline{2-6}
                                   & \multirow{4}{*}{pour\_tube}             & $e_\text{ft}$ (cm) & $6.0_{\pm 0.0}$            & \textbf{$1.3_{\pm 0.7}$}   & $5.2_{\pm 3.9}$            \\
                                   &                                         & $e_j$ (rad)        & \textbf{$0.05_{\pm 0.00}$} & $0.12_{\pm 0.12}$          & $0.08_{\pm 0.14}$          \\
                                   &                                         & $e_t$ (cm)         & $51.1_{\pm 0.0}$           & $22.7_{\pm 8.5}$           & \textbf{$10.5_{\pm 3.4}$}  \\
                                   &                                         & $e_r$ (deg)        & $112.5_{\pm 0.0}$          & $95.0_{\pm 27.6}$          & \textbf{$58.6_{\pm 9.7}$}  \\
        \cline{2-6}
                                   & \multirow{4}{*}{stir\_beaker}           & $e_\text{ft}$ (cm) & $6.8_{\pm 0.0}$            & \textbf{$2.7_{\pm 0.5}$}   & $5.0_{\pm 0.3}$            \\
                                   &                                         & $e_j$ (rad)        & $0.05_{\pm 0.00}$          & $0.05_{\pm 0.05}$          & \textbf{$0.03_{\pm 0.00}$} \\
                                   &                                         & $e_t$ (cm)         & $48.2_{\pm 0.0}$           & $9.8_{\pm 2.4}$            & \textbf{$7.0_{\pm 0.9}$}   \\
                                   &                                         & $e_r$ (deg)        & $105.2_{\pm 0.0}$          & $45.0_{\pm 7.3}$           & \textbf{$21.7_{\pm 0.5}$}  \\
        \cline{2-6}
                                   & \multirow{4}{*}{uncap\_alcohol\_burner} & $e_\text{ft}$ (cm) & $5.9_{\pm 0.0}$            & \textbf{$1.6_{\pm 0.2}$}   & $2.4_{\pm 0.2}$            \\
                                   &                                         & $e_j$ (rad)        & $0.06_{\pm 0.00}$          & $0.04_{\pm 0.00}$          & \textbf{$0.02_{\pm 0.00}$} \\
                                   &                                         & $e_t$ (cm)         & $8.4_{\pm 0.0}$            & $5.1_{\pm 1.5}$            & \textbf{$3.4_{\pm 0.3}$}   \\
                                   &                                         & $e_r$ (deg)        & $119.2_{\pm 0.0}$          & $22.1_{\pm 4.3}$           & \textbf{$10.8_{\pm 1.6}$}  \\
        \cline{2-6}
                                   & \multirow{4}{*}{unplug}                 & $e_\text{ft}$ (cm) & $6.4_{\pm 0.0}$            & \textbf{$1.4_{\pm 0.1}$}   & $3.8_{\pm 0.3}$            \\
                                   &                                         & $e_j$ (rad)        & $0.05_{\pm 0.00}$          & \textbf{$0.02_{\pm 0.00}$} & $0.03_{\pm 0.00}$          \\
                                   &                                         & $e_t$ (cm)         & $43.7_{\pm 0.0}$           & \textbf{$1.3_{\pm 0.4}$}   & $3.9_{\pm 0.7}$            \\
                                   &                                         & $e_r$ (deg)        & $125.5_{\pm 0.0}$          & \textbf{$12.4_{\pm 1.1}$}  & $32.0_{\pm 12.3}$          \\
        \cline{2-6}
                                   & \multirow{4}{*}{wipe\_board}            & $e_\text{ft}$ (cm) & $7.1_{\pm 0.0}$            & \textbf{$2.5_{\pm 0.1}$}   & $3.1_{\pm 0.2}$            \\
                                   &                                         & $e_j$ (rad)        & $0.05_{\pm 0.00}$          & $0.03_{\pm 0.00}$          & \textbf{$0.02_{\pm 0.00}$} \\
                                   &                                         & $e_t$ (cm)         & $21.5_{\pm 0.0}$           & $3.9_{\pm 0.3}$            & \textbf{$2.6_{\pm 1.2}$}   \\
                                   &                                         & $e_r$ (deg)        & $113.0_{\pm 0.0}$          & $21.5_{\pm 0.2}$           & \textbf{$8.5_{\pm 6.6}$}   \\
        \cline{2-6}
                                   & \multirow{4}{*}{\textbf{Avg.}}          & $e_\text{ft}$ (cm) & $6.4_{\pm 0.4}$            & \textbf{$2.1_{\pm 0.6}$}   & $3.8_{\pm 1.1}$            \\
                                   &                                         & $e_j$ (rad)        & $0.06_{\pm 0.01}$          & $0.04_{\pm 0.03}$          & \textbf{$0.03_{\pm 0.02}$} \\
                                   &                                         & $e_t$ (cm)         & $39.5_{\pm 20.2}$          & $7.1_{\pm 6.8}$            & \textbf{$5.2_{\pm 3.0}$}   \\
                                   &                                         & $e_r$ (deg)        & $108.4_{\pm 15.2}$         & $32.4_{\pm 27.6}$          & \textbf{$24.1_{\pm 17.5}$} \\
        \midrule
        \multirow{16}{*}{gigahand} & \multirow{4}{*}{p36-tea}                & $e_\text{ft}$ (cm) & $6.9_{\pm 0.0}$            & $5.8_{\pm 0.5}$            & \textbf{$4.2_{\pm 1.0}$}   \\
                                   &                                         & $e_j$ (rad)        & $0.07_{\pm 0.00}$          & \textbf{$0.07_{\pm 0.00}$} & $0.12_{\pm 0.03}$          \\
                                   &                                         & $e_t$ (cm)         & $30.6_{\pm 0.0}$           & \textbf{$3.6_{\pm 0.4}$}   & $8.3_{\pm 4.2}$            \\
                                   &                                         & $e_r$ (deg)        & $114.2_{\pm 0.0}$          & \textbf{$19.4_{\pm 1.1}$}  & $23.3_{\pm 4.4}$           \\
        \cline{2-6}
                                   & \multirow{4}{*}{p44-dog}                & $e_\text{ft}$ (cm) & $5.2_{\pm 1.2}$            & $5.7_{\pm 0.3}$            & \textbf{$4.1_{\pm 0.0}$}   \\
                                   &                                         & $e_j$ (rad)        & $0.16_{\pm 0.16}$          & \textbf{$0.07_{\pm 0.00}$} & $0.07_{\pm 0.00}$          \\
                                   &                                         & $e_t$ (cm)         & $97.8_{\pm 36.5}$          & \textbf{$4.8_{\pm 0.4}$}   & $12.5_{\pm 0.8}$           \\
                                   &                                         & $e_r$ (deg)        & $90.0_{\pm 31.8}$          & \textbf{$5.9_{\pm 0.4}$}   & $11.7_{\pm 0.9}$           \\
        \cline{2-6}
                                   & \multirow{4}{*}{p52-instrument}         & $e_\text{ft}$ (cm) & \textbf{$1.7_{\pm 0.0}$}   & $4.3_{\pm 1.2}$            & $2.9_{\pm 0.2}$            \\
                                   &                                         & $e_j$ (rad)        & \textbf{$0.05_{\pm 0.00}$} & $0.07_{\pm 0.01}$          & $0.06_{\pm 0.00}$          \\
                                   &                                         & $e_t$ (cm)         & $10.7_{\pm 0.0}$           & $5.1_{\pm 1.3}$            & \textbf{$3.4_{\pm 0.7}$}   \\
                                   &                                         & $e_r$ (deg)        & $16.0_{\pm 0.0}$           & $17.5_{\pm 2.7}$           & \textbf{$4.7_{\pm 0.8}$}   \\
        \cline{2-6}
                                   & \multirow{4}{*}{\textbf{Avg.}}          & $e_\text{ft}$ (cm) & $4.6_{\pm 2.2}$            & $5.3_{\pm 0.7}$            & \textbf{$3.7_{\pm 0.6}$}   \\
                                   &                                         & $e_j$ (rad)        & $0.10_{\pm 0.04}$          & \textbf{$0.07_{\pm 0.00}$} & $0.09_{\pm 0.03}$          \\
                                   &                                         & $e_t$ (cm)         & $46.4_{\pm 37.3}$          & \textbf{$4.5_{\pm 0.7}$}   & $8.1_{\pm 3.7}$            \\
                                   &                                         & $e_r$ (deg)        & $73.4_{\pm 41.8}$          & $14.3_{\pm 6.0}$           & \textbf{$13.2_{\pm 7.7}$}  \\
        \bottomrule
    \end{tabular}
    \caption{Per-trajectory tracking metrics for \texttt{xhand}. Mean $\pm$ std across 5 seeds.}
    \label{tab:tracking_detail_xhand}
\end{table}
\newpage

\subsection{\maniptrans Retargeting Results}
\label{subsec:maniptrans_results}
To match the exact setting of \maniptrans and avoid discrepancies caused by different simulators, we directly port the training environment and evaluation protocol from \maniptrans on the \oakink dataset.
The following are the details of the training and evaluation protocol when comparing \spider with the \maniptrans RL baseline.

\textbf{RL Baseline Training. }
We train one residual RL policy per trajectory using the \maniptrans framework on 16 randomly picked single-hand picking trajectories from the \oakink dataset.
Each policy is trained for 100 epochs on a single trajectory with the Inspire dexterous hand using the same 12-DoF simulation model as in the original \maniptrans paper.
The action space consists of 18-dimensional residual actions (6 wrist force/torque + 12 joint DOF positions), applied on top of a pre-trained base imitator that provides nominal actions from the current observations.
We use an action smoothing factor of 0.4 as suggested in the baseline repository.

\textbf{RL Evaluation. }
We evaluate each trained policy with 5 random seeds (0-4) to capture stochasticity.
For each seed, we run 2048 rollouts across 256 parallel IsaacGym environments without early termination, deterministic initial states from the demo trajectory, and
save the 256 longest rollouts to HDF5.

\textbf{SPIDER Retargeting. } We run \spider with 1024 parallel IsaacGym environments.
\spider optimizes only the 18-dim residual actions; the base imitator provides nominal actions at each step.
The action smoothing factor is 0.4, matching RL.
For the reward, we use the same reward function as the RL baseline but remove the action penalty term for efficiency (since our action is smooth by interpolation).
Each trajectory is run with 5 seeds. All hyperparameters are set as in \Cref{tab:spider-maniptrans-params}.

\begin{table*}[h]
    \centering
    \caption{SPIDER hyperparameters for the ManipTrans (IsaacGym) experiments.}
    \label{tab:spider-maniptrans-params}
    \begin{tabular}{lcc}
        \toprule
        \textbf{Parameter}              & \textbf{Symbol}          & \textbf{Value}         \\
        \midrule
        \multicolumn{3}{l}{\textit{Simulation}}                                             \\
        Simulator                       & --                       & IsaacGym (GPU PhysX)   \\
        Simulation timestep             & $\Delta t_{\text{sim}}$  & $1/120$ s              \\
        Control timestep                & $\Delta t_{\text{ctrl}}$ & $0.2$ s                \\
        Knot point spacing              & $\Delta t_{\text{knot}}$ & $0.2$ s                \\
        Action dimensions               & $n_u$                    & $18$                   \\
        \midrule
        \multicolumn{3}{l}{\textit{Sampling-based Optimization}}                            \\
        Planning horizon                & $H$                      & $0.6$ s (36 sim steps) \\
        Parallel samples                & $N_w$                    & $1024$                 \\
        Max sampling iterations         & $N$                      & $32$                   \\
        Early stopping threshold        & $\epsilon$               & $0.001$                \\
        Early stopping patience         & --                       & $4$ iterations         \\
        \midrule
        \multicolumn{3}{l}{\textit{Noise Schedule}}                                         \\
        Initial noise scale             & $\sigma_{\text{first}}$  & $0.5$                  \\
        Final noise scale (horizon end) & $\sigma_{\text{last}}$   & $1.0$                  \\
        Noise decay per iteration       & $\beta$                  & $0.93$                 \\
        Wrist pos/rot noise scale       & --                       & $0.0$                  \\
        Joint noise scale               & --                       & $1.0$                  \\
        \midrule
        \multicolumn{3}{l}{\textit{Reward}}                                                 \\
        Terminal reward scale           & $\alpha_T$               & $10.0$                 \\
        Sampling temperature            & $\tau$                   & $1.0$                  \\
        \midrule
        \multicolumn{3}{l}{\textit{ManipTrans-specific}}                                    \\
        Actions moving average          & $\gamma$                 & $0.4$                  \\
        \bottomrule
    \end{tabular}
\end{table*}

\textbf{Evaluation Metrics. } We compute four metrics against ground-truth demo data from reference motion, following the \maniptrans evaluation protocol.
Evaluation metrics are:
\begin{itemize}
    \item $e_t$: mean $L_2$ object translation error (cm)
    \item $e_r$: geodesic object rotation error (degrees), computed as the angle of the axis-angle representation of $R_{gt} R_{pred}^T$
    \item $e_j$: mean $L_2$ hand joint position error (cm), computed over all body positions including wrist
    \item $e_{ft}$: mean $L_2$ fingertip position error (cm), computed over 5 fingertip joints (thumb, index, middle, ring, pinky)
\end{itemize}
All metrics are averaged over the trajectory length, then reported as mean $\pm$ std across 5 seeds per trajectory.

\textbf{Experiment Results. }
We randomly select 16 trajectories (8 left-hand, 8 right-hand) from the \oakink dataset.
Full results are shown in \Cref{tab:maniptrans_comparison}.
For some trajectories, \spider's standard deviation is $0$ because the object is not moving and \spider perfectly tracks the hand motion.

\begin{table*}[t]
    \centering
    \caption{Multi-seed comparison between \spider and \maniptrans RL baseline on 16 trajectories from \oakink dataset. Each method is evaluated with 5 seeds per trajectory. Metrics: $e_t$ = object translation error (cm), $e_r$ = object rotation error (deg), $e_j$ = hand joint position error (cm), $e_{ft}$ = fingertip position error (cm). Values reported as mean $\pm$ std.}
    \label{tab:maniptrans_comparison}
    \resizebox{\textwidth}{!}{%
        \begin{tabular}{llc|cccc}
            \toprule
            \textbf{Data ID}                  & \textbf{Side}          & \textbf{Method} & $e_t$ (cm) $\downarrow$    & $e_r$ (deg) $\downarrow$    & $e_j$ (cm) $\downarrow$  & $e_{ft}$ (cm) $\downarrow$ \\
            \midrule
            \multirow{2}{*}{01857@10}         & \multirow{2}{*}{right} & \spider         & $34.81 \pm 0.30$           & $101.76 \pm 1.12$           & $\mathbf{4.73 \pm 0.01}$ & $\mathbf{4.10 \pm 0.01}$   \\
                                              &                        & \maniptrans     & $\mathbf{34.50 \pm 3.34}$  & $\mathbf{112.67 \pm 19.26}$ & $25.98 \pm 19.13$        & $25.59 \pm 18.77$          \\
            \midrule
            \multirow{2}{*}{031b7@0}          & \multirow{2}{*}{right} & \spider         & $40.74 \pm 0.74$           & $47.17 \pm 22.58$           & $\mathbf{5.03 \pm 0.01}$ & $\mathbf{3.95 \pm 0.01}$   \\
                                              &                        & \maniptrans     & $\mathbf{40.17 \pm 0.07}$  & $\mathbf{35.20 \pm 3.39}$   & $101.47 \pm 2.08$        & $98.49 \pm 2.02$           \\
            \midrule
            \multirow{2}{*}{03ac9@0}          & \multirow{2}{*}{right} & \spider         & $\mathbf{44.27 \pm 0.00}$  & $80.72 \pm 0.00$            & $\mathbf{4.80 \pm 0.00}$ & $\mathbf{3.94 \pm 0.00}$   \\
                                              &                        & \maniptrans     & $44.31 \pm 0.00$           & $\mathbf{80.54 \pm 0.00}$   & $293.28 \pm 20.40$       & $294.40 \pm 20.62$         \\
            \midrule
            \multirow{2}{*}{07301@0}          & \multirow{2}{*}{right} & \spider         & $25.96 \pm 0.02$           & $\mathbf{27.95 \pm 0.67}$   & $\mathbf{5.09 \pm 0.01}$ & $\mathbf{4.05 \pm 0.00}$   \\
                                              &                        & \maniptrans     & $\mathbf{13.40 \pm 4.65}$  & $137.87 \pm 2.71$           & $10.20 \pm 1.02$         & $10.03 \pm 1.07$           \\
            \midrule
            \multirow{2}{*}{07bb1@1}          & \multirow{2}{*}{right} & \spider         & $\mathbf{31.09 \pm 15.33}$ & $\mathbf{29.26 \pm 3.27}$   & $\mathbf{4.86 \pm 0.06}$ & $\mathbf{4.03 \pm 0.07}$   \\
                                              &                        & \maniptrans     & $34.43 \pm 2.16$           & $57.47 \pm 40.98$           & $11.12 \pm 2.91$         & $10.04 \pm 3.18$           \\
            \midrule
            \multirow{2}{*}{0b3a0@6}          & \multirow{2}{*}{left}  & \spider         & $26.88 \pm 0.01$           & $\mathbf{40.46 \pm 0.16}$   & $\mathbf{4.79 \pm 0.00}$ & $\mathbf{3.96 \pm 0.00}$   \\
                                              &                        & \maniptrans     & $\mathbf{25.64 \pm 0.38}$  & $73.22 \pm 10.33$           & $8.05 \pm 0.05$          & $7.63 \pm 0.05$            \\
            \midrule
            \multirow{2}{*}{0f4a7@1}          & \multirow{2}{*}{left}  & \spider         & $\mathbf{5.99 \pm 2.61}$   & $\mathbf{56.57 \pm 10.51}$  & $\mathbf{4.57 \pm 0.07}$ & $\mathbf{3.94 \pm 0.09}$   \\
                                              &                        & \maniptrans     & $21.33 \pm 14.55$          & $57.51 \pm 12.53$           & $9.93 \pm 2.07$          & $9.68 \pm 2.29$            \\
            \midrule
            \multirow{2}{*}{0f900@0}          & \multirow{2}{*}{left}  & \spider         & $20.56 \pm 10.22$          & $\mathbf{52.75 \pm 31.28}$  & $\mathbf{4.45 \pm 0.01}$ & $\mathbf{3.89 \pm 0.04}$   \\
                                              &                        & \maniptrans     & $\mathbf{11.66 \pm 10.62}$ & $58.05 \pm 41.20$           & $7.19 \pm 2.73$          & $6.34 \pm 3.06$            \\
            \midrule
            \multirow{2}{*}{155cb@0}          & \multirow{2}{*}{right} & \spider         & $45.45 \pm 0.00$           & $51.92 \pm 0.00$            & $\mathbf{4.89 \pm 0.02}$ & $\mathbf{3.98 \pm 0.01}$   \\
                                              &                        & \maniptrans     & $\mathbf{45.41 \pm 0.00}$  & $\mathbf{51.81 \pm 0.00}$   & $54.05 \pm 1.19$         & $52.83 \pm 1.17$           \\
            \midrule
            \multirow{2}{*}{235ac@0}          & \multirow{2}{*}{right} & \spider         & $22.06 \pm 0.40$           & $\mathbf{59.61 \pm 2.24}$   & $\mathbf{4.94 \pm 0.00}$ & $\mathbf{3.91 \pm 0.00}$   \\
                                              &                        & \maniptrans     & $\mathbf{17.04 \pm 3.14}$  & $70.61 \pm 44.86$           & $21.54 \pm 7.76$         & $23.50 \pm 6.88$           \\
            \midrule
            \multirow{2}{*}{2ddc8@2}          & \multirow{2}{*}{left}  & \spider         & $\mathbf{14.00 \pm 0.00}$  & $85.78 \pm 0.00$            & $\mathbf{4.60 \pm 0.01}$ & $\mathbf{3.96 \pm 0.01}$   \\
                                              &                        & \maniptrans     & $14.03 \pm 0.00$           & $\mathbf{85.77 \pm 0.00}$   & $6.40 \pm 0.03$          & $5.36 \pm 0.02$            \\
            \midrule
            \multirow{2}{*}{380c7@0}          & \multirow{2}{*}{left}  & \spider         & $\mathbf{11.36 \pm 4.52}$  & $23.57 \pm 12.31$           & $\mathbf{5.82 \pm 0.57}$ & $\mathbf{5.87 \pm 0.81}$   \\
                                              &                        & \maniptrans     & $16.97 \pm 0.17$           & $\mathbf{15.78 \pm 8.39}$   & $19.06 \pm 0.81$         & $20.73 \pm 0.51$           \\
            \midrule
            \multirow{2}{*}{39e5e@5}          & \multirow{2}{*}{left}  & \spider         & $\mathbf{12.64 \pm 0.01}$  & $\mathbf{33.06 \pm 0.30}$   & $\mathbf{4.73 \pm 0.02}$ & $\mathbf{4.01 \pm 0.02}$   \\
                                              &                        & \maniptrans     & $18.61 \pm 4.90$           & $99.13 \pm 26.68$           & $19.24 \pm 2.30$         & $18.77 \pm 2.90$           \\
            \midrule
            \multirow{2}{*}{49793@0}          & \multirow{2}{*}{right} & \spider         & $\mathbf{21.04 \pm 9.41}$  & $\mathbf{49.52 \pm 3.41}$   & $\mathbf{4.64 \pm 0.12}$ & $\mathbf{4.03 \pm 0.18}$   \\
                                              &                        & \maniptrans     & $23.99 \pm 3.82$           & $122.98 \pm 20.92$          & $1618.07 \pm 215.12$     & $1617.23 \pm 214.88$       \\
            \midrule
            \multirow{2}{*}{70ca2@0}          & \multirow{2}{*}{left}  & \spider         & $6.25 \pm 4.36$            & $85.93 \pm 13.46$           & $\mathbf{4.55 \pm 0.02}$ & $\mathbf{3.99 \pm 0.04}$   \\
                                              &                        & \maniptrans     & $\mathbf{3.93 \pm 0.12}$   & $\mathbf{50.81 \pm 5.08}$   & $11.91 \pm 3.74$         & $10.67 \pm 3.93$           \\
            \midrule
            \multirow{2}{*}{81a95@5}          & \multirow{2}{*}{left}  & \spider         & $6.77 \pm 5.14$            & $44.97 \pm 21.95$           & $\mathbf{4.90 \pm 0.13}$ & $\mathbf{3.71 \pm 0.04}$   \\
                                              &                        & \maniptrans     & $\mathbf{3.13 \pm 2.85}$   & $\mathbf{38.57 \pm 25.20}$  & $6.61 \pm 2.34$          & $4.88 \pm 2.33$            \\
            \midrule
            \midrule
            \multirow{2}{*}{\textbf{Overall}} & \multirow{2}{*}{--}    & \spider         & $23.12 \pm 12.91$          & $\mathbf{54.44 \pm 22.43}$  & $\mathbf{4.84 \pm 0.31}$ & $\mathbf{4.08 \pm 0.47}$   \\
                                              &                        & \maniptrans     & $\mathbf{23.03 \pm 12.93}$ & $71.75 \pm 32.30$           & $139.01 \pm 388.16$      & $138.51 \pm 388.12$        \\
            \bottomrule
        \end{tabular}%
    }
\end{table*}

\subsection{\dexmachina Retargeting Results}
\label{subsec:dexmachina_results}

We evaluate \spider on bimanual dexterous manipulation tasks from the \arctic dataset~\citep{fanARCTICDatasetDexterous2023}, using the \dexmachina framework~\citep{mandiDexMachinaFunctionalRetargeting2025} and the Genesis physics engine.
We use the Inspire dexterous hand in a bimanual configuration (two hands manipulating articulated objects simultaneously).
To ensure a fair comparison, \spider directly uses the same RL training environment and reward function as \dexmachina, and all evaluation metrics are computed identically for both methods. We remove the object virtual joint curriculum from \dexmachina.

\textbf{RL Baseline Training.}
We use the \dexmachina RL policy as our baseline.
\dexmachina trains one RL policy per \arctic clip using the Genesis physics engine with the Inspire hand in a bimanual configuration.
The action space consists of 36 dimensions in hybrid mode: for each hand, 6 wrist force/torque dimensions (3 position + 3 rotation) and 12 finger joint position targets.
The RL policy is deterministic at evaluation time; since the initial state is fixed, the RL policy produces identical rollouts across random seeds.

\textbf{RL Evaluation.}
We evaluate the PPO policy once per task with 5 seeds and report the mean and std of the metrics.

\textbf{Reward Setting.}
To ensure \spider optimizes the same objective as the RL baseline, we delegate reward computation to \dexmachina's reward function but remove the action penalty term for efficiency (since our action is smooth by interpolation).

\textbf{Curriculum Learning.}
In \spider retargeting, we \emph{disable curriculum learning} in \dexmachina, so the object joints are always free (passive) throughout optimization.
For \dexmachina, we use the default curriculum learning settings where object joint stiffness and damping are annealed from high values (providing a guiding force) to zero across training iterations.

\textbf{SPIDER Retargeting.}
We run \spider with 1024 parallel Genesis environments.
\spider optimizes the full 36-dim action space (both hands' wrist and finger actions) using sampling-based MPC.
Each task is run with 5 random seeds.
All hyperparameters are listed in \Cref{tab:spider-dexmachina-params}.

\begin{table*}[h]
    \centering
    \caption{\spider hyperparameters for the \dexmachina (Genesis) experiments.}
    \label{tab:spider-dexmachina-params}
    \begin{tabular}{lcc}
        \toprule
        \textbf{Parameter}              & \textbf{Symbol}          & \textbf{Value}         \\
        \midrule
        \multicolumn{3}{l}{\textit{Simulation}}                                             \\
        Simulator                       & --                       & Genesis (GPU)          \\
        Simulation timestep             & $\Delta t_{\text{sim}}$  & $0.02$ s (50 Hz)       \\
        Control timestep                & $\Delta t_{\text{ctrl}}$ & $0.2$ s (5 Hz)         \\
        Knot point spacing              & $\Delta t_{\text{knot}}$ & $0.2$ s                \\
        Action dimensions               & $n_u$                    & $36$                   \\
        \midrule
        \multicolumn{3}{l}{\textit{Sampling-based Optimization}}                            \\
        Planning horizon                & $H$                      & $0.4$ s (20 sim steps) \\
        Parallel samples                & $N_w$                    & $1024$                 \\
        Max sampling iterations         & $N$                      & $32$                   \\
        Early stopping threshold        & $\epsilon$               & $0.01$                 \\
        Early stopping patience         & --                       & $4$ iterations         \\
        \midrule
        \multicolumn{3}{l}{\textit{Noise Schedule}}                                         \\
        Initial noise scale             & $\sigma_{\text{first}}$  & $0.5$                  \\
        Final noise scale (horizon end) & $\sigma_{\text{last}}$   & $1.0$                  \\
        Noise decay per iteration       & $\beta$                  & $0.87$                 \\
        Wrist pos/rot noise scale       & --                       & $1.0$                  \\
        Joint noise scale               & --                       & $1.0$                  \\
        \midrule
        \multicolumn{3}{l}{\textit{Reward}}                                                 \\
        Reward mode                     & --                       & RL training reward     \\
        Terminal reward scale           & $\alpha_T$               & $10.0$                 \\
        Sampling temperature            & $\tau$                   & $1.0$                  \\
        \midrule
        \multicolumn{3}{l}{\textit{DexMachina-specific}}                                    \\
        Imitation reward weight         & $w_{\text{imi}}$         & $0.3$                  \\
        Contact reward weight           & $w_{\text{con}}$         & $3.0$                  \\
        BC reward weight                & $w_{\text{bc}}$          & $0.3$                  \\
        Object joint stiffness          & $k_p$                    & $0.0$ (no curriculum)  \\
        Object joint damping            & $k_v$                    & $0.0$ (no curriculum)  \\
        \bottomrule
    \end{tabular}
\end{table*}

\textbf{Evaluation Metrics.}
We evaluate object tracking performance using three metrics computed at each simulation timestep in the original \dexmachina paper~\citep{mandiDexMachinaFunctionalRetargeting2025}, comparing the simulated object state against the ground-truth reference trajectory from the \arctic dataset:
\begin{itemize}
    \item $d_\text{pos}$: L2 position distance (m) between simulated and reference object positions, $d_\text{pos} = \lVert \mathbf{p} - \mathbf{p}^* \rVert_2$.
    \item $d_\text{rot}$: geodesic rotation distance (rad) on SO(3) between simulated and reference orientations, computed as $d_\text{rot} = 2 \arcsin\bigl(\min(\lVert \mathrm{Im}(\mathbf{q} \otimes \bar{\mathbf{q}}^*) \rVert_2,\; 1)\bigr)$, where $\mathbf{q}$ and $\mathbf{q}^*$ are unit quaternions in $(w, x, y, z)$ convention and $\bar{\mathbf{q}}^*$ denotes the conjugate.
    \item $d_\text{arti}$: articulation distance (rad), the absolute difference in the object's articulated joint angle, $d_\text{arti} = |a - a^*|$.
\end{itemize}
All metrics are averaged over the trajectory length, then reported as mean $\pm$ std across 5 seeds per task for \spider.
For the RL baseline, we report the single deterministic evaluation, as it produces identical rollouts across seeds (std $= 0$).
In the original \dexmachina paper, they also report the success rate by computing Average Distance Area Under the Curve (ADD-AUC) for each object part separately.
However, since this metric is not open source, we directly report the raw metrics for comparison.

\textbf{Experiment Results.}
We evaluate on 7 bimanual dexterous manipulation tasks from the \arctic dataset (box, ketchup, ketchup-u02, laptop, mixer, notebook, waffleiron), each defined by a 200-frame clip at 50\,Hz.
Each task is named in the format of \texttt{<object-name>-<start-frame>-<end-frame>}, similar to the original \dexmachina paper.
Full results are shown in \Cref{tab:dexmachina_comparison}.

\begin{table*}[t]
    \centering
    \caption{Multi-seed comparison between \spider and \dexmachina RL baseline on 7 bimanual tasks from the \arctic dataset. Both methods are evaluated with 5 seeds per task. Metrics: $d_\text{pos}$ = object position distance (m), $d_\text{rot}$ = object rotation distance (rad), $d_\text{arti}$ = articulation distance (rad). Values reported as mean $\pm$ std. Best per task in \textbf{bold}.}
    \label{tab:dexmachina_comparison}
    \resizebox{\textwidth}{!}{%
        \begin{tabular}{lc|ccc}
            \toprule
            \textbf{Task}                           & \textbf{Method} & $d_\text{pos}$ (m) $\downarrow$ & $d_\text{rot}$ (rad) $\downarrow$ & $d_\text{arti}$ (rad) $\downarrow$ \\
            \midrule
            \multirow{2}{*}{box-30-230}             & \spider         & $\mathbf{0.0172 \pm 0.0069}$    & $\mathbf{0.0988 \pm 0.0198}$      & $\mathbf{0.0929 \pm 0.0801}$       \\
                                                    & \dexmachina     & $0.0906 \pm 0.0373$             & $0.3709 \pm 0.1155$               & $0.2897 \pm 0.0522$                \\
            \midrule
            \multirow{2}{*}{ketchup-30-230}         & \spider         & $\mathbf{0.8917 \pm 0.1311}$    & $\mathbf{1.7862 \pm 0.1723}$      & $\mathbf{0.7836 \pm 0.0850}$       \\
                                                    & \dexmachina     & $1.1865 \pm 0.3021$             & $2.0616 \pm 0.1768$               & $0.8057 \pm 0.0806$                \\
            \midrule
            \multirow{2}{*}{ketchup-30-230-s01-u02} & \spider         & $\mathbf{0.7573 \pm 0.3397}$    & $\mathbf{1.5418 \pm 0.3500}$      & $\mathbf{1.1717 \pm 0.0227}$       \\
                                                    & \dexmachina     & $1.1569 \pm 0.2231$             & $2.0379 \pm 0.2021$               & $1.1863 \pm 0.0851$                \\
            \midrule
            \multirow{2}{*}{laptop-30-230}          & \spider         & $\mathbf{0.0427 \pm 0.0171}$    & $0.2155 \pm 0.0726$               & $1.0548 \pm 0.5011$                \\
                                                    & \dexmachina     & $0.0573 \pm 0.0840$             & $\mathbf{0.1980 \pm 0.2344}$      & $\mathbf{0.4094 \pm 0.3296}$       \\
            \midrule
            \multirow{2}{*}{mixer-30-230}           & \spider         & $\mathbf{0.0176 \pm 0.0098}$    & $\mathbf{0.1203 \pm 0.0360}$      & $\mathbf{0.1440 \pm 0.0118}$       \\
                                                    & \dexmachina     & $0.1211 \pm 0.1563$             & $0.7360 \pm 0.2814$               & $0.3654 \pm 0.1230$                \\
            \midrule
            \multirow{2}{*}{notebook-30-230}        & \spider         & $0.9968 \pm 0.0245$             & $1.9491 \pm 0.2248$               & $\mathbf{0.7024 \pm 0.0007}$       \\
                                                    & \dexmachina     & $\mathbf{0.4969 \pm 0.0682}$    & $\mathbf{1.1062 \pm 0.2850}$      & $0.7468 \pm 0.1131$                \\
            \midrule
            \multirow{2}{*}{waffleiron-30-230}      & \spider         & $\mathbf{0.2224 \pm 0.1585}$    & $\mathbf{1.0033 \pm 0.2985}$      & $0.5528 \pm 0.3142$                \\
                                                    & \dexmachina     & $0.5113 \pm 0.0682$             & $1.1791 \pm 0.0903$               & $\mathbf{0.5480 \pm 0.0753}$       \\
            \midrule
            \midrule
            \multirow{2}{*}{\textbf{Overall}}       & \spider         & $\mathbf{0.4208 \pm 0.4096}$    & $\mathbf{0.9593 \pm 0.7562}$      & $0.6432 \pm 0.3840$                \\
                                                    & \dexmachina     & $0.5172 \pm 0.4485$             & $1.0985 \pm 0.6854$               & $\mathbf{0.6216 \pm 0.2915}$       \\
            \bottomrule
        \end{tabular}%
    }
\end{table*}

\spider wins on 6/7 tasks for $d_\text{pos}$, 5/7 for $d_\text{rot}$, and 5/7 for $d_\text{arti}$.
On average, \spider achieves lower position error ($0.421$ vs.\ $0.517$\,m) and rotation error ($0.959$ vs.\ $1.099$\,rad), while the RL baseline achieves slightly lower articulation error ($0.622$ vs.\ $0.643$\,rad).
Notably, \spider achieves these results without any task-specific training, using only test-time optimization with the same reward function as the RL baseline.

\section{Virtual Contact Guidance Details}
\label{sec:contact_guidance_details}

\graybold{Solution Ambiguity.}
For a retargeting method to be valid, it is also important to preserve the human-preferred contact in the demonstration.
Take the stir stick manipulation in~\Cref{fig:guidance} as an example: the robot can hold the stick either between the thumb and index finger or between the index and middle finger; both achieve the task, but the former is more natural and aligns with the human demonstration. However, due to the non-convex cost landscape of \Cref{eq:dynamical_retargeting_problem}, there may be multiple solutions that achieve similar object motion with different contact behaviors.
As a result, a sampling-based optimizer may converge to alternative contact modes, leading to implausible contact trajectories in a demonstration even when the object trajectory is reproduced.

\graybold{Virtual Contact Guidance.}
To guide sampling towards the desired contact mode, we introduce \emph{virtual contact guidance}. This guidance factor expands the feasible solution set and biases the optimization process towards the target configuration.
Unlike a soft contact reward or cost - which may still fail when the desired mode is harder to sample - virtual contact guidance explicitly \emph{enlarges} the basin of attraction to make it easier to sample from, as illustrated in~\Cref{fig:guidance}.
Our method applies a virtual constraint between intended contact points on the object and robot fingers, ``sticking'' the object to the target configuration in early stages and gradually relaxing this constraint in a curriculum-like fashion.
Concretely, we impose a virtual constraint that maintains the relative position between contact pairs. For the $k$-th object-hand contact pair (e.g., handle-thumb), we define the current relative position as ${}^\text{robot}p_{k,t}^\text{object} = p^\text{robot}_{k,t} - p^\text{object}_{k,t}$ and the desired reference position as ${}^\text{robot}p_{k,t}^{\text{object,ref}} = p^\text{robot,ref}_{k,t} - p^\text{object,ref}_{k,t}$.
The constraint is activated when the contact indicator $c_{k,t} = 1$, which occurs when the reference relative position is within the contact threshold: $c_{k,t} = \mathbf{1}(\|{}^\text{robot}p_{k,t}^{\text{object,ref}}\|_2 \leq \epsilon_\text{contact})$.
The constraint strength is controlled by a penalty parameter $\eta_i \rightarrow \infty$ when $i \rightarrow N$:
\begin{equation}
    c_{k,t} \|{}^\text{robot}p_{k,t}^\text{object} - {}^\text{robot}p_{k,t}^{\text{object,ref}}\|_2^2 \leq \eta_i
\end{equation}
This strategy connects to prior ideas: virtual object constraint~\citep{mandiDexMachinaFunctionalRetargeting2025} $\|p_{k,t}^\text{object} - p_{k,t}^{\text{object,ref}}\|_2^2 \leq \eta$, which expands feasibility around absolute object states in RL, and contact cost~\citep{lakshmipathyKinematicMotionRetargeting2024}, biasing optimization towards desired hand--object \emph{relative} states. Our formulation reduces sampling complexity while preserving the intended contact sequence by maintaining \emph{relative} hand--object relationships.

\graybold{Robustness against imperfect reference contact.} It is important to note that, in order to maintain robustness against imperfect demonstrations with noisy or unstable contact, virtual constraints should be selectively relaxed rather than enforced indiscriminately.
To achieve this, we apply a contact filter that detects unstable interactions: if a contact duration is shorter than $t_{c,\text{min}}$ or if the contact point drifts more than $d_{c,\text{max}}$ during that period, the contact is classified as unstable and the corresponding virtual constraint is disabled. This implementation ensures that only reliable contacts contribute to guidance, thereby preventing noisy demonstrations from biasing the optimization process.

\end{document}